\documentclass{article}

\usepackage{microtype}
\usepackage{graphicx}
\usepackage{booktabs} %
\usepackage{todonotes}
\usepackage{amsmath,amsfonts,amssymb,amsthm,bm}
\usepackage{nicefrac}

\usepackage{caption}
\usepackage{subcaption}

\usepackage[acronym, nomain, nonumberlist]{glossaries}
\makenoidxglossaries

\newacronym{mnist}{MNIST}{Modified National Institute of Standards and Technology}
\newacronym{ode}{ODE}{Ordinary Differential Equation}
\newacronym{sgd}{SGD}{Stochastic Gradient Descent}
\newacronym{relu}{ReLU}{Rectified Linear Unit}
\newacronym{scm}{SCM}{Soft Committee Machine}
\newacronym{ewc}{EWC}{Elastic Weight Consolidation}
\newacronym{iid}{i.i.d.}{independent and identically distributed}
\newacronym{ntk}{NTK}{Neural Tangent Kernel}

\usepackage{tikz}
\usepackage{pgfplots}
\pgfplotsset{compat=newest}

\graphicspath{{figs/}{figs/}{figs/}}
\usepackage{hyperref}

\usepackage[accepted]{icml2022}

\icmltitlerunning{\HPname~Hypothesis for Catastrophic Forgetting}

\usepackage{setspace}
\usepackage{enumitem}
\setlist[itemize]{noitemsep}

\usepackage{xcolor}

\newcommand{\hpname}{Maslow's hammer}
\newcommand{\HPname}{Maslow's Hammer}

\allowdisplaybreaks

\begin{document}

\twocolumn[
\icmltitle{\HPname~for Catastrophic Forgetting:\\ Node Re-Use vs Node Activation}

\begin{icmlauthorlist}
\icmlauthor{Sebastian Lee}{imperial,swc,gatsby}
\icmlauthor{Stefano Sarao Mannelli}{swc,gatsby}
\icmlauthor{Claudia Clopath}{imperial,swc}
\icmlauthor{Sebastian Goldt}{sissa}
\icmlauthor{Andrew Saxe}{swc,gatsby,cifar}
\end{icmlauthorlist}

\icmlaffiliation{imperial}{Imperial College, London, UK}
\icmlaffiliation{sissa}{International School of Advanced Studies (SISSA), Trieste, Italy}
\icmlaffiliation{cifar}{CIFAR Azrieli Global Scholars program, CIFAR, Toronto, Canada}
\icmlaffiliation{gatsby}{Gatsby Computational Neuroscience Unit, UCL}
\icmlaffiliation{swc}{Sainsbury Wellcome Centre, UCL}

\icmlcorrespondingauthor{Sebastian Lee}{sebastian.lee14@imperial.ac.uk}
\icmlcorrespondingauthor{Andrew Saxe}{a.saxe@ucl.ac.uk}

\icmlkeywords{Machine Learning, ICML}

\vskip 0.3in
]

\printAffiliationsAndNotice{}  %

\begin{abstract}
Continual learning---learning new tasks in sequence while maintaining performance on old tasks---remains particularly challenging for artificial neural networks. Surprisingly, 
the amount of forgetting does not increase with the dissimilarity between the learned tasks, but appears to be worst in an intermediate similarity regime.

In this paper we theoretically analyse both a synthetic teacher-student framework and a real data setup to provide an explanation of this phenomenon that we name \hpname~hypothesis. Our analysis reveals the presence of a trade-off between node activation and node re-use that results in worst forgetting in the intermediate regime. Using this understanding we reinterpret popular algorithmic interventions for catastrophic interference in terms of this trade-off, and identify the regimes in which they are most effective.

\end{abstract}

\newcommand{\xb}{\mathbf{x}}
\newcommand{\wb}{\mathbf{W}}
\newcommand{\wvecb}{\mathbf{w}}
\newcommand{\vb}{\mathbf{v}}
\newcommand{\hb}{\mathbf{h}}
\newcommand{\zb}{\mathbf{Z}}

\colorlet{shadecolor2}{white}
\colorlet{shadecolor3}{white}

\section{Introduction}

Artificial neural networks have reached astonishing performance in a number of different applications~\cite{silver2016mastering, rajkomar2019machine, devlin2019bert, tunyasuvunakool2021highly}, but they tend to perform poorly when they have to solve a sequence of learned tasks~\cite{kemker2018measuring}. The ineffectiveness of deep learning algorithms in this learning paradigm---known as continual learning or lifelong learning---is strikingly different from observations in human and animal learning, where tasks can effectively be learned sequentially and interference is a rarity~\cite{barnett2002and, calvert2004handbook, mareschal2007neuroconstructivism, pallier2003brain}. Neuroscientists and psychologists have been interested in the mechanisms underpinning this ability for some time~\cite{mccloskey1989catastrophic, cichon2015branch, yang2014sleep, flesch2018comparing}; more recently the engagement of the machine learning community with this paradigm has also grown as the focus shifts from performance on single tasks to distributions of tasks and learning from the real world~\cite{parisi2019continual}. 

The key difficulty of continual learning is avoiding so-called~\emph{catastrophic forgetting} or~\emph{catastrophic interference}~\cite{mccloskey1989catastrophic,ratcliff1990connectionist}, 
the phenomenon of deteriorating performance on earlier tasks when learning later tasks. 
Humans are very good continual learners, and various biological mechanisms have 
been proposed to account for the brain's ability to combat forgetting~\cite{McClelland1995}. 
On the other hand artificial systems, in particular neural networks trained 
with gradient descent algorithms, suffer badly from catastrophic forgetting~\cite{goodfellow2013empirical}. 
This has prompted research into methods to augment vanilla gradient descent with additional 
elements specifically aimed at reducing forgetting~\cite{parisi2019continual}, including some that take inspiration form the aforementioned biological mechanisms~\cite{hinton1987using, robins1995catastrophic, gepperth2016bio, rebuffi2017icarl}. 

In addition to this algorithmic line of research, there is growing interest in understanding \emph{why} forgetting affects deep
learning so severely and what the main drivers of forgetting are. \citet{ramasesh2020anatomy}  performed a series of systematic experiments in a range of architectures and training setups and made the counterintuitive observation that catastrophic forgetting is worst between tasks of \emph{intermediate} similarity. \citet{lee2021continual} then analysed the impact of task similarity on continual learning in a solvable model of two-layer neural networks and found the same non-monotonic relationship between task similarity and forgetting. Despite these results, the precise mechanism that makes intermediate task similarity the worst has remained unclear.

Here, we describe a possible mechanism that drives catastrophic forgetting in two-layer neural networks trained in a teacher-student setup. Our main contributions can be summarised as follows:
\begin{itemize}
    \item \hpname~hypothesis to explain the observation that intermediate similarity is worst for forgetting ~\cite{ramasesh2020anatomy,lee2021continual} in terms of a trade-off between node (hidden unit) re-use and node activation;
    \item Evidence from both the student-teacher framework and a data-mixing image classification paradigm to support the \hpname~hypothesis;
    \item An empirical study of how various methods of alleviating forgetting impact the relationship between task similarity and forgetting;
    \item Observation of `\textit{catastrophic slowing}', whereby despite tremendous advantages in aligned and orthogonal task settings, interleaving can be inferior to regularisation methods in intermediate similarity regimes.
\end{itemize}

\textbf{Further related work}

Recent investigations into theoretical questions related to continual learning include work by~\citet{mirzadeh2021wide}, who study the relationship between network width, depth and forgetting; showing that wider and shallower networks are less affected by forgetting. \citet{bell2021behavioral} take inspiration from methodologies in psychology to design tasks aimed at investigating catastrophic forgetting, including in relation to loss surfaces and the interplay between semantic and perceptual information. Meanwhile \citet{shen2021algorithmic} gained interesting insights into algorithmic components of two-layer neural circuits---not unlike those studied here---that allow fruit flies to mitigate interference, including sparse coding and associative learning. Arguably closest in nature to our work are those of~\citet{asanuma2021statistical} and~\citet{doan2020theoretical}, who also investigate the effect of task similarity on forgetting but in linear regression and~\gls{ntk}~\cite{jacot2018neural} regimes respectively.

A related theoretical line of research concerns transfer learning where the focus is not on forgetting, but on the boost that features learned on upstream tasks can provide new tasks \cite{tan2018survey}. \citet{dhifallah2021phase}~\&~\citet{gerace2021probing} have analysed, on single-layer and two-layer networks respectively, the effect of similarity between tasks and data scarcity on performance in the downstream task using methods similar to ours.

On the more applied side of work on continual learning, concerned with developing
algorithms to combat forgetting in neural networks, there is a larger 
body of literature (see e.g.~\citet{parisi2019continual} for a review). Methods can broadly be split into 
three categories: regularisation~\cite{pmlr-v70-zenke17a, li2017learning,kirkpatrick2017overcoming}; 
dynamic architectures~\cite{rusu2016progressive, draelos2017neurogenesis,zhou2012online}; and replay~\cite{McClelland1995,shin2017continual}. 
A more recent set of approaches concerns itself with explicitly learning modular representations for compositionality~\cite{mendez2021lifelong, veniat2021efficient, ostapenko2021continual}; our investigation into node specialisation connects naturally to some of these concepts.
In this work we look at
aspects of each of these method families: insofar as our setups have separate
heads for each task, we implicitly consider adaptive architectures; in~\autoref{sec: combating} we explicitly investigate how~\gls{ewc} and interleaved replay affect the relationship between
task similarity and forgetting.
\section{Continual Learning Setup}
\label{sec:setup}
In this work we consider two paradigms to study continual learning: a synthetic framework using the teacher-student model, and a real data framework where similarity is parameterized by a mixing parameter.

\paragraph{Teacher-Student Framework.}\label{sec: teacher_student}

\input{figs_code/figure_1}

The key idea of the teacher-student setup is to train a neural network, the student, on a data set generated by taking random inputs and propagating them through a fixed neural network with random weights called the teacher~\cite{gardner1989, seung1992statistical, engel2001statistical}. We will consider two-layer networks with output 
$\phi(\xb; \wb, \vb) = \sum_{l=1}^L {\vb}_l g\left(\nicefrac{{\wvecb}_l\cdot\xb}{\sqrt{D}}\right)$;
where~$D$ is the input dimension, $L$ is the number of hidden units, 
$\wb\in\mathbb{R}^{L\times D}$ are the first layer weights, 
$\vb\in\mathbb{R}^{L}$ are the second layer weights, $g$ is the activation function, 
and $\xb\in\mathbb{R}^D$ is the input vector. These inputs are sampled i.i.d. (independent and identically distributed) from the standard normal distribution. 

While the framework allows for any number of tasks, for concreteness we consider training the student on a succession of two tasks (which we denote throughout by $\dagger$ and $\ddag$). In the $i^{\text{th}}$ phase of training 
(i.e. training on the $i^{\text{th}}$ task), the supervision labels for the 
student are generated from the~$i^{\text{th}}$ teacher by $y^i=\phi(\xb; \wb^i, \vb^i)$ and the student 
outputs are given by $\hat{y}^i=\phi(\xb; \wb, \hb^i)$. Training is performed with~\gls{sgd} on the squared error $(y^i-\hat{y}^i)^2$. Note that the student shares first layer weights~$\wb$ across tasks but has separate head weights $\hb^i$ for each task. A sketch of this setup is shown in~\autoref{fig:1a}. 

The key quantity that we would like to compute is the  generalisation error of the student with respect to the $i^{\text{th}}$ teacher,
\begin{multline}
  \label{eq:eg}
  \epsilon^i(\wb, \hb^i, \wb^i, \vb^i)
  \equiv\\
  \frac{1}{2}\left\langle\right.[\phi(\xb; \wb^i, \vb^i) \left.- \phi(\xb; \wb, \hb^i)]^2\right\rangle.
\end{multline}
Note that due to the separate heads, the generalisation errors are well-defined with respect to both teachers regardless of which is currently providing the labels. From these generalisation errors, one can further define quantities analogous to forgetting and transfer from one teacher/task to the next as differences in generalisation error.

The advantage of the teacher-student setup is that by providing full control over the input distribution, the similarity between tasks can be precisely tuned by controlling the relationship between teacher weights. Furthermore, we can give the student the right number of parameters to learn all teachers perfectly, at least in principle. Finally, the dynamics of the student in this setup can be solved exactly, yielding an~\gls{ode} that describes its average dynamics (shown for single teacher-student by \citet{saad1995exact}, and later for continual learning by~\citet{lee2021continual}). A key observation from this framework is that intermediate task similarity is worst for forgetting; we reproduce this in~\autoref{fig:1b} (c.f.~Fig. 3 in \citet{lee2021continual} and ~\autoref{app: exp_details} for details).
In this work, we propose a mechanism responsible for this behaviour, and investigate the impact 
of various methods for combating forgetting on this relationship.

\paragraph{Data-Mixing Framework.}\label{sec: data_mixing}

To probe how well our findings translate to more realistic data distributions, we complement the teacher-student framework with a data-mixing approach similar to that 
introduced by~\citet{ramasesh2020anatomy}. This procedure gives a notion of control
over the similarity between any pair of tasks in that the input distribution is composed of a mixture between two separate datasets, where the mixing factor determines the similarity. 
More specifically, consider two separate datasets with equal cardinality in which 
inputs and outputs have the same dimensions across both datasets, i.e. 
$\mathcal{D}_1=\{\xb^1_i, y^1_i\}_{i=1}^N$ and $\tilde{\mathcal{D}}_2=\{\tilde{\xb}^2_i, \tilde{y}^2_i\}_{i=1}^N$. 
We can control the `similarity' between two successive tasks by first training on $\mathcal{D}_1$, 
followed by a mixture between $\mathcal{D}_1$ and $\tilde{\mathcal{D}}_2$. 
For a given mixing factor $\alpha$, this second dataset is given by
\begin{equation}
    \mathcal{D}_2^\alpha = \{\alpha\xb^1_i + (1-\alpha)\tilde{\xb}^2_i, \alpha y^1_i + (1-\alpha)\tilde{y}^2_i\}_{i=1}^N.
\end{equation}
Under this protocol, $\alpha=0$ corresponds to a completely new dataset and an 
entirely new task, whereas $\alpha=1$ corresponds to continuing to train on the 
first task.
\section{Intermediate Task Similarity}\label{sec: why}

Although numerous independent studies have found non-monotonic relationships between task similarity 
and forgetting in artificial neural networks~\cite{ramasesh2020anatomy, lee2021continual, asanuma2021statistical}, a convincing explanation for this result is still missing. Here, we propose such a mechanism, which we call \hpname~hypothesis. The starting point is to think about how individual nodes in the hidden layers of two-layer networks are re-purposed during continual learning. We first outline the intuition behind the hypothesis before 
presenting supporting evidence from both the teacher-student setup and networks trained on image data.

\subsection{\HPname~Hypothesis}\label{sec: hypothesis}

\input{figs_code/figure_2}

The trade-off between node re-use and node activation finds an analogy in a well-known cognitive bias in psychology: in the words of Abraham Maslow,  ``\textit{[...] it is tempting, if the only tool you have is a hammer, to treat everything as if it were a nail}''~\cite{maslow1966psychology}. In other words: if we have a nail we can (and should) use a hammer, but for a screw
we really need a different tool and should avoid using the hammer.

This phenomenon illustrates the choice that the student network makes in learning the new task. Consider the simplest case of a student with two hidden units trained on a pair of teachers with a single hidden unit each. Since the student has one set of head weights for each teacher, it has the capacity to achieve zero test error on both teachers at the same time, cf.~\autoref{fig:2a}.

During the first task, we expect the student to learn the first teacher. In doing so, we \emph{assume} a high degree 
of specialisation in the student whereby a subset of units in the network being trained become very important for the task
while others remain inactive or unimportant. In this specific case, this results in the student using one node to learn the first teacher and leaving the second node virtually inactive. We will show that this is the case in both the teacher-student setup and on the data mixture (cf. ~\autoref{fig:intermediate_evidence_overlaps}, ~\autoref{fig:data_mixing_evidence} and~\autoref{app: specialisation})

After the switch point there are three ways in which the student can learn the second teacher:~\emph{re-use} of the node that specialised to the first teacher,~\emph{activation} of the second (previously inactive) node, or a hybrid of the two, cf.~\autoref{fig:2c}. 
In order to minimise forgetting, the student could use the inactive units to learn the second teacher and leave the node that has specialised on 
the first teacher untouched, as schematised in~\autoref{fig:2b}. If the tasks are very related, however, it may be convenient to re-use previously activated nodes and leverage the features extracted in the first task, as in transfer learning \cite{tan2018survey}.
If the tasks are very dissimilar, the orthogonal teacher regime, the student network typically chooses to begin using its inactive node leaving the specialised one fairly intact, thus guarding against catastrophic interference.
The intermediate case represents the most difficult case for the student: the previously specialised node will be somewhat aligned to the second teacher, 
so there would be some transfer benefit to fine tuning the orientation of this node (\emph{re-use}), and making up any remaining difference with the previously inactive node. However under such a policy, unlike for the fully aligned teacher case, this would result in interference since the specialised node is moved. In sum, the Maslow's hammer hypothesis states that gradient descent dynamics bias towards re-using nodes when tasks are more similar (using a hammer for increasingly nail-like objects) and towards activation when tasks are more dissimilar (finding a different tool for decreasingly nail-like objects). This is most damaging when tasks are somewhat related, akin to breaking a screw when attempting to use a hammer.

\subsection{Evidence from the Teacher-Student Framework}\label{sec: student_teacher_evidence}

\input{figs_code/figure_3}

We first focus on the teacher-student setting where the model assures precise control. The framework detailed in~\autoref{sec: teacher_student} can be analysed exactly.
Indeed, a long line of work has shown that, as the input dimension tends to infinity, the generalisation error concentrates and can be understood purely in
terms of so-called `order parameters' of the system~\cite{mezard1987spin, engel2001statistical}. Here, the concentration hypothesis will be taken as a working hypothesis and verified numerically by comparing theory and simulations. 

We focus on the simplest setup to exhibit catastrophic forgetting, a two-task setting with a single task switch. Each teacher has a single hidden node and an output weight of norm one; the student has two hidden nodes. Among the crucial order parameters in the two-layer teacher-student scenario are the teacher-student overlaps
    $r_{km} \equiv \frac{1}{D} \wvecb_k^T \wvecb^\dagger_m$ and $u_{kp} \equiv \frac{1}{D} \wvecb_k^T \wvecb^\ddag_p$,
which measure the alignment between the weight of the $m^\text{th}$ ($p^\text{th}$) teacher $\dagger$ ($\ddag$) node and the $k^\text{th}$ student node. At the beginning of training, the random teacher weights and the randomly initialised student vector have very little overlap; throughout training, the student will improve its alignment with the teacher providing the labels, and hence its test error. Another order parameter is the overlap between student weights,
   $q_{k\ell} \equiv \frac{1}{D} \wvecb_k^T \wvecb_\ell$,
where the diagonal elements give the student node norms. The final crucial order parameter for our continual learning analysis, is the overlap between the first-layer weights of the two teachers,
    $v_{mp} = \frac{1}{D} (\wvecb^\dagger_m)^T \wvecb^\ddag_p$, which we abbreviate to $V$ (see~\autoref{app: ode}~\&~\autoref{app: overlap_generation} for details).
This order parameter describes the task similarity: orthogonal tasks ($V=0$) correspond to independent teacher weights, whereas perfectly similar tasks ($V=1$) have identical teacher weights up to permutations of the second-layer weights.

These order parameters obey a closed set of differential equations that describes their dynamics when training the student using~\gls{sgd} as described in~\autoref{sec:setup}. These were first derived by \citet{riegler1995} and \citet{saad1995exact}, and recently shown to be asymptotically exact by~\citet{goldt2019dynamics}. These methods have been used to explore a broad range of phenomena in neural networks, see \citet{saad2009line} for a summary of early work and \citet{yoshida2019datadependence, goldt2019modelling, refinetti2021classifying, saglietti2021analytical} for recent results. Here, we follow the derivation of similar equations for continual learning provided by \citet{lee2021continual}. 
In~\autoref{fig:intermediate_evidence_overlaps} we show the evolution of the solution to these~\glspl{ode} for a student trained on two successive teachers with several values of the similarity parameter $V$  (see~\autoref{app: ode} for details) These plots show evidence for the \hpname~hypothesis outlined in~\autoref{sec: hypothesis}. 
It is clear from~\autoref{fig:3a}. and~\autoref{fig:3b}. 
that there is strong specialisation in the student network. The magnitude of one student node is close to 1, 
and that node is almost fully aligned with the first teacher before the switch. 
Meanwhile the other node is essentially inactive before the switch. 
In order to minimise the amount of forgetting, the specialised node should not move after the switch (as per~\autoref{fig:2b}). 
However after the switch there is movement in both nodes. Let us consider different levels of similarity separately: 

\textbf{Fully aligned case} (yellow lines): there is an initial phase of movement 
in the specialised node before a reversion to the solution found for the first task. 
Meanwhile the previously inactive node remains largely inactive.
Although this solution does not use the spare capacity available to the student (outside the initial transient), it is very close to optimal behaviour in terms of forgetting.

\textbf{Fully orthogonal case} (purple lines): there is only a minimal deviation of the specialised node before it reverts 
to the solution found for the first task. On the other hand, 
there is complete activation and alignment to the second teacher 
of the second node. This can be seen in the darkest dashed line 
moving close to 1 in~\autoref{fig:3a}, and the darkest dashed line in~\autoref{fig:3c} 
moving close to -1 (sign is flipped by learned head weight). 
This solution is very close to the optimal one proposed by~\autoref{fig:hypothesis_sketch}. 

\textbf{Intermediate case} (turquoise lines): between the aligned and orthogonal cases, 
the student does a combination of re-using the previously 
specialised node and activating the previously inactive node. 
The former can be seen in the movement away from 1 in the solid lines 
in~\autoref{fig:3a} and the
movement away from -1 in the solid lines in~\autoref{fig:3b}. The latter can be seen in the movements away from 0 of the dashed lines in~\autoref{fig:3a} and the dashed lines in~\autoref{fig:3c}.

\input{figs_code/figure_4}

\paragraph{Re-use vs. activation tendency.}

An important aspect of this explanation is that more similar tasks will bias the network towards re-use whereas more orthogonal tasks will bias the network towards new activation. This is not obvious~\emph{a priori} and warrants closer inspection.
In~\autoref{fig:3d} we show the trajectory of the specialised student node norm around the switch point. Immediately after the switch there is a clear monotonic relationship between the teacher-teacher similarity and rate of movement in $q_{11}$. On the other hand,~\autoref{fig:3e} shows the inverse relationship for movement away from 0 in the norm of the inactive student node. To complete this part of the picture,~\autoref{fig:3f} shows the asymptotic generalisation error of the student with respect to the first teacher under a complete re-use scheme, which we compute retrospectively via:
\begin{equation}
    \epsilon^{\dagger}_{[\text{re-use}]} = \frac{1}{2}\left\langle[\phi(\xb; \wb^\dagger, \vb^\dagger) - \phi(\xb; \wb^\ddag, \hb^{\dagger*})]^2\right\rangle,\label{eq: reuse}
\end{equation}
where $\wb^\dagger$ is the first teacher feature weights, $\vb^\dagger$ is the first teacher head, $\wb^\ddag$ is the second teacher feature weights and $\hb^{\dagger*}$ is the component of the student's first head weight reading from the specialised node at the switch point. There is another clear decreasing monotonic relationship between cost in re-using the node and task similarity. Together these plots show the re-use and activation tendencies of the student in various similarity regimes, as well as the costs associated with these tendencies.

\subsection{Evidence from Data Mixing Framework}

In this section we use the protocol described in~\autoref{sec: data_mixing} to supplement the analysis in the synthetic framework with real data experiments. 
It is worth bearing in mind the following implications of changing settings:
(i) While previously task similarity was defined over input-output mappings and the input distribution was constant, here the similarity parameter influences both input-output mappings and input distributions;
(ii) Since the tasks themselves share many features (e.g. edges for image datasets), this contributes additional similarity such that even no mixing ($\alpha=0$) will give similar tasks to some extent.

In the teacher-student framework we can understand clearly how the student solution
relates to the task given by the teacher network from the overlap matrices. In the standard supervised learning setting there is no analogue.
Instead we define an empirical measure of node `\textit{importance}' that we use to 
investigate how different nodes in the network contribute to forgetting. 
For a given node $i$, we define its importance $I_i^t$ in relation to some task $t$ to be the change in test error when the output of that node is masked (see~\autoref{app: node_importance_def} for details). 
If a node is important to the network for a given task, the error will increase substantially when this node is masked and $I_i^t$ will be high.

In~\autoref{fig:4b} we see that for a two-layer network trained
on a Fashion MNIST~\cite{xiao2017/online} binary classification task, one or two nodes dominate in terms of 
importance at the switch point. Any forgetting that occurs after the switch point 
will hence be dominated by the behaviour of these nodes. Empirically, we then observe that these nodes remain more important for the second task when the task similarity is higher (see~\autoref{fig:4b} for a single seed). To visualise this more generally, we consider the dot product of the vector of node importances for the first task at the switch point $I^1$ with the vector of node importances for the second task at the end of training on both tasks $I^2$. If similar nodes are important for both tasks (re-use), this quantity is high; while if different nodes are important (activation), it is low. \autoref{fig:4c} shows that, just as in the teacher-student (see~\autoref{fig:3g}), $I^1\cdot I^2$ generally increases as a function of similarity in the data-mixing setup (see~\autoref{app:ramasesh_statistics} for details on statistics).

\section{Methods for Combating Forgetting}\label{sec: combating}

With a better understanding of the basis for the relationship between task similarity and catastrophic forgetting, 
a natural question to ask is how various commonly used methods to combat forgetting impact the picture.
These methods typically fall into three broad groups: dynamic architectures, where
capacity is added to deal with new tasks; regularisation, where a
penalty is added to the objective of later tasks to bias the network to solutions
compatible with earlier tasks; and replay, where data from previous tasks (or
representations thereof) are interleaved throughout training of later tasks.
By using the conventional continual learning protocol of one head per task, we
are arguably already operating in the dynamic architecture regime. Beyond that, we
study in this section one of the most widely used algorithms for combating 
forgetting,~\gls{ewc}; as well as interleaved replay, which we 
can implement straightforwardly in the teacher-student framework without storing 
data or training additional generative models.

\subsection{Elastic Weight Consolidation}

\gls{ewc}~\cite{kirkpatrick2017overcoming} applies a quadratic penalty to weight movement away from the solution for an earlier task and is modulated by the Fisher information of the weight for the earlier task. The penalty is motivated by a Laplace approximation of the posterior (conditional probability of the parameters given the data from the first task) where the mean and variance of the Gaussian approximation are given by the weights at the end of the first task, and the diagonal of the Fisher information matrix respectively. For a pair of tasks $A$ and $B$, and a neural network parameterised by $\theta$, the objective function for training on the second task is thus given by:
\vspace{-2mm}
\begin{equation}
    \mathcal{L}(\theta) = \mathcal{L}_B(\theta) + \frac{\lambda}{2}\sum_i F_i(\theta_i-\theta_{A,i}^*)^2, \label{eq: ewc}
\end{equation}
where $F_i$ is the $i^{\text{th}}$ element along the diagonal of the Fisher information matrix, $\theta_{A,i}^*$ is the value of the $i^{\text{th}}$ weight at the end of training on task $A$, and $\lambda$ is an additional hyperparameter controlling the strength of consolidation.
Specifically in the online learning setting of the two-layer student-teacher framework,~\gls{ewc} affects only the first layer weights since the head weights are not shared across tasks.

\begin{figure}
	\pgfplotsset{
		width=1.1\textwidth,
		height=0.8\textwidth,
		scaled x ticks=false,
		every tick label/.append style={font=\tiny},
		y label style={at={(axis description cs:-0.16, 0.5)}, rotate=0, anchor=south},
		x label style={at={(axis description cs:0.5, -0.45)}, rotate=0, anchor=south},
		xlabel={\scriptsize step, $s$},
		}
	\begin{subfigure}[b]{0.235\textwidth}
	\begin{tikzpicture}
        \fill[shadecolor2, opacity=0] (0, 0) rectangle (\textwidth, 0.75\textwidth);
        \node [anchor=north west] at (0, 0.75\textwidth) {\emph{(a)}};
        \node [anchor=north west] at (1.6, 0.75\textwidth) {\scriptsize $\lambda=0$};
		\begin{axis}
			[
			at={(0.9cm, 0.75cm)},
			anchor=south west,
			xmin=0, xmax=12000000,
			ymin=-5, ymax=0,
			ylabel={\scriptsize $\log\epsilon^\dagger$}
		]
		\addplot graphics [xmin=0, xmax=12000000,ymin=-5,ymax=0] {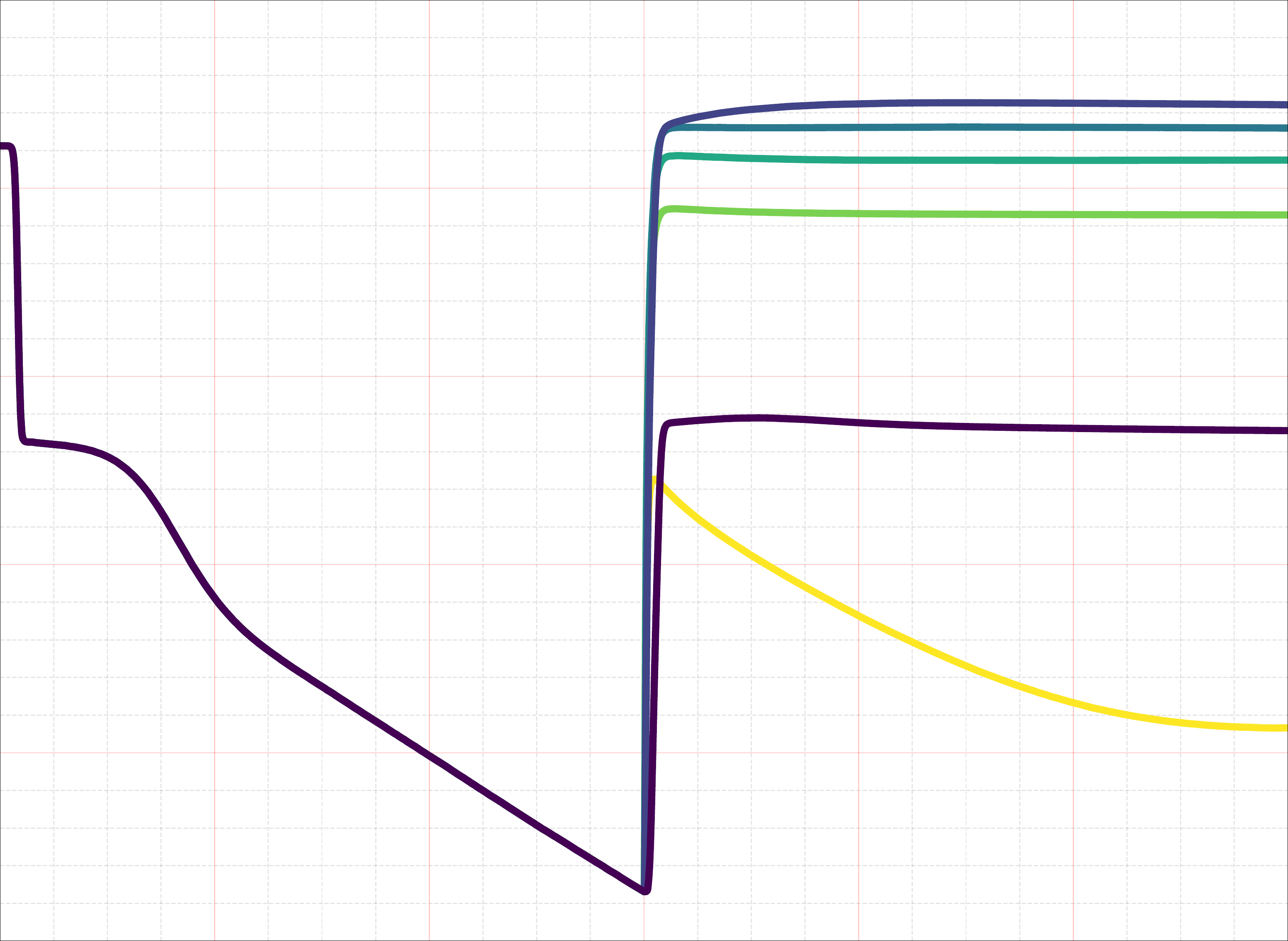};
		\end{axis}
	\end{tikzpicture}%
	\phantomcaption
	\label{fig:5a}
	\end{subfigure}
	\begin{subfigure}[b]{0.235\textwidth}
	\begin{tikzpicture}
        \fill[shadecolor2, opacity=0] (0, 0) rectangle (\textwidth, 0.75\textwidth);
        \node [anchor=north west] at (0, 0.75\textwidth) {\emph{(b)}};
        \node [anchor=north west] at (1.6, 0.75\textwidth) {\scriptsize $\lambda=100$};
		\begin{axis}
			[
			at={(0.9cm, 0.75cm)},
			anchor=south west,
			xmin=0, xmax=12000000,
			ymin=-5, ymax=0,
			ylabel={\scriptsize $\log\epsilon^\dagger$}
		]
		\addplot graphics [xmin=0, xmax=12000000,ymin=-5,ymax=0] {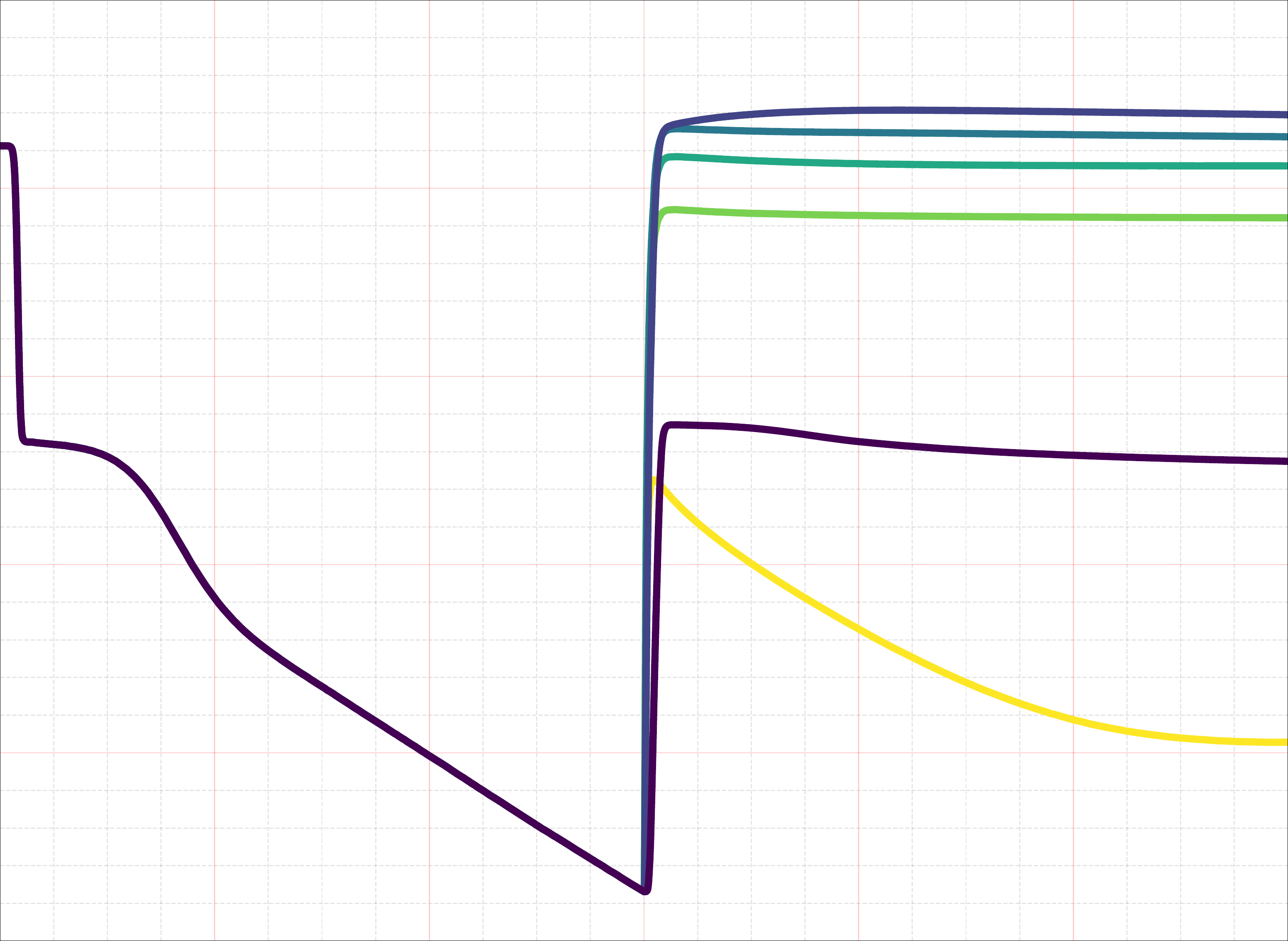};
		\end{axis}
	\end{tikzpicture}%
	\phantomcaption
	\label{fig:5b}
	\end{subfigure}
	\hspace{0em}
	\begin{subfigure}[b]{0.235\textwidth}
	\begin{tikzpicture}
        \fill[shadecolor2, opacity=0] (0, 0) rectangle (\textwidth, 0.75\textwidth);
        \node [anchor=north west] at (0, 0.75\textwidth) {\emph{(c)}};
        \node [anchor=north west] at (1.6, 0.75\textwidth) {\scriptsize $\lambda=1000$};
		\begin{axis}
			[
			at={(0.9cm, 0.75cm)},
			anchor=south west,
			xmin=0, xmax=12000000,
			ymin=-5, ymax=0,
			ylabel={\scriptsize $\log\epsilon^\dagger$}
		]
		\addplot graphics [xmin=0, xmax=12000000,ymin=-5,ymax=0] {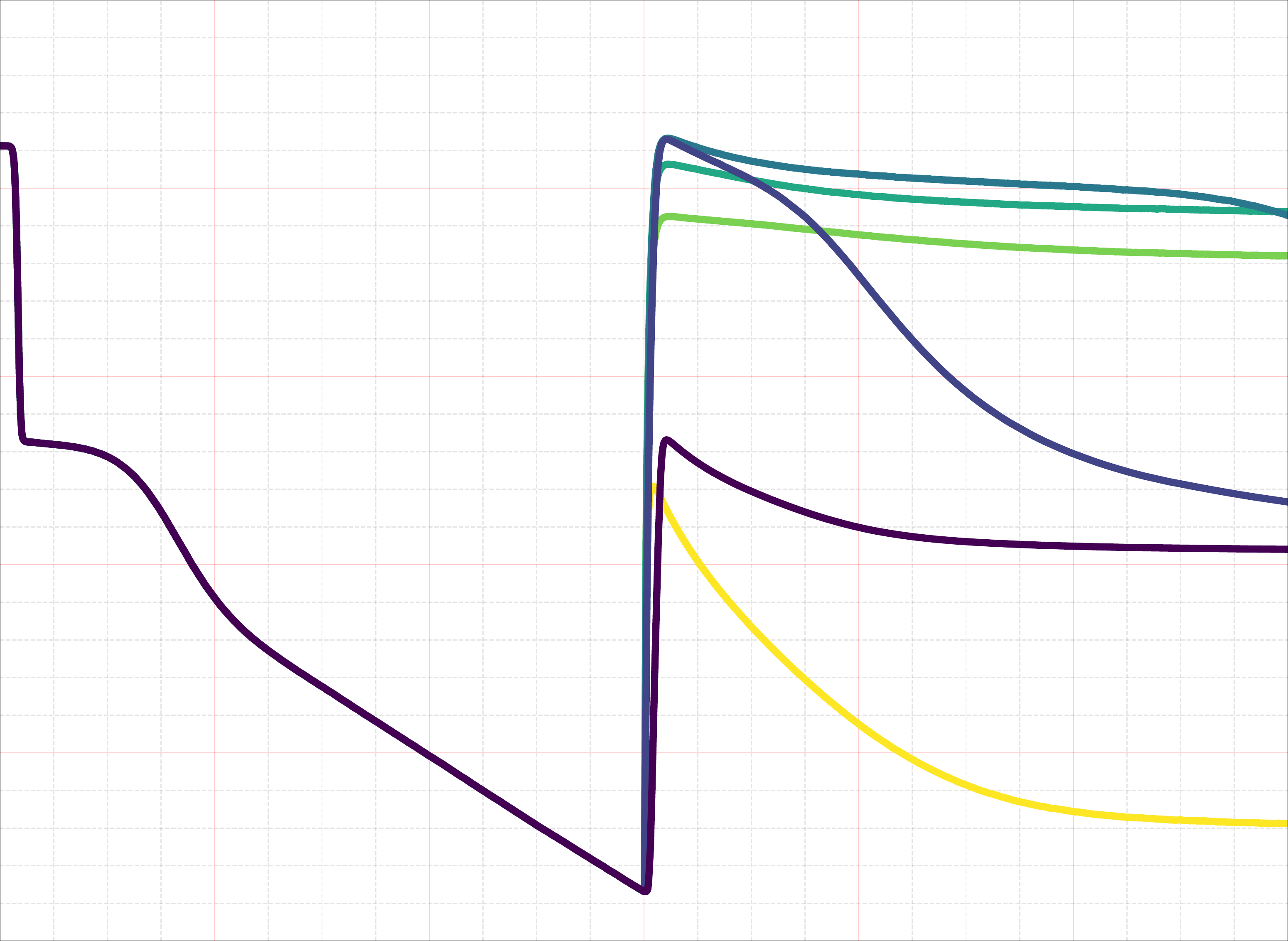};
		\end{axis}
	\end{tikzpicture}%
	\phantomcaption
	\label{fig:5c}
	\end{subfigure}
	\begin{subfigure}[b]{0.235\textwidth}
	\begin{tikzpicture}
        \fill[shadecolor2, opacity=0] (0, 0) rectangle (\textwidth, 0.75\textwidth);
        \node [anchor=north west] at (0, 0.75\textwidth) {\emph{(d)}};
        \node [anchor=north west] at (1.6, 0.75\textwidth) {\scriptsize $\lambda=10000$};
		\begin{axis}
			[
			at={(0.9cm, 0.75cm)},
			anchor=south west,
			xmin=0, xmax=12000000,
			ymin=-5, ymax=0,
			ylabel={\scriptsize $\log\epsilon^\dagger$}
		]
		\addplot graphics [xmin=0, xmax=12000000,ymin=-5,ymax=0] {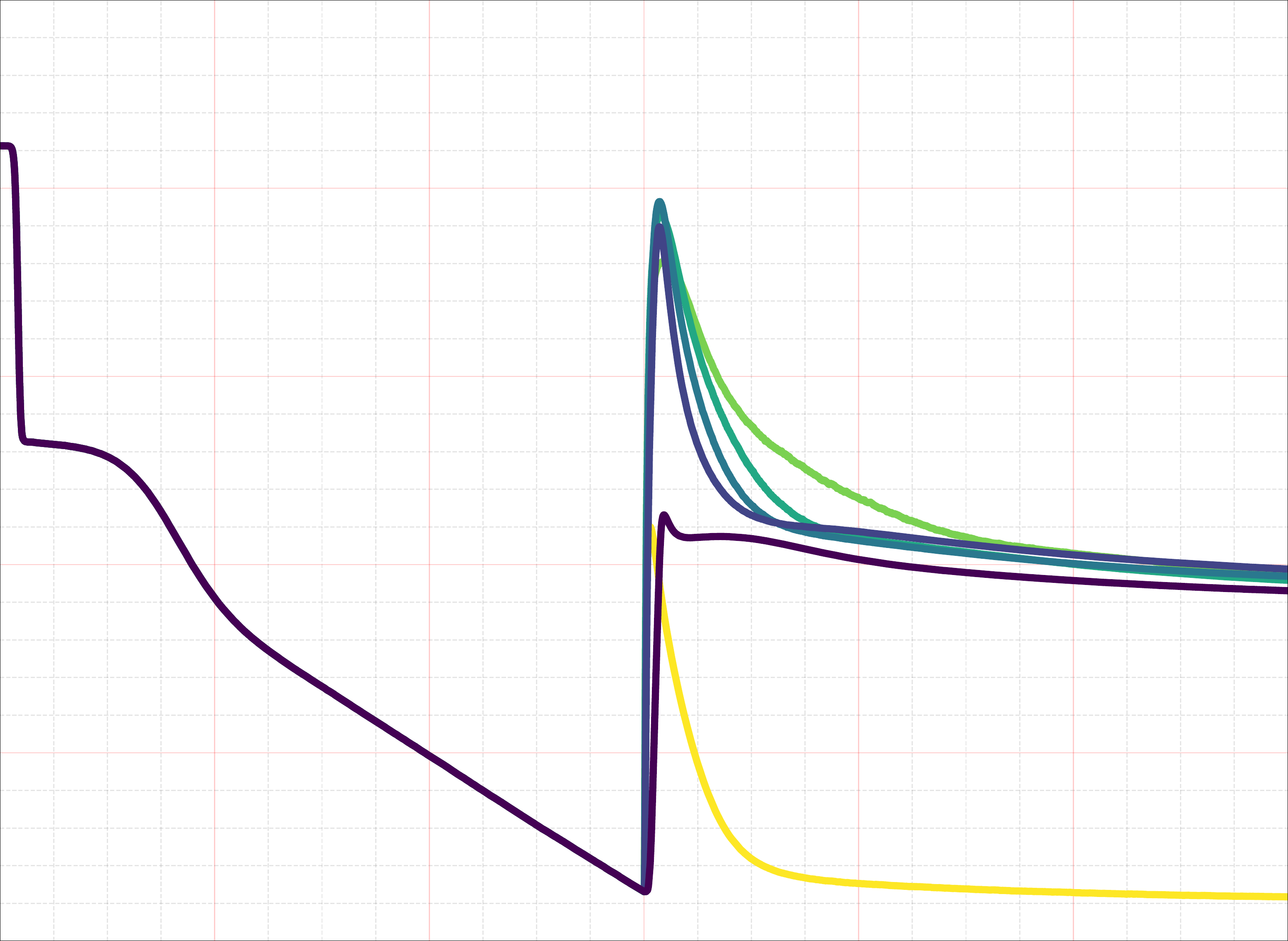};
		\end{axis}
	\end{tikzpicture}%
	\phantomcaption
	\label{fig:5d}
	\end{subfigure}
	\hspace{0em}
	\vspace{-2em}
	\caption[]{\textbf{Effect of EWC on task similarity vs. forgetting}  As the consolidation strength $\lambda$ increases, so too does the bias towards node activation. This happens first for networks trained on more similar teachers until for $\lambda=10000$, all trajectories regardless of $V$ have essentially collapsed onto one.\label{fig:ewc}}
	\vspace{-1em}
\end{figure}

In~\autoref{fig:ewc} we show the generalisation error curves for a student trained on a succession of two teachers with various degrees of similarity, this time training with the modified~\gls{ewc} objective in the second phase. Each subplot shows a different value for the strength of consolidation ($\lambda$ in~\autoref{eq: ewc}).

As the importance parameter increases and more weight is given to consolidation in the objective, forgetting generally reduces. In particular we see that the more similar the tasks are, the greater $\lambda$ needs to be to have an impact. Eventually for the largest value of $\lambda$ shown, all trajectories have collapsed onto a very similar learning trajectory. The exception is the trajectory corresponding to fully aligned teachers, which despite some improvement is comparatively less affected by~\gls{ewc}. We can understand these results through the lens of~\hpname: for fully aligned teachers, the student does not need to activate dormant nodes in the second task and can feasibly continue to use the specialised node, hence~\gls{ewc} has little effect. As the task similarity reduces, the propensity of the student to instead activate a new node increases. The effect of~\gls{ewc} is to intensify this increased propensity, in other words to amplify the bias to fresh node activation since movement in the weights contributing to the specialised node is penalised. Additionally, as the bias towards node re-use is  higher for more similar tasks, it takes a stronger push in the other direction (i.e. higher $\lambda$) to impact these trajectories. With a high enough $\lambda$ such that the bias to node re-use is effectively removed for all trajectories regardless of teacher similarity, the task in the second phase of learning is akin to learning a new random teacher with a \emph{tabula rasa} node. Hence the learning trajectories collapse onto one. We show later that although the conservatism induced by strong~\gls{ewc} (comparable in this setting to freezing a node), limits even the fully aligned setting, it can still be a favourable method in intermediate similarity settings.

\subsection{Interleaved Replay}

One set of methods for combating forgetting involves showing examples from previous tasks during training of later tasks. This can be done in a range of ways from explicitly storing data from previous tasks to training a generative model from which to sample data during later tasks~\cite{shin2017continual,draelos2017neurogenesis}. These methods are inspired by systems consolidation theories in neuroscience, e.g. hippocampal replay~\cite{kumaran2016learning}.

\begin{figure}[t!]
	\pgfplotsset{
		width=1.1\textwidth,
		height=0.8\textwidth,
		scaled x ticks=false,
		every tick label/.append style={font=\tiny},
		y label style={at={(axis description cs:-0.16, 0.5)}, rotate=0, anchor=south},
		x label style={at={(axis description cs:0.5, -0.45)}, rotate=0, anchor=south},
		xlabel={\scriptsize step, $s$},
		}
	\begin{subfigure}[b]{0.235\textwidth}
	\begin{tikzpicture}
        \fill[shadecolor2, opacity=0] (0, 0) rectangle (\textwidth, 0.75\textwidth);
        \node [anchor=north west] at (0, 0.75\textwidth) {\emph{(a)}};
        \node [anchor=north west] at (1.6, 0.75\textwidth) {\scriptsize $T=100$};
		\begin{axis}
			[
			at={(0.9cm, 0.75cm)},
			anchor=south west,
			xmin=0, xmax=12000000,
			ymin=-5, ymax=0,
			ylabel={\scriptsize $\log\epsilon^\dagger$}
		]
		\addplot graphics [xmin=0, xmax=12000000,ymin=-5,ymax=0] {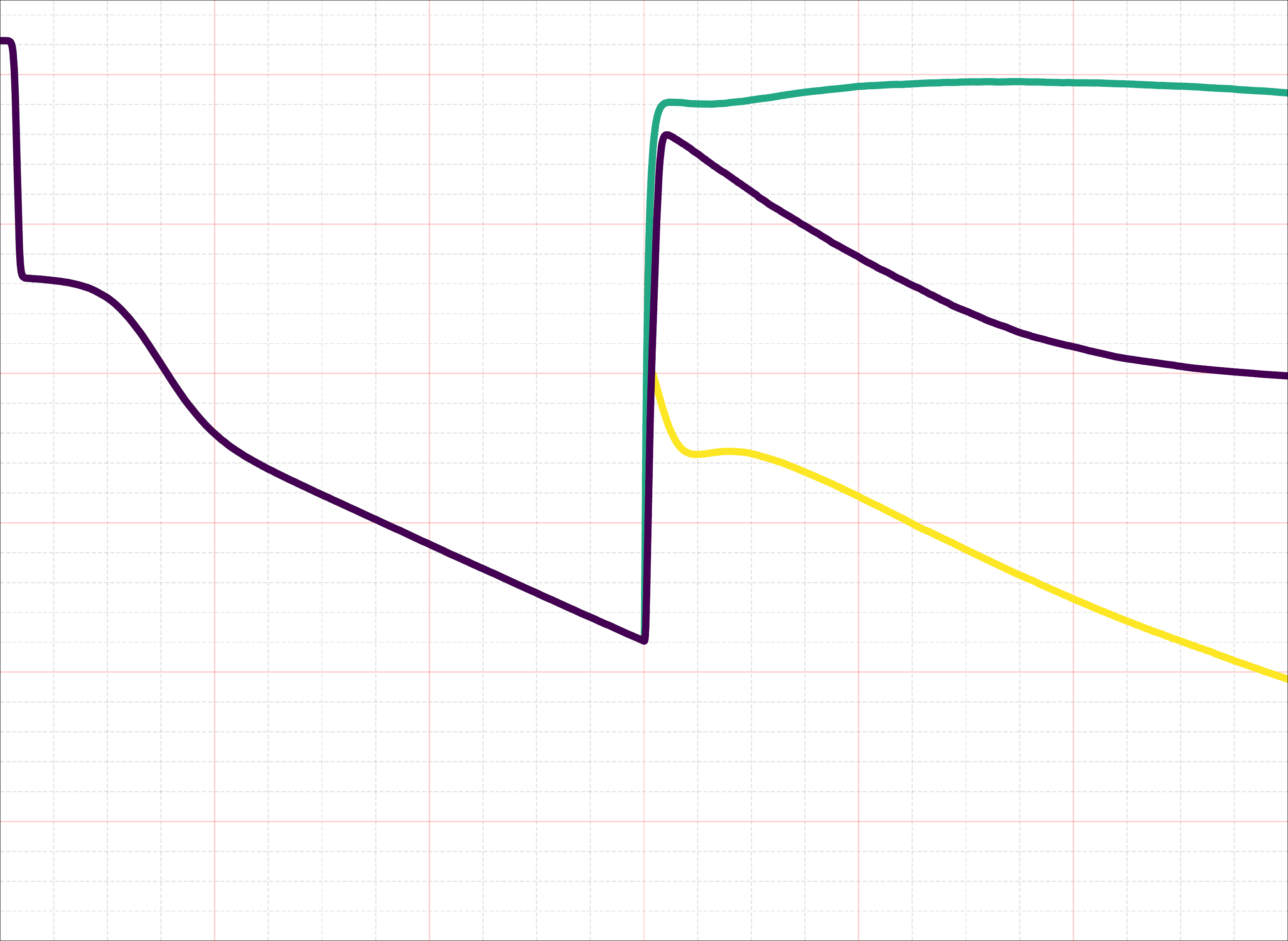};
		\end{axis}
	\end{tikzpicture}%
	\phantomcaption
	\label{fig:6a}
	\end{subfigure}
	\begin{subfigure}[b]{0.235\textwidth}
	\begin{tikzpicture}
        \fill[shadecolor2, opacity=0] (0, 0) rectangle (\textwidth, 0.75\textwidth);
        \node [anchor=north west] at (0, 0.75\textwidth) {\emph{(b)}};
        \node [anchor=north west] at (1.6, 0.75\textwidth) {\scriptsize $T=10$};
		\begin{axis}
			[
			at={(0.9cm, 0.75cm)},
			anchor=south west,
			xmin=0, xmax=12000000,
			ymin=-5, ymax=0,
			ylabel={\scriptsize $\log\epsilon^\dagger$}
		]
		\addplot graphics [xmin=0, xmax=12000000,ymin=-5,ymax=0] {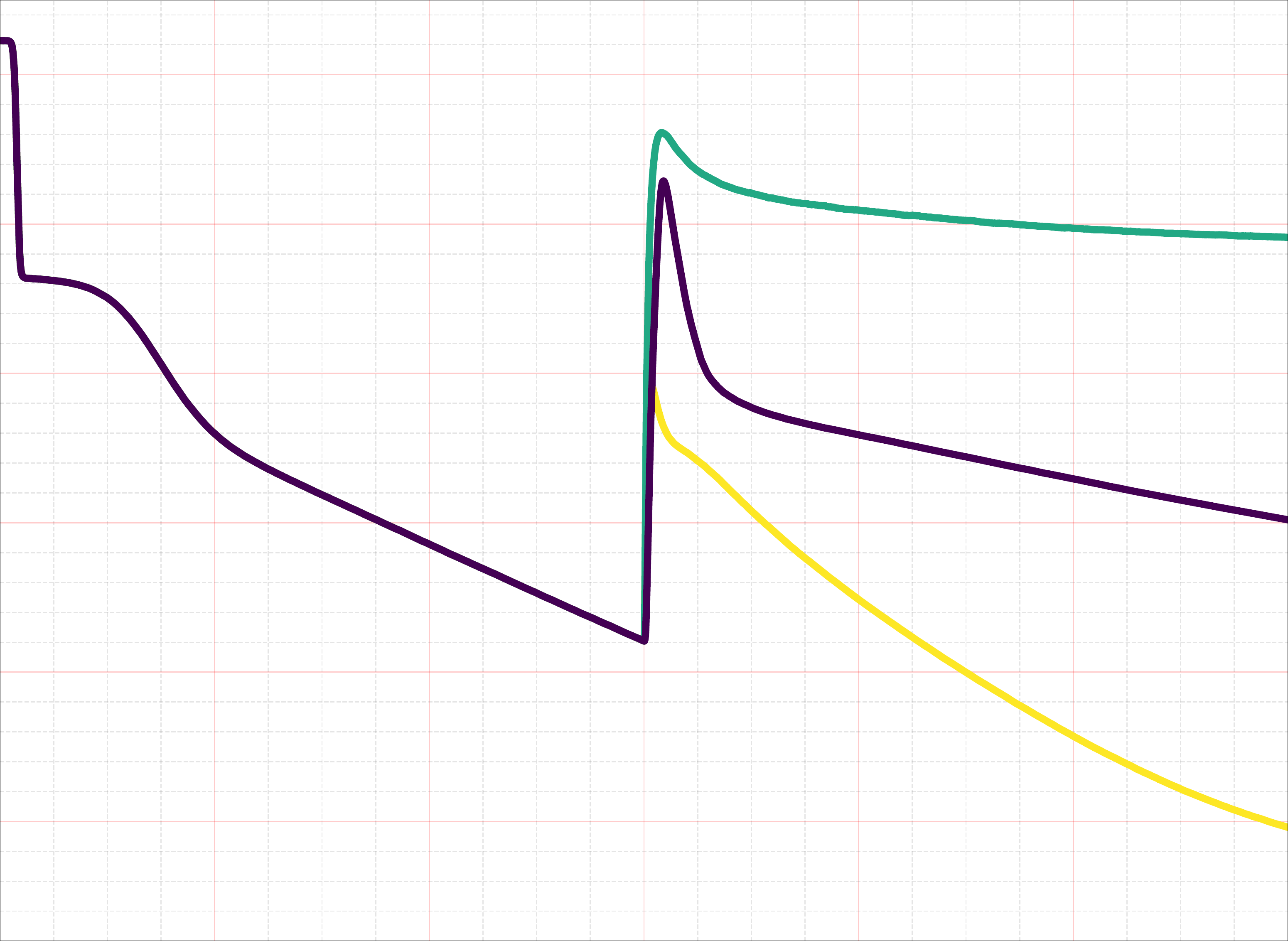};
		\end{axis}
	\end{tikzpicture}%
	\phantomcaption
	\label{fig:6b}
	\end{subfigure}
	\hspace{0em}
	\begin{subfigure}[b]{0.235\textwidth}
	\begin{tikzpicture}
        \fill[shadecolor2, opacity=0] (0, 0) rectangle (\textwidth, 0.75\textwidth);
        \node [anchor=north west] at (0, 0.75\textwidth) {\emph{(c)}};
        \node [anchor=north west] at (1.6, 0.75\textwidth) {\scriptsize $T=2$};
		\begin{axis}
			[
			at={(0.9cm, 0.75cm)},
			anchor=south west,
			xmin=0, xmax=12000000,
			ymin=-5, ymax=0,
			ylabel={\scriptsize $\log\epsilon^\dagger$}
		]
		\addplot graphics [xmin=0, xmax=12000000,ymin=-5,ymax=0] {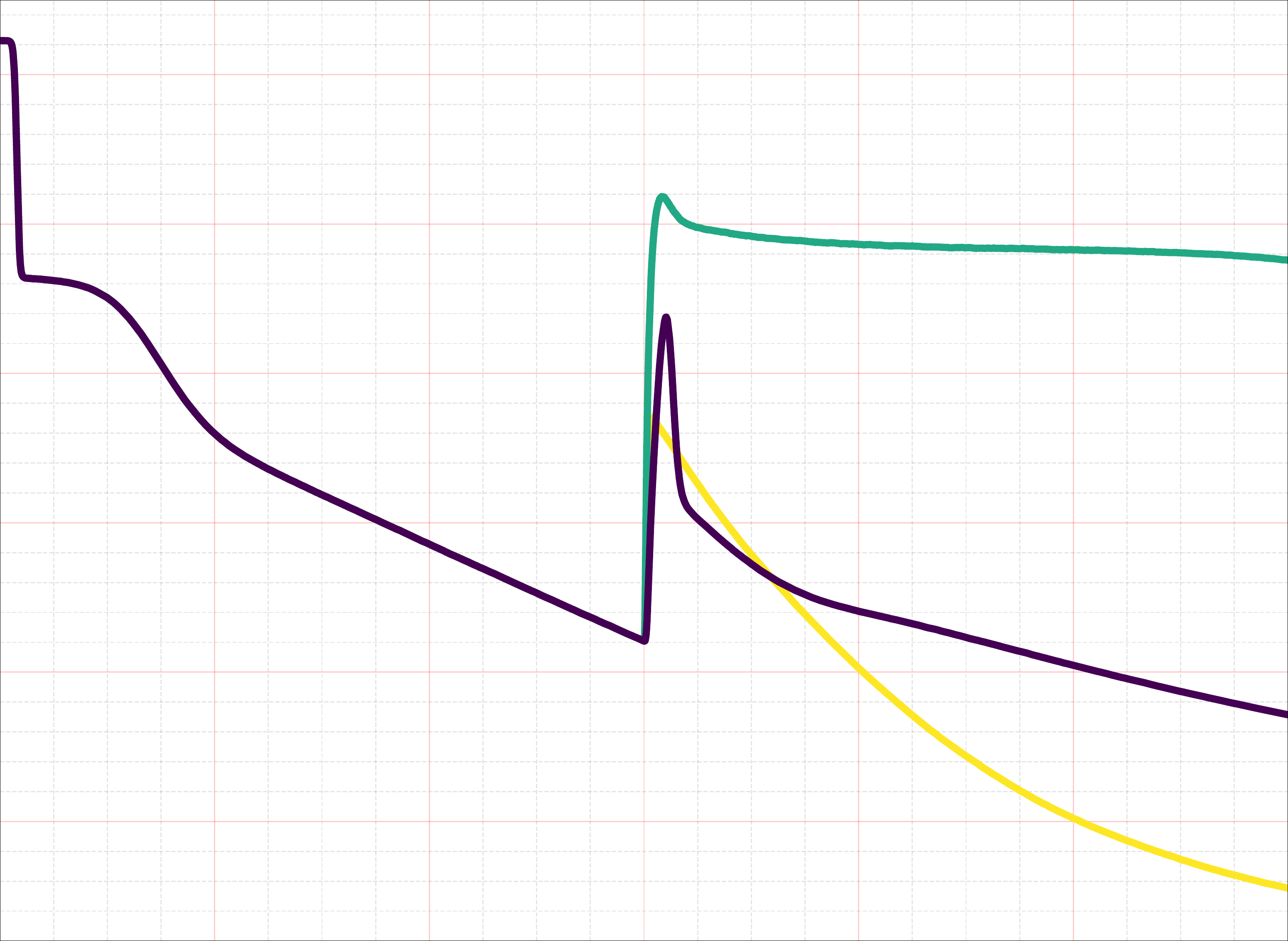};
		\end{axis}
	\end{tikzpicture}%
	\phantomcaption
	\label{fig:6c}
	\end{subfigure}
	\begin{subfigure}[b]{0.235\textwidth}
	\begin{tikzpicture}
        \fill[shadecolor2, opacity=0] (0, 0) rectangle (\textwidth, 0.75\textwidth);
        \node [anchor=north west] at (0, 0.75\textwidth) {\emph{(d)}};
        \node [anchor=north west] at (1.6, 0.75\textwidth) {\scriptsize $T=1$};
		\begin{axis}
			[
			at={(0.9cm, 0.75cm)},
			anchor=south west,
			xmin=0, xmax=12000000,
			ymin=-5, ymax=0,
			ylabel={\scriptsize $\log\epsilon^\dagger$}
		]
		\addplot graphics [xmin=0, xmax=12000000,ymin=-5,ymax=0] {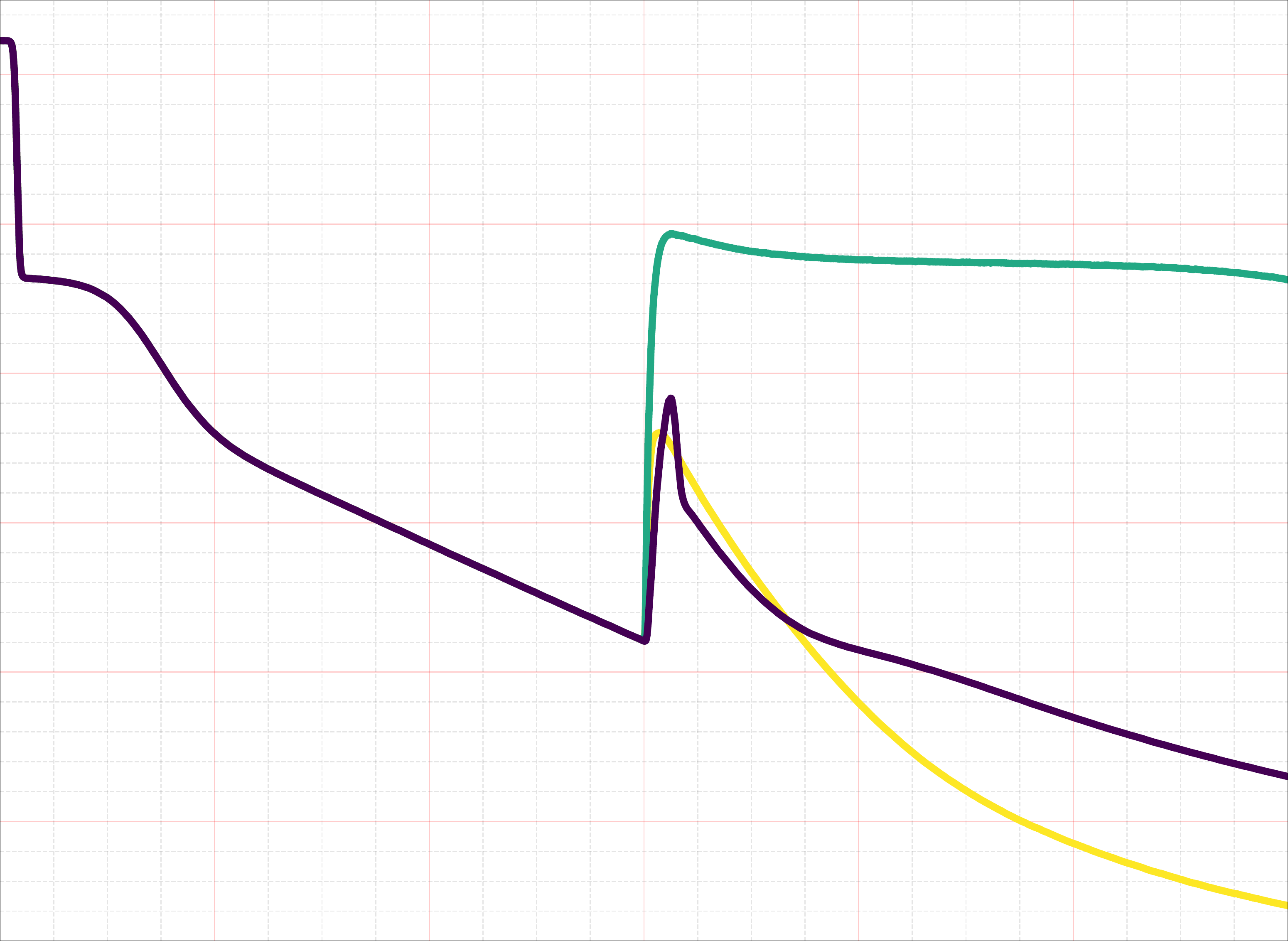};
		\end{axis}
	\end{tikzpicture}%
	\phantomcaption
	\label{fig:6d}
	\end{subfigure}
	\hspace{0em}
	\vspace{-2em}
	\caption[]{\textbf{Effect of replay on task similarity vs. forgetting} Although interleaving is superior to EWC in highly aligned or highly orthogonal regimes, it remains very poor in intermediate regimes even for fully interleaved training.\label{fig:interleave}}
 	\vspace{-2em}
\end{figure}

In the student-teacher framework, it is straightforward to implement this kind of algorithm since the teacher~\emph{is} the generative model; this means we can intermittently sample from previous teachers during training on later teachers. Here we focus in particular on interleaving a single example from the first teacher at different periods, $T$, during training of the second teacher. Other durations of interleaving and study of potential prioritisation schemes are left for future work. This sensitivity analysis is shown in~\autoref{fig:interleave}.

As the period of interleaving reduces, forgetting reduces and indeed the student begins to co-learn in the orthogonal and aligned regimes (this can be thought of as a kind of backward-transfer). Unlike for~\gls{ewc} where high $\lambda$ collapses the trajectories of the student onto one regardless of teacher-teacher similarity, even for the strongest interleaving ($T=1$) intermediate similarity remains the most difficult regime. 
Again this boils down to a difficult and ultimately costly trade-off between node re-use and node activation which interleaved training, unlike strong regularisation, cannot mitigate (see~\autoref{app: overlaps}). 

\begin{figure}[t!]
	\pgfplotsset{
		width=1.1\textwidth,
		height=\textwidth,
		scaled x ticks=false,
		every tick label/.append style={font=\tiny},
		y label style={at={(axis description cs:-0.16, 0.5)}, rotate=0, anchor=south},
		x label style={at={(axis description cs:0.5, -0.3)}, rotate=0, anchor=south},
		xlabel={\scriptsize step, $s$},
		}
	\begin{subfigure}[b]{0.235\textwidth}
	\begin{tikzpicture}
        \fill[shadecolor2, opacity=0] (0, 0) rectangle (\textwidth, 0.9\textwidth);
        \node [anchor=north west] at (0, 0.9\textwidth) {\emph{(a)}};
		\begin{axis}
			[
			at={(0.9cm, 0.95cm)},
			anchor=south west,
			xmin=0, xmax=20000000,
			xtick={6000000, 20000000},
			ymin=-7.5, ymax=-0.5,
			ylabel={\scriptsize $\log\epsilon^\dagger$}
		]
		\addplot graphics [xmin=0, xmax=20000000,ymin=-7.5,ymax=-0.5] {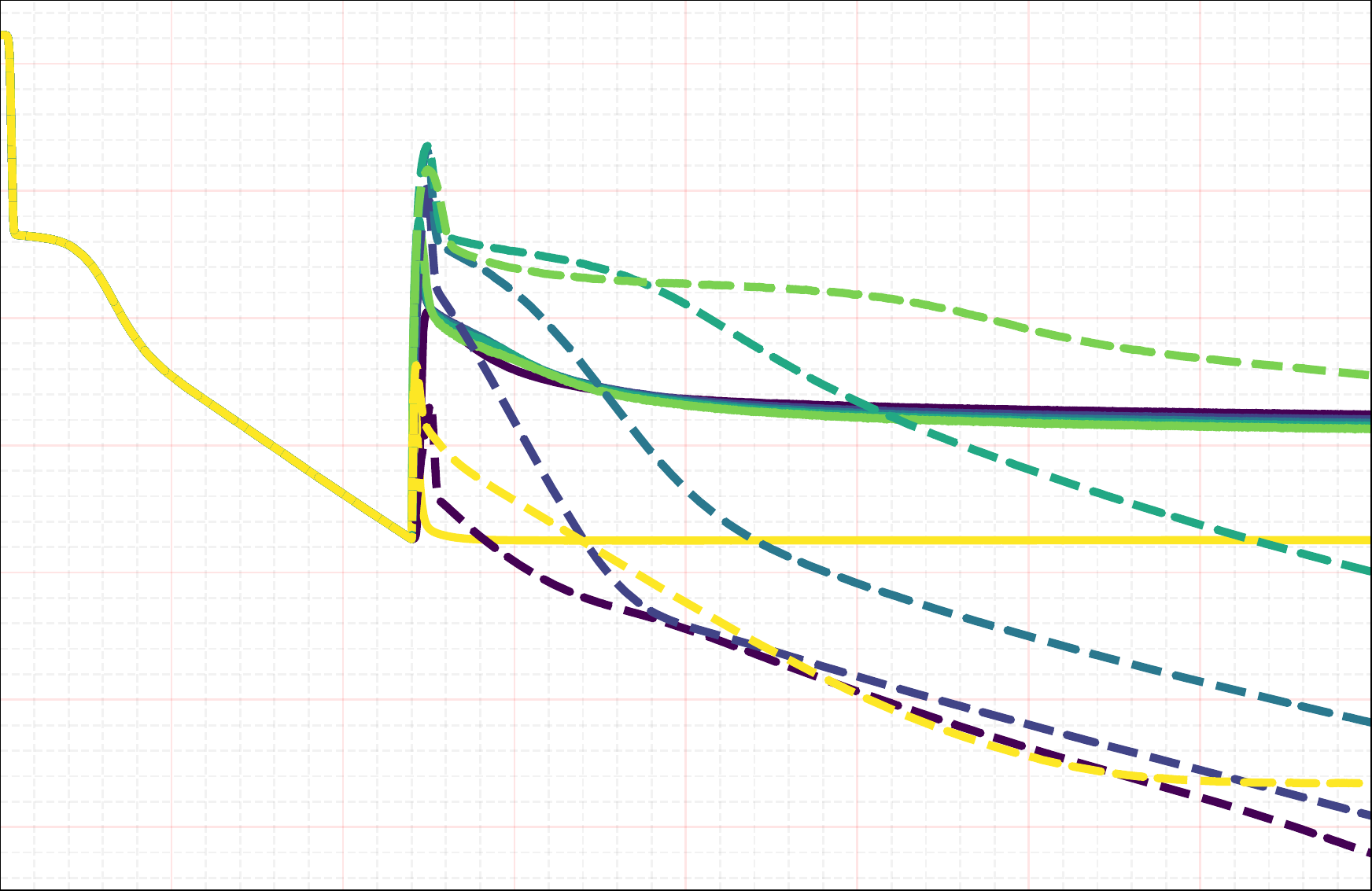};
		\end{axis}
	\end{tikzpicture}%
	\phantomcaption
	\label{fig:7a}
	\end{subfigure}
	\begin{subfigure}[b]{0.235\textwidth}
	\begin{tikzpicture}
        \fill[shadecolor2, opacity=0] (0, 0) rectangle (\textwidth, 0.9\textwidth);
        \node [anchor=north west] at (0, 0.9\textwidth) {\emph{(b)}};
		\begin{axis}
			[
			at={(0.9cm, 0.95cm)},
			anchor=south west,
			xmin=0, xmax=12000000,
			ymin=-6.8, ymax=-0.5,
			ylabel={\scriptsize $\log\epsilon^\dagger$}
		]
		\addplot graphics [xmin=0, xmax=12000000,ymin=-6.8,ymax=-0.5] {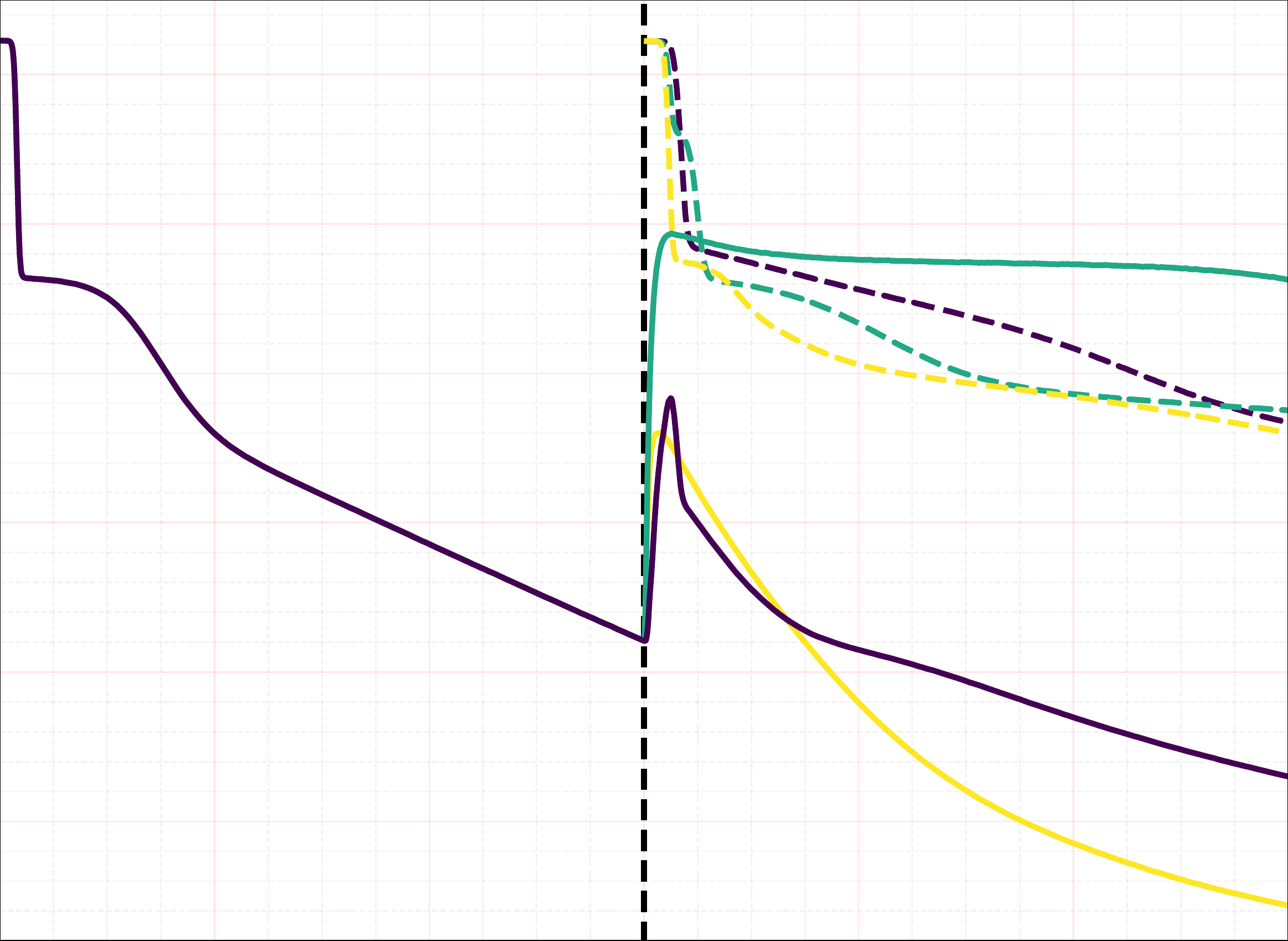};
		\end{axis}
	\end{tikzpicture}%
	\phantomcaption
	\label{fig:7b}
	\end{subfigure}
	\vspace{-1.5em}
	\caption[]{\textbf{EWC vs. interleaved replay}~\emph{(a)} Generalisation error on first teacher $\log\epsilon^\dagger$ vs step $s$ for a set of teacher-teacher similarities $V$. Dashed (solid) lines represent $T=1$ interleaved replay of the first teacher ($\lambda=1000$ EWC penalty) in the second phase of training. Although interleaving is far superior on the extremes of the spectrum, there is a range of task similarity where interleaved replay is worse than the trajectories of EWC.
	\textbf{Catastrophic slowing}~\emph{(b)} Generalisation error $\log\epsilon^\dagger$ vs. step $s$ for extremes of task similarity spectrum $V$. Solid lines represent interleaved replay from the task switch whereas dashed lines represent a re-initialisation followed by interleaved training from the task switch. If you allow access to the first teacher during the second phase of training, re-initialising at the task switch is better for forgetting (and transfer) in the intermediate task similarity regime.}
	\label{fig:comparison-catastrophic_slowing}
 	\vspace{-1em}
\end{figure}

It is informative to compare directly the generalisation error trajectories for vanilla~\gls{sgd}, strong interleaving, and strong~\gls{ewc} (see~\autoref{fig:7a}).
Although interleaving is far superior in the orthogonal and aligned regimes (effectively allowing backward transfer),~\gls{ewc} is better in the intermediate similarity case despite heavily stagnating on the first task due to the strong deviation penalty. Given interleaving is generally considered to be the gold standard~\cite{kumaran2016learning}, this may be unexpected. Is this a consequence of poor initialisation~\cite{NEURIPS2020_618491e2,gerace2021probing}, or is the intermediate similarity task structure inherently challenging for interleaved training? We investigated this question by comparing to the drastic strategy of completely re-initialising at the task boundary and performing interleaved training from a~\emph{tabula rasa} network. As we show in~\autoref{fig:7b}, there is a clear benefit in terms of forgetting in interleaving experiences starting from the solution to the first task when teachers are orthogonal or aligned. In the intermediate regime however re-initialising entirely is actually better! This cannot be explained away by a trade-off with superior performance on the second task as this is also better when re-initialising (see~\autoref{app: transfer_supp}). This suggests the presence of a~\emph{catastrophic slowing} effect in the intermediate regime where interleaving is not a viable combating method due to the tight balance between re-use and activation.
    
\section{Discussion}\label{sec: conc}

This work has introduced the \hpname~hypothesis, which shows how a trade-off between re-use and activation at the node level gives rise to a non-monotonic relationship between task similarity and forgetting such that intermediate task similarity is most damaging. 
The universality of this explanation remains to be fully established: In the teacher-student setup, the isotropy of the input distribution implies that the feature weights must pay attention to every part of the input distribution, and behavioural signatures at the node level capture those of interest at the weight level. In the data-mixing framework, where this data isotropy is broken and we can expect the learned representation in the feature weights to be sparser, we still see evidence for a trade-off between re-use and activation at the node level even if this is a coarsening of more complicated interactions at the weight level. In other settings, this trade-off may play out at a super-node level in clusters or sub-networks. We leave the study of these trade-offs over different components of the network for future work. 
We use the insights gained from~\hpname~to rethink two popular combating methods for catastrophic forgetting (\gls{ewc} and interleaved replay), and identify among other properties of these methods an effect we term~\emph{catastrophic slowing} where interleaving experiences in an intermediate task similarity regime is worse than re-initialising the network, from both a transfer and forgetting perspective. Moving forward, we hope that~\hpname~for catastrophic forgetting can help elucidate related phenomena in continual learning and cognate paradigms.

\footnotesize
Code: \href{github.com/seblee97/student\_teacher\_catastrophic}{github.com/seblee97/student\_teacher\_catastrophic}
\normalsize

\section{Acknowledgements}

SL is supported by an EPSRC DTP Studentship. CC would like to acknowledge support from BBSRC (BB/N013956/1, BB/N019008/1), Wellcome Trust (200790/Z/16/Z), the Simons Foundation (564408) and EPSRC(EP/R035806/1). This work was further supported by the Sainsbury Wellcome Centre Core Grant from Wellcome (219627/Z/19/Z) and the Gatsby Charitable Foundation (GAT3755) (SL/SSM/AS). AS is a CIFAR Azrieli Global Scholar in the Learning in Machines \& Brains programme.

\nocite{langley00}

\bibliography{main}
\bibliographystyle{icml2022}

\newpage
\appendix

\onecolumn

\title{Appendix}\date{}
\maketitle
\section{Experiment Details}\label{app: exp_details}

Unless mentioned otherwise in the main text, the following parameters were used in all teacher-student runs:
\begin{itemize}
    \item Input dimension = 1000;
    \item Test set size = 50,000;
    \item \gls{sgd} optimiser;
    \item Mean squared error loss;
    \item Teacher weight initialisation: normal distribution with variance 1;
    \item Student weight initialisation: normal distribution with variance 0.001;
    \item Student hidden dimension: 4; 
    \item Teacher hidden dimension: 2;
    \item Learning rate: 0.1;
    \item Nonlinearity = scaled error function.
\end{itemize}

For the data mixing framework, we use the Fashion~\gls{mnist} dataset, with the following parameters:

\begin{itemize}
    \item $\mathcal{D}_1$: class 0, class 5;
    \item $\tilde{\mathcal{D}}_2$: class 2, class 7;
    \item \gls{sgd} optimiser;
    \item Mean squared error loss;
    \item Batch size = 1;
    \item Input dimension = 1024;
    \item Hidden dimension = 8;
    \item Nonlinearity = sigmoid;
    \item Learning rate: 0.001.
\end{itemize}

We grayscale the data and apply an early stopping regime such that the weights used for the network in the second task are those with the lowest test error obtained during the first phase of training. This is to avoid any additional effects from overfitting.

Code for the experiments can be found at: \href{github.com/seblee97/student\_teacher\_catastrophic}{github.com/seblee97/student\_teacher\_catastrophic}
\section{Specialisation Assumption}\label{app: specialisation}

Implicit in some of the discussion around~\hpname~is that there is specialisation in the network during the first task, and that there is additional capacity in the network to learn later tasks. We show in the ODE solutions for the teacher-student setup and empirically in the data mixing framework that this is the case (even in a relatively small network). We do not show the plots here, but we also confirmed this assumption to be robust in the teacher-student framework over every aspect of the problem we looked at (activation function, level of over-parameterisation, size of hidden dimension, noise and noiseless teachers). In settings where this assumption does not hold at all,~\hpname~will likely not hold in the way it is currently stated. We leave investigation of this regime for future work.

\section{ODE Formulation}\label{app: ode}

Below we outline the ODE analysis that is used in the teacher-student components of this paper. Since our setup is very similar to that of~\citet{lee2021continual}, the derivations below are also the same. We reproduce it here to facilitate a self-contained paper. 

\subsection{Order Parameters}\label{app: order_params}

The full set of order parameters for the two-teacher student-teacher networks in the large input limit is given by:

\begin{align}
	\text{\footnotesize Student-Student Overlap, }\mathbf{Q}:  q_{kl} & \equiv \langle\lambda_k\lambda_l\rangle %
	= \frac{1}{N}\wvecb_k\wvecb_l;\\
	\text{\footnotesize Teacher}^\dagger\text{\footnotesize -Teacher}^\dagger\text{\footnotesize Overlap, }\mathbf{T}: t_{nm} & \equiv \langle\rho_m\rho_n\rangle %
	= \frac{1}{N}\wvecb^\dagger_m\wvecb^\dagger_n  ;\\
	\text{\footnotesize Student-Teacher}^\dagger\text{\footnotesize Overlap, }\mathbf{R}:  r_{km} & \equiv \langle\lambda_k\rho_m\rangle %
	= \frac{1}{N}\wvecb_k\wvecb^\dagger_m;\\
	\text{\footnotesize Teacher}^\ddag\text{\footnotesize -Teacher}^\ddag\text{\footnotesize Overlap, }\mathbf{S}: s_{pq} & \equiv \langle\eta_p\eta_q\rangle %
	= \frac{1}{N}\wvecb^\ddag_p\wvecb^\ddag_q ;\\
	\text{\footnotesize Student-Teacher}^\ddag\text{\footnotesize Overlap, }\mathbf{U}:  u_{kp}&\equiv \langle\lambda_k\eta_p\rangle %
	= \frac{1}{N}\wvecb_k\wvecb^\ddag_p ; \\
	\text{\footnotesize Teacher}^\dagger\text{\footnotesize -Teacher}^\ddag\text{\footnotesize Overlap, }\mathbf{V}: v_{mp} & \equiv\langle\rho_m\eta_p\rangle%
	= \frac{1}{N}\wvecb^\dagger_m\wvecb^\ddag_p;
\end{align}
along with the head weights, $\vb^\dagger$, $\vb^\ddag$, $\hb^\dagger$, $\hb^\ddag$.

Throughout, we denote any quantities associated with the first task with $\dagger$, any quantity associated with the second task with $\ddag$. In any quantity or equation that generally holds for $\dagger$ or $\ddag$, we represent this by marking it with $*$.

\subsection{Generalisation Error in terms of Order Parameters}

Our aim is to formulate the generalisation error in terms of the macroscopic order parameters. 

The~\gls{sgd} update equations in the two-layer teacher-student setup are given by:
\begin{subequations}
    \label{eq:sgd}
    \begin{align}
    	\wvecb_k^{\mu+1} &= \wvecb_k^{\mu} - \frac{\alpha_\wb}{\sqrt{D}}v_k^{*\mu}g'(\lambda_k^\mu)\Delta^{*\mu} \xb^\mu \label{eq: wupdate} \\
	    h_k^{*\mu+1} &= h_k^{*\mu} -
                       \frac{\alpha_\hb}{D}g(\lambda_k^\mu)\Delta^{*\mu}, \label{eq:
                       vupdate}
    \end{align}
\end{subequations}
where $\alpha_\wb$ is the learning rate for the feature weights, $\alpha_\hb$ is the learning rate for the head weights, and
\begin{align}
	\Delta^{\dagger\mu} &\equiv \sum_k h_k^{\dagger\mu}g(\lambda_k^\mu) - \sum_m v_m^\dagger g(\rho_m^\mu);\\
	\Delta^{\ddag\mu} &\equiv \sum_k h_k^{\ddag\mu}g(\lambda_k^\mu) - \sum_p v_p^\ddag g(\eta_p^\mu).
\end{align}
We have also introduced the \emph{local fields} 
\begin{equation}
    \rho_m \equiv \frac{\wvecb_m\xb}{\sqrt{D}}, \qquad \eta_p \equiv \frac{\wvecb_p\xb}{\sqrt{D}}, \qquad \lambda_k \equiv \frac{\wvecb_k\xb}{\sqrt{D}}
\end{equation}
of the $m^{\text{th}}$ teacher $\dagger$ unit, $n^{\text{th}}$ teacher $\ddag$ unit, and $k^{\text{th}}$ student unit, respectively. In general, 
indices $i, j, k, l$ are used for hidden units of the student; $m, n$ for hidden units of $\dagger$; and $p, q$ for hidden units of $\ddag$.

Let us begin 
by multiplying out~\autoref{eq:eg},
\begin{align}
	\epsilon_g^\dagger = \frac{1}{2}\left\langle\left[\sum_{i, k}h_i^\dagger h_k^\dagger g(\lambda_i)g(\lambda_k)
	+\sum_{m, n}v_m^\dagger v_n^\dagger g(\rho_m)g(\rho_n) - 2\sum_{i,n}h_i^\dagger v_n^\dagger g(\lambda_i)g(\rho_n)\right]\right\rangle.
\end{align}

These generalisation errors involve averages of local fields, which can be computed as integrals over a joint 
multivariate Gaussian probability distribution, all of the form
\begin{equation}
    \mathcal{P}(\beta, \gamma) = \frac{1}{\sqrt{(2\pi)^{F + H}|\tilde{\mathbf{C}}|}}\exp{\left\{-\frac{1}{2}(\beta, \gamma)^T\tilde{\mathbf{C}}^{-1}(\beta,\gamma)\right\}},
\end{equation}
where $\beta$ and $\gamma$ are local fields with number of units $F$ and $H$ respectively, and $\tilde{\mathbf{C}}$ is 
a covariance matrix suitably projected down from
\[
\mathbf{C} = 
\begin{pmatrix} 
	\mathbf{Q} & \mathbf{R} & \mathbf{U}\\
	\mathbf{R}^T & \mathbf{T} & \mathbf{V}\\
	\mathbf{U}^T & \mathbf{V}^T & \mathbf{S}
\end{pmatrix}.
\]
We define
\begin{equation}
	I_2(f, h) \equiv \langle g(\beta)g(\gamma)\rangle, \label{eq: i2}
\end{equation}
where $f, h$ are the indices corresponding to the units of the local fields $\beta$ and $\gamma$. 
This allows us to write the generalisation errors as 
\begin{align}
	\epsilon_g^\dagger &= \frac{1}{2}\sum_{i,k}h^\dagger_ih^\dagger_kI_2(i,k) + \frac{1}{2}\sum_{n,m}v^\dagger_nv^\dagger_mI_2(n,m) - \sum_{i,n}h^\dagger_iv^\dagger_nI_2(i,n)\label{eq: generror1I}\\
	\epsilon_g^\ddag &= \frac{1}{2}\sum_{i,k}h^\ddag_ih^\ddag_kI_2(i,k) + \frac{1}{2}\sum_{p,q}v^\ddag_pv^\ddag_qI_2(p,q) - \sum_{i,p}h^\ddag_iv^\ddag_pI_2(i,p).\label{eq: generror2I}
\end{align}

\subsubsection{Sigmoidal Activation}
For the scaled error activation function, $g(x) = \text{erf}(x/\sqrt{2})$, there is an analytic 
expression for the $I_2$ integral purely in terms of the order parameters~\cite{saad1995exact}:
\begin{equation}
	I_2(i,k) = \frac{1}{\pi}\arcsin{\frac{q_{ik}}{\sqrt{(1 + q_{ii})(1 + q_{kk})}}}.
\end{equation}
In turn, we can similarly write the generalisation errors in terms of the order parameters only:
\begin{multline}
	\epsilon_g^\dagger = \frac{1}{\pi}\sum_{i,k} h^\dagger_ih^\dagger_k\arcsin{\frac{q_{ik}}{\sqrt{(1 + q_{ii})(1 + q_{kk})}}} + \frac{1}{\pi}\sum_{n,m} v^\dagger_nv^\dagger_m\arcsin{\frac{t_{nm}}{\sqrt{(1 + t_{nn})(1 + t_{mm})}}}\\
	+ \frac{2}{\pi}\sum_{i,n} h^\dagger_iv^\dagger_n\arcsin{\frac{r_{in}}{\sqrt{(1 + q_{ii})(1 + t_{nn})}}}
\end{multline}
\begin{multline}
	\epsilon_g^\ddag = \frac{1}{\pi}\sum_{i,k} h^\ddag_ih^\ddag_k\arcsin{\frac{q_{ik}}{\sqrt{(1 + q_{ii})(1 + q_{kk})}}} + \frac{1}{\pi}\sum_{p,q} v^\ddag_pv^\ddag_q\arcsin{\frac{s_{pq}}{\sqrt{(1 + s_{pp})(1 + s_{qq})}}}\\
	+ \frac{2}{\pi}\sum_{i,p} h^\ddag_iv^\ddag_p\arcsin{\frac{u_{ip}}{\sqrt{(1 + q_{ii})(1 + s_{pp})}}}.
\end{multline}

\subsection{Order Parameter Evolution (Training on $\dagger$)}

Having arrived at expressions for the generalisation error of both teachers in terms of the order parameters, we want to determine equations 
of motion for these order parameters from the weight update equations (\autoref{eq: wupdate} \& \autoref{eq: vupdate}). Trivially, the order 
parameters associated with the two teachers, $\mathbf{T}$ and $\mathbf{S}$ are constant over time, as are the head weights of the teachers, $\vb^\dagger, \vb^\ddag$. 
When training on $\dagger$, the student head weights corresponding to $\ddag$ are also stationary; it remains for us to find equations 
of motion for $\mathbf{R}, \mathbf{Q}, \mathbf{U}$ and $\mathbf{h}^\dagger$, which we derive below. The equivalent derivations
when training on teacher $\ddag$ can be made by using the update in~\autoref{eq: vupdate} instead.

\subsubsection{\gls{ode} for $\mathbf{R}$}

Consider the inner product of \autoref{eq: wupdate} (in the case of * = $\dagger$) with $\wvecb_n^\dagger$:
\begin{align}
	\wvecb_k^{\mu+1}\wvecb_n^\dagger - \wvecb_k^\mu \wvecb_n^\dagger &= -\frac{\alpha_\wb}{\sqrt{D}}h_k^{\dagger\mu} g'(\lambda_k^\mu)\Delta^{\dagger\mu} \xb^\mu \wvecb_n^\dagger\\
	&= -\alpha_\wb h_k^{\dagger\mu} g'(\lambda_k^\mu)\Delta^{\dagger\mu} \rho_n^\mu\\ 
	r_{kn}^{\mu+1} - r_{kn}^\mu&= -\frac{\alpha_\wb}{D} h_k^{\dagger\mu} g'(\lambda_k^\mu)\Delta^{\dagger\mu} \rho_n^\mu
\end{align}
If we let $\tau \equiv \mu/D$ and take the thermodynamic limit of $D\to\infty$, the time parameter becomes continuous and we can write:
\begin{equation}
	\frac{dr_{in}}{d\tau} = -\alpha_\wb h_i^\dagger \langle g'(\lambda_i)\Delta^\dagger \rho_n\rangle,
\end{equation}
where we have re-indexed $k\to i$.

\subsubsection{\gls{ode} for $\mathbf{Q}$}

Consider squaring \autoref{eq: wupdate} (here we can simply use * to denote training on either teacher).
\begin{align}
	\wvecb_k^{\mu+1}\wvecb_i^{\mu+1} - \wvecb_k^\mu \wvecb_i^\mu &=-\frac{\alpha_\wb}{\sqrt{D}}h_i^{*\mu} g'(\lambda_i^\mu)\Delta^{*\mu} \xb^\mu \wvecb_k^\mu -\frac{\alpha_\wb}{\sqrt{D}}h_k^{*\mu} g'(\lambda_k^\mu)\Delta^{*\mu} \xb^\mu \wvecb_i^\mu \nonumber\\ 
	&\quad \quad + \frac{\alpha_\wb^2}{D} h_i^{*\mu} g'(\lambda_i^\mu)h_k^{*\mu} g'(\lambda_k^\mu)(\Delta^{*\mu} \xb^\mu)^2 \\
	&= - \alpha_\wb h_i^{*\mu} g'(\lambda_i^\mu)\Delta^{*\mu}\lambda_k^\mu -\alpha_\wb h_k^{*\mu} g'(\lambda_k^\mu)\Delta^{*\mu} \lambda_i^\mu \nonumber\\
	&\quad \quad+\frac{\alpha_\wb^2}{D} h_i^{*\mu} g'(\lambda_i^\mu)h_k^{*\mu} g'(\lambda_k^\mu)(\Delta^{*\mu}\xb^\mu)^2 \\
	q^{\mu+1}_{ki} - q^{\mu}_{ki}&= -\frac{\alpha_\wb}{D} h_i^{*\mu} g'(\lambda_i^\mu)\Delta^{*\mu}\lambda_k^\mu - \frac{\alpha_\wb}{D} h_k^{*\mu} g'(\lambda_k^\mu)\Delta^{*\mu} \lambda_i^\mu\nonumber\\
	&\quad \quad + \frac{\alpha_\wb^2}{D^2} h_i^{*\mu} g'(\lambda_i^\mu)h_k^{*\mu} g'(\lambda_k^\mu)(\Delta^{*\mu}\xb^\mu)^2. 
\end{align}
Performing the same reparameterisation of $\mu$ and the same thermodynamic limit, we get:
\begin{align}
	\frac{dq_{ik}}{d\tau} = -\alpha_\wb h_i^* \langle g'(\lambda_i)\Delta^* \lambda_k\rangle - \alpha_\wb h_k^* \langle g'(\lambda_k)\Delta^* \lambda_i\rangle +\alpha_\wb^2 h_i^*h_k^*\langle g'(\lambda_i)g'(\lambda_k)\Delta^{*2}\rangle.
\end{align}
Note: in the limit, $(\xb^\mu)^2 \to D$ since individual samples are taken from a unit normal. Hence the $1/D$ limit 
remains the same decay rate for each term.

\subsubsection{\gls{ode} for $\mathbf{U}$}

Consider the inner product of \autoref{eq: wupdate} (in the case of * = $\dagger$) with $\wvecb_p^\ddag$:
\begin{align}
	\wvecb_k^{\mu+1}\wvecb_p^\ddag - \wvecb_k^\mu \wvecb_p^\ddag &= -\frac{\alpha_\wb}{\sqrt{D}}h_k^{\dagger\mu} g'(\lambda_k^\mu)\Delta^{\dagger\mu} \xb^\mu \wvecb_p^\ddag\\
	&= -\alpha_\wb h_k^{\dagger\mu} g'(\lambda_k^\mu)\Delta^{\dagger\mu} \eta_p^\mu\\
	u_{kp}^{\mu+1} - u_{kp}^\mu&= -\frac{\alpha_\wb}{D} h_k^{\dagger\mu} g'(\lambda_k^\mu)\Delta^{\dagger\mu} \eta_p^\mu.
\end{align}
If we let $\tau \equiv \mu/D$ and take the thermodynamic limit of $D\to\infty$:
\begin{equation}
	\frac{du_{ip}}{d\tau} = -\alpha_\wb h_i^* \langle g'(\lambda_i)\Delta^* \eta_p\rangle. \label{eq: uip}
\end{equation}

\subsubsection{\gls{ode} for $\hb^*$}

Here, we simply take the thermodynamic limit of \autoref{eq: vupdate} (for * = $\dagger$):
\begin{equation}
	\frac{dh_i^\dagger}{d\tau} = -\alpha_h\langle\Delta^\dagger g(\lambda_i)\rangle
\end{equation}

\section{Explicit Formulation}

We can go one step further and write the right hand sides of the \glspl{ode} in terms of more concise integrals. 
Recall that for no noise
\begin{equation}
	\Delta^{\dagger\mu} \equiv \sum_k h_k^{\dagger\mu}g(\lambda_k^\mu) - \sum_m v_m^\dagger g(\rho_m^\mu).
\end{equation}
Substituting this term into the \glspl{ode} above gives us the expanded versions below:
\begin{align}
	\frac{dr_{in}}{d\tau} &= -\alpha_\wb h_i^\dagger \left\langle g'(\lambda_i)\left[\sum_k h_k^\dagger g(\lambda_k) - \sum_m v_m^\dagger g(\rho_m)\right] \rho_n\right\rangle;\\
	\frac{dq_{ik}}{d\tau} &= -\alpha_\wb h_i^\dagger \left\langle g'(\lambda_i)\left[\sum_j h_j^\dagger g(\lambda_j) - \sum_m v_m^\dagger g(\rho_m)\right]\lambda_k\right\rangle \nonumber \\
	&\quad \quad -\alpha_\wb h_k^\dagger \left\langle g'(\lambda_k)\left[\sum_j h_j^\dagger g(\lambda_j) - \sum_m v_m^\dagger g(\rho_m)\right]\lambda_i\right\rangle \nonumber \\
	&\quad \quad \quad+ \alpha_\wb^2 h_i^\dagger h_k^\dagger \left\langle g'(\lambda_i)g'(\lambda_k)\left[\sum_j h_j^\dagger g(\lambda_j) - \sum_m v_m^\dagger g(\rho_m)\right]^2\right\rangle;\\
	\frac{du_{ip}}{d\tau} &= -\alpha_\wb h_i^\dagger \left\langle g'(\lambda_i)\left[\sum_k h_k^\dagger g(\lambda_k) - \sum_m v_m^\dagger g(\rho_m)\right]\eta_p\right\rangle;\\
	\frac{dh_i^\dagger}{d\tau} &= -\alpha_\hb\left\langle\left[\sum_k h_k^\dagger g(\lambda_k) - \sum_m v_m^\dagger g(\rho_m)\right]g(\lambda_i)\right\rangle.         
\end{align}
Similarly to the $I_2$ integral defined in~\autoref{eq: i2}, we further define:
\begin{align}
	&I_3(d, f, h) = \langle g'(\zeta)\beta g(\gamma)\rangle,\\
	&I_4(d, e, f, h) = \langle g'(\zeta)g'(\iota)g(\beta)g(\gamma)\rangle;
\end{align}
where $\zeta, \iota$ are local fields of the student with indices $d, e$; and $\beta, \gamma$ can be local 
fields of either student or teacher with indices $f, h$.
Substituting these definitions into the expanded \gls{ode} formulations gives:
\begin{align}
	\frac{dr_{in}}{d\tau} &= \alpha_\wb h_i^\dagger\left[\sum_m^Mv_m^*I_3(i,n,m) - \sum_k^Kh_k^\dagger I_3(i,n,k)\right]; \label{eq: oder}\\
	\frac{dq_{ik}}{d\tau} &= \alpha_\wb h_i^\dagger \left[\sum_m^Mv^\dagger_mI_3(i,k,m) - \sum_j^Kh^\dagger_jI_3(i,k,j)\right] \nonumber\\
	& \quad + \alpha_\wb h_k^\dagger \left[\sum_m^Mv^\dagger_mI_3(k,i,m) - \sum_j^Kh^\dagger_jI_3(k,i,j)\right] \nonumber\\
	& \quad \quad +\alpha_\wb^2 h_i^\dagger h_k^\dagger\left[\sum_{j,l}^Kh^\dagger_jh^\dagger_lI_4(i,k,j,l) + \sum_{m,n}^Mv^\dagger_mv^\dagger_nI_4(i,k,m,n)\right. \nonumber \\
	& \quad \quad \quad \quad \quad \quad \left.- 2\sum_j^K\sum_m^Mv^\dagger_mh^\dagger_jI_4(i,k,j,m)\right]; \label{eq: odeq}\\
	\frac{du_{ip}}{d\tau} &= \alpha_\wb h_i^\dagger \left[\sum_m^Mv^\dagger_mI_3(i,p,m) - \sum_k^Kh^\dagger_kI_3(i,p,k)\right]; \label{eq: uipexp}\\
	\frac{dh_i^\dagger}{d\tau} &= \alpha_\hb\left[\sum_m^Mv^\dagger_mI_2(m,i) - \sum_k^Kh^\dagger_kI_2(k,i)\right]. \label{eq: hstar_exp}
\end{align}
This completes the picture for the dynamics of the generalisation error. It can be expressed purely 
in terms of the head weights and the $I$ integrals. For the case of the scaled error function we can 
evaluate the $I_2, I_3$, and $I_4$ analytically meaning we have an exact formulation of the generalisation 
error dynamics of the student with respect to both teachers in the thermodynamic limit. Further details 
on the integrals can be found in~\autoref{app: gaussian_integrals}. The next chapter introduces the 
experimental framework that compliments the theoretical formalism presented above.

\subsection{Gaussian Integrals under Scaled Error Function}\label{app: gaussian_integrals}

In the derivations above, we introduce a set of integrals over multivariate Gaussian distributions,
labelled $I_2$, $I_3$ and $I_4$. They are defined as:

\begin{align}
    I_2(f, h) &\equiv \langle g(\beta)g(\gamma)\rangle,\\
	I_3(d, f, h) &\equiv \langle g'(\zeta)\beta g(\gamma)\rangle,\\
	I_4(d, e, f, h) &\equiv \langle g'(\zeta)g'(\iota)g(\beta)g(\gamma)\rangle;
\end{align}
where $\zeta, \iota$ are local fields of the student with indices $d, e$; and $\beta, \gamma$ can be local 
fields of either student or teacher with indices $f, h$; and $g$ is the activation function. 

These integrals do not have closed form solutions for the ReLU activation. For the scaled error function however,
they can all be solved analytically. They are given by:

\begin{align}
    I_2 &= \frac{1}{\pi}\arcsin{\frac{c_{12}}{\sqrt{(1 + c_{11})(1 + c_{22})}}};\\
    I_3 &= \frac{2c_{23}(1 + c_{11}) - 2c_{12}c_{13}}{\sqrt{\Lambda_3}(1 + c_{11})};\\
    I_4 &= \frac{4}{\pi^2\sqrt{\Lambda_4}}\arcsin{\frac{\Lambda_0}{\sqrt{\Lambda_1\Lambda_2}}};
\end{align}
where 
\begin{align}
    \Lambda_0 &= \Lambda_4c_{34} - c_{23}c_{24}(1 + c_{11}) - c_{13}c_{14}(1 + c_{22}) + c_{12}c_{13}c_{24} + c_{12}c_{14}c_{23};\\
    \Lambda_1 &= \Lambda_4(1 + c_{33}) - c_{23}^2(1 + c_{11}) - c_{13}^2(1 + c_{22}) + 2c_{12}c_{13}c_{23};\\
    \Lambda_2 &= \Lambda_4(1 + c_{44}) - c_{24}^2(1 + c_{11}) - c_{14}^2(1 + c_{22}) + 2c_{12}c_{14}c_{24};\\
    \Lambda_3 &= (1 + c_{11})(1 + c_{33}) - c_{13}^2;\\
\end{align}
and where $c$ is the relevant projected down covariance matrix.
\section{Overlap Generation}\label{app: overlap_generation}

Throughout this work we tune the similarity of tasks in the teacher-student setup by manipulating the feature weights of the second teacher in relation to the first. Here we outline the procedure used. Once again this is largely similar to the procedure used in~\cite{lee2021continual}.

In the main text we mention that we abbreviate the matrix elements $v_{mp}$ to $V$. The motivation for this is that in the single hidden unit case the input-hidden weights are vectors and their dot products are simply scalars. When we move to the multi-hidden unit setting (see below), there is no single similarity measure from the overlap matrix but we construct our similarity based on a scalar interpolation between two random matrices, so continue to denote similarity simply by $V$.

For teachers with a single hidden unit we simply need a procedure to generate two $D$-dimensional vectors (where $D$ is the input dimension),
$\mathbf{v}_1$, $\mathbf{v}_2$, with an angle $\theta$ between them such that:
\begin{equation}
    \mathbf{v}_1\cdot \mathbf{v}_2 = \theta.
\end{equation}
Fortunately there is a standard algorithm for this. First we define two vectors
\[
\tilde{\mathbf{v}}_1 = 
\begin{pmatrix} 
	0\\
	1
\end{pmatrix};
\quad
\tilde{\mathbf{v}}_2 = 
\begin{pmatrix} 
	\sin{\theta}\\
	\cos{\theta}
\end{pmatrix}.
\]
Second, we generate an $D\times D$ orthogonal matrix, $R$. There is a standard sicpy implementation for this
based on QR decomposition of a random Gaussian 
matrix\footnote{\href{https://docs.scipy.org/doc/scipy/reference/generated/scipy.stats.ortho\_group.html}{SciPy Stats Module Docs}}.

Finally, multiply the first two columns of $R$ with either vector to generate the rotated vectors:
\begin{align}
    \mathbf{v}_1 = R[:, 1:2]\cdot \tilde{\mathbf{v}}_1;\\
    \mathbf{v}_2 = R[:, 1:2]\cdot \tilde{\mathbf{v}}_2.
\end{align}

For the more general case of multi-hidden units, $V$ is closer to an interpolation than a rotation. Specifically it is an interpolation between two random matrices such that $V=0$ gives a new random matrix that is orthogonal to the first, and $V=1$ gives back the same matrix as the first. Formally for similarity measure $V$ and first teacher feature weight matrix, $\wb^\dagger$, the second teacher feature weight matrix is given by:
\begin{equation}
    \wb^\ddag = V\wb^\dagger + \sqrt{(1 - V^2)}\mathbf{Z},
\end{equation}
where $\mathbf{Z}$ is a $D\times D$ random matrix.
\section{Empirical Node Importance}\label{app: node_importance_def}

While specialisation measures are very clearly identifiable via the overlap matrices in the teacher-student, analogues do not exist in the standard supervised learning setup. For this we use an empirical measure of node `importance' defined as the drop in test error when the node is masked. Formally for a two-layer network, node index $i$ and task index $t$:
\begin{align}
  \label{eq:node_importance_def}
  I_i^t\equiv\frac{1}{2}  \! &\left\langle  \! \left[\sum_{\substack{l=1 \\ l\neq i}}^L {\vb}_l g({\wb}_l\xb_j^t) \!\!-  \!\!y_j^t\right]^2 
   \!\!\!\!\! -  \!\!
  \left[\sum_{l=1}^L {\vb}_l g({\wb}_l\xb_j^t) \!\! - \!\! y_j^t\right]^2  \!\right\rangle
\end{align}
where $\vb$ are the second layer weights, $g$ is the activation function, $\wb$ are the first layer weights, and the averages are taken over the test dataset pairs $(\xb_j, y_j)$.

\newpage

\section{Detailed Statistics of Node Re-Use Tendency in Data Mixing}\label{app:ramasesh_statistics}

In~\autoref{fig:4c}, we show the $I^1\cdot I^2$ vs. $\alpha$. Specifically we plot the mean of 200 random seeds. At each value of $\alpha$ there is a high level of variance in $I^1\cdot I^2$. However each decile of the distribution of $I^2\cdot I^2$ follows a similar monotonic distribution, which we show in the plots below. This demonstrates that while there is a large width in the distribution of $I^1\cdot I^2$, there is a systematic shift upwards as $\alpha$ increases from 0 to 1.

\begin{figure}[h]
\centering
	\pgfplotsset{
		width=1.1\textwidth,
		height=0.8\textwidth,
		scaled x ticks=false,
		every tick label/.append style={font=\tiny},
		y label style={at={(axis description cs:-0.16, 0.5)}, rotate=0, anchor=south},
		x label style={at={(axis description cs:0.5, -0.42)}, rotate=0, anchor=south},
		xlabel={\scriptsize $\alpha$},
		ylabel={\scriptsize $I^1\cdot I^2$},
		xmin=0, xmax=1,
		}
	\begin{subfigure}[b]{0.235\textwidth}
	\begin{tikzpicture}
        \fill[shadecolor2, opacity=0] (0, 0) rectangle (\textwidth, 0.75\textwidth);
        \node [anchor=north west] at (0, 0.75\textwidth) {\emph{(a)}};
        \node [anchor=north west] at (1.6, 0.75\textwidth) {\scriptsize $0^\text{th}$ Percentile};
		\begin{axis}
			[
			at={(0.9cm, 0.75cm)},
			anchor=south west,
			ymin=0, ymax=0.009,
		]
		\addplot graphics [xmin=0, xmax=1,ymin=0,ymax=0.009] {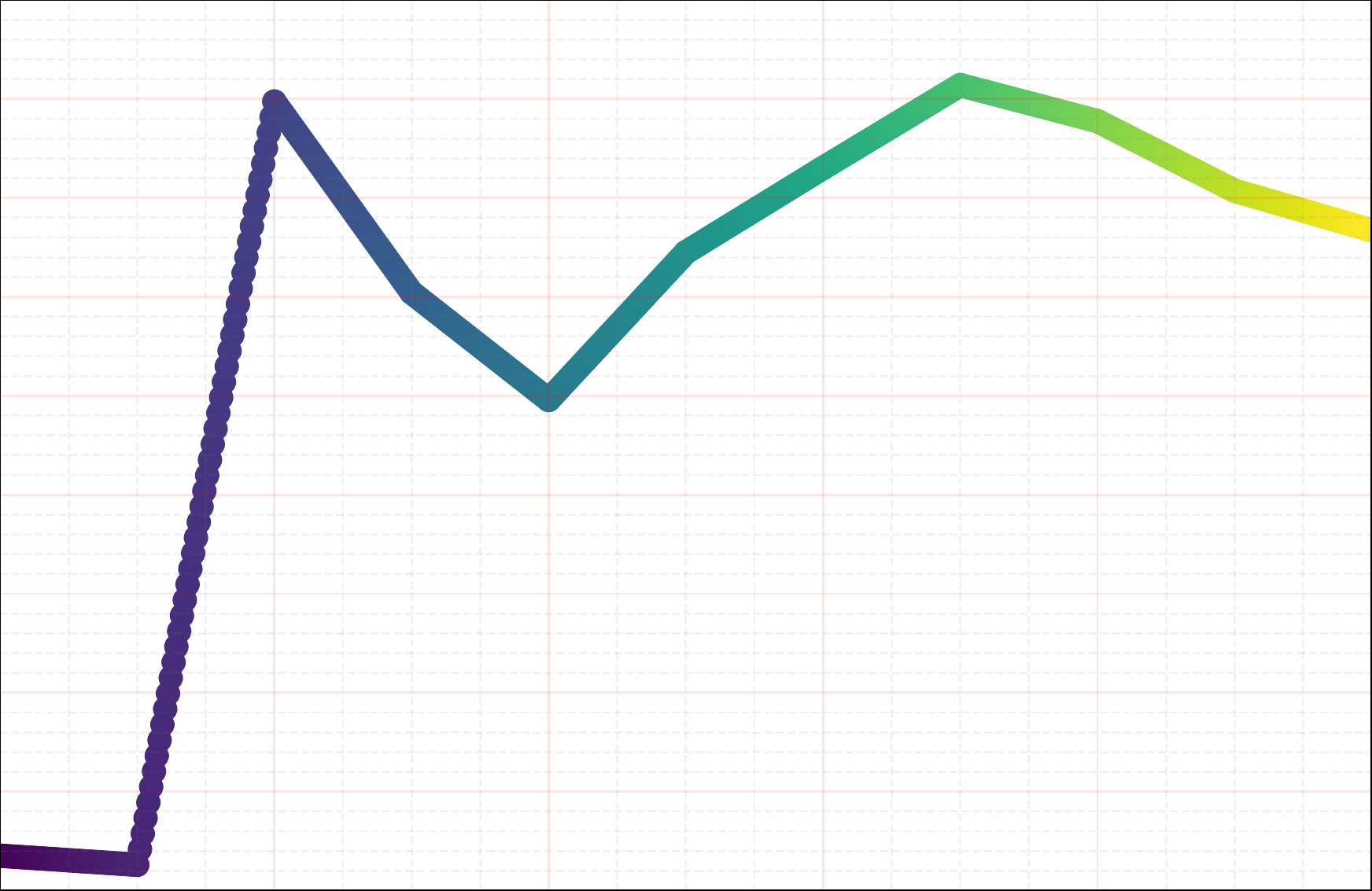};
		\end{axis}
	\end{tikzpicture}%
	\phantomcaption
	\end{subfigure}
	\begin{subfigure}[b]{0.235\textwidth}
	\begin{tikzpicture}
        \fill[shadecolor2, opacity=0] (0, 0) rectangle (\textwidth, 0.75\textwidth);
        \node [anchor=north west] at (0, 0.75\textwidth) {\emph{(b)}};
        \node [anchor=north west] at (1.6, 0.75\textwidth) {\scriptsize $10^\text{th}$ Percentile};
		\begin{axis}
			[
			at={(0.9cm, 0.75cm)},
			anchor=south west,
			xmin=0, xmax=1,
			ymin=0.005, ymax=0.03,
		]
		\addplot graphics [xmin=0, xmax=1,ymin=0.005, ymax=0.03] {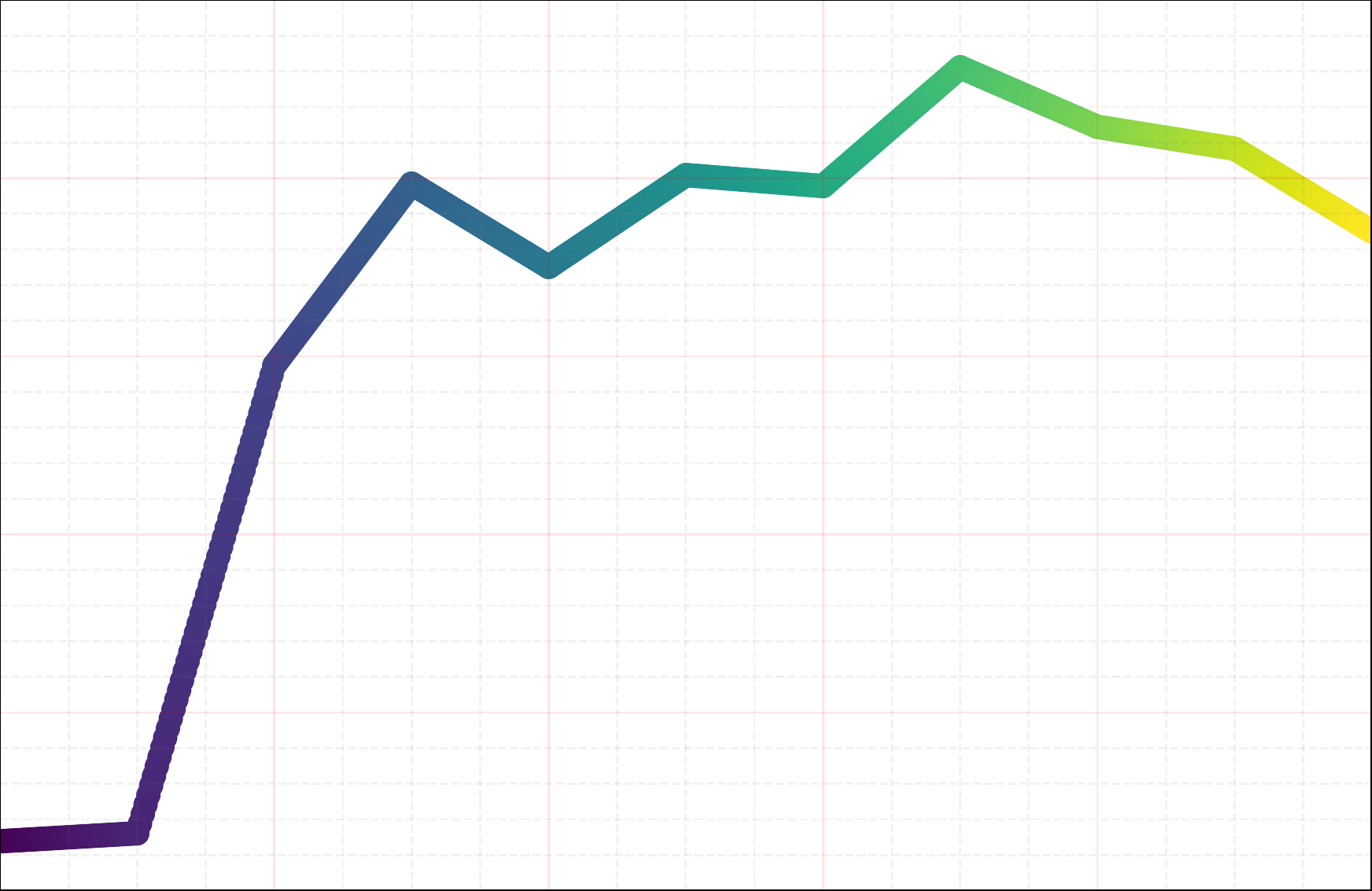};
		\end{axis}
	\end{tikzpicture}%
	\phantomcaption
	\end{subfigure}
	\hspace{0em}
	\begin{subfigure}[b]{0.235\textwidth}
	\begin{tikzpicture}
        \fill[shadecolor2, opacity=0] (0, 0) rectangle (\textwidth, 0.75\textwidth);
        \node [anchor=north west] at (0, 0.75\textwidth) {\emph{(c)}};
        \node [anchor=north west] at (1.6, 0.75\textwidth) {\scriptsize $20^\text{th}$ Percentile};
		\begin{axis}
			[
			at={(0.9cm, 0.75cm)},
			anchor=south west,
			xmin=0, xmax=1,
			ymin=0.01, ymax=0.06,
		]
		\addplot graphics [xmin=0, xmax=1,ymin=0.01,ymax=0.06] {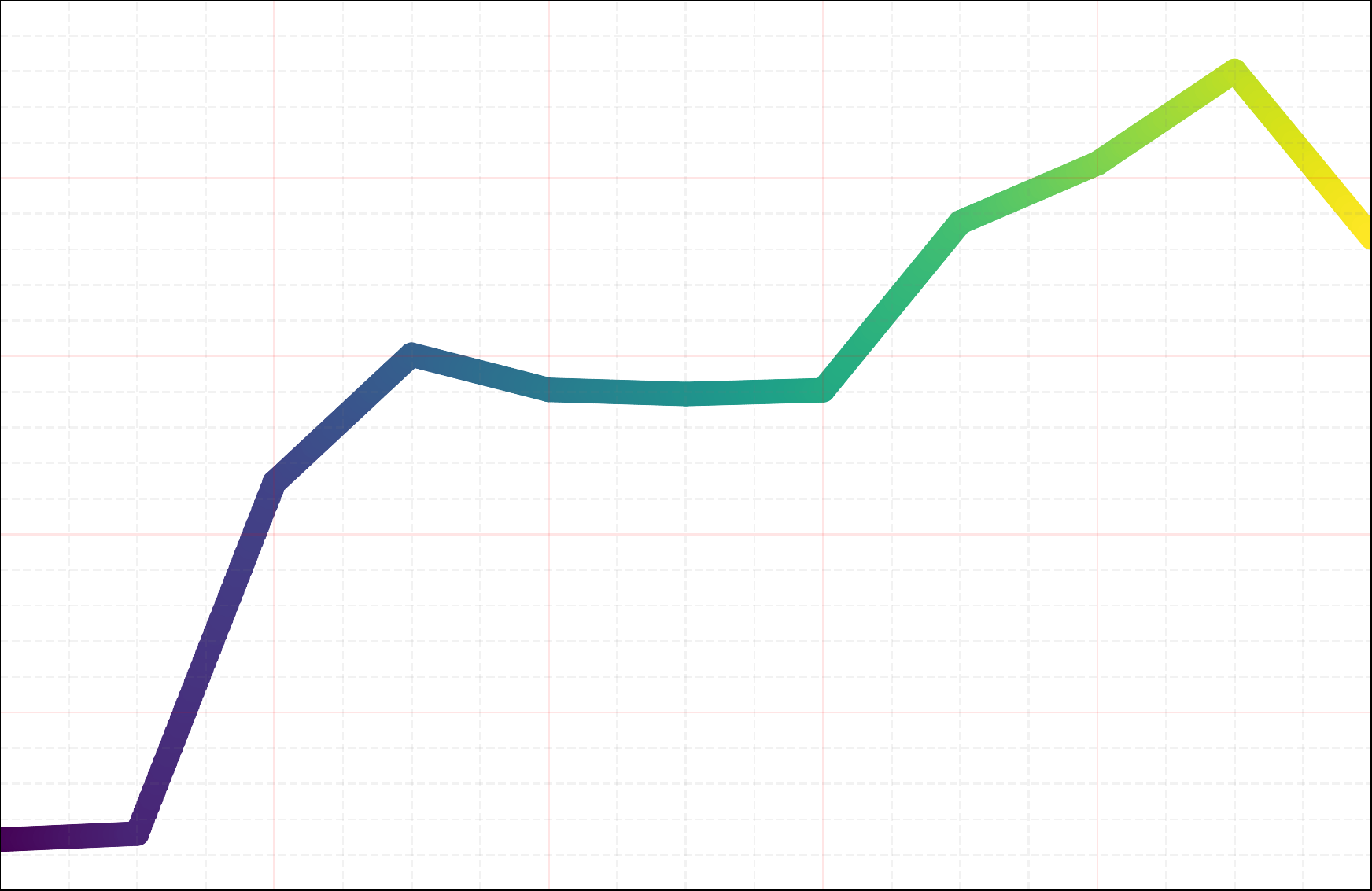};
		\end{axis}
	\end{tikzpicture}%
	\phantomcaption
	\end{subfigure}
	\begin{subfigure}[b]{0.235\textwidth}
	\begin{tikzpicture}
        \fill[shadecolor2, opacity=0] (0, 0) rectangle (\textwidth, 0.75\textwidth);
        \node [anchor=north west] at (0, 0.75\textwidth) {\emph{(d)}};
        \node [anchor=north west] at (1.6, 0.75\textwidth) {\scriptsize $30^\text{th}$ Percentile};
		\begin{axis}
			[
			at={(0.9cm, 0.75cm)},
			anchor=south west,
			xmin=0, xmax=1,
			ymin=0.015, ymax=0.09,
		]
		\addplot graphics [xmin=0, xmax=1,ymin=0.015,ymax=0.09] {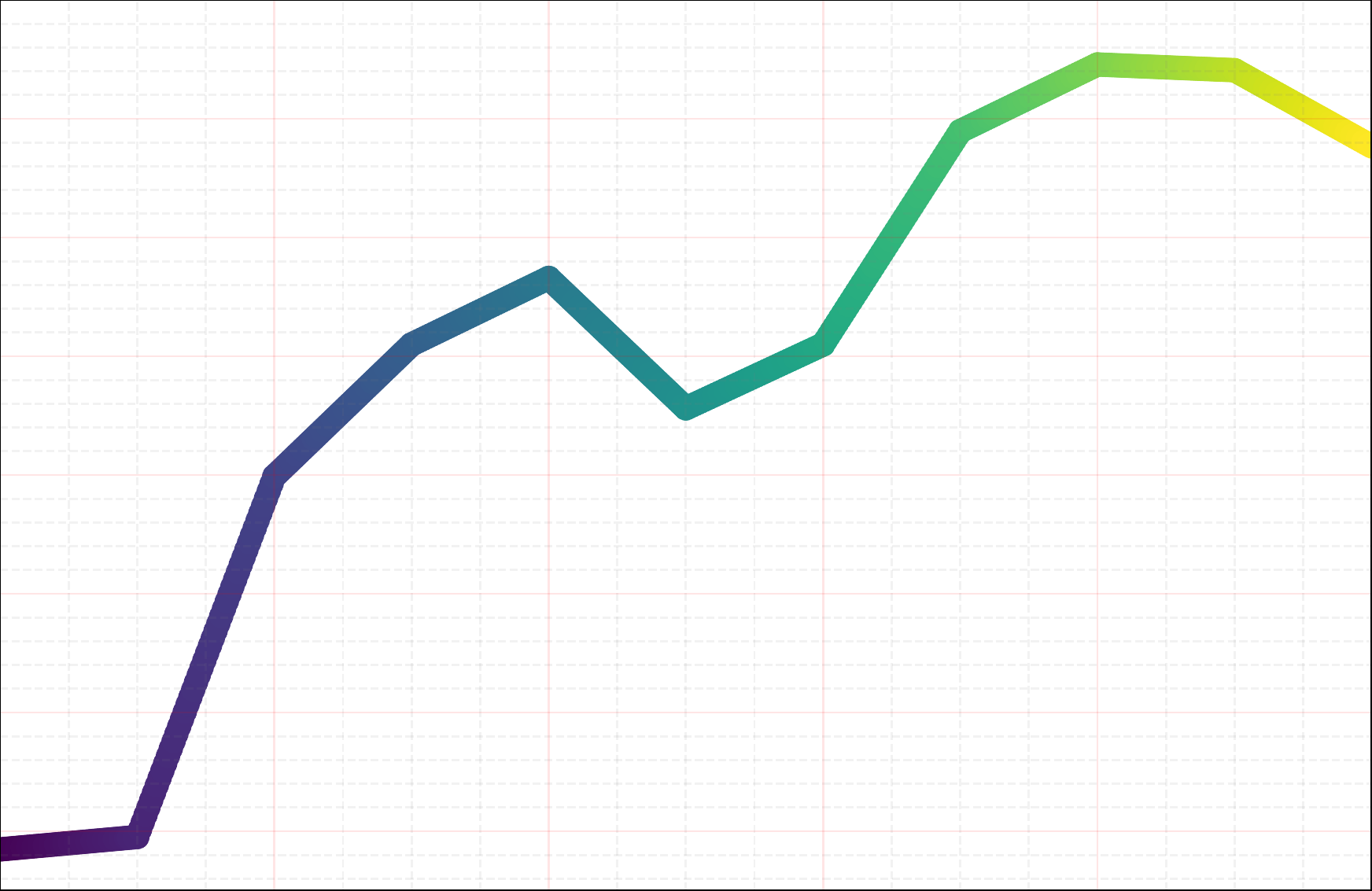};
		\end{axis}
	\end{tikzpicture}%
	\phantomcaption
	\end{subfigure}
	\begin{subfigure}[b]{0.235\textwidth}
	\begin{tikzpicture}
        \fill[shadecolor2, opacity=0] (0, 0) rectangle (\textwidth, 0.75\textwidth);
        \node [anchor=north west] at (0, 0.75\textwidth) {\emph{(e)}};
        \node [anchor=north west] at (1.6, 0.75\textwidth) {\scriptsize $40^\text{th}$ Percentile};
		\begin{axis}
			[
			at={(0.9cm, 0.75cm)},
			anchor=south west,
			xmin=0, xmax=1,
			ymin=0.02, ymax=0.13,
		]
		\addplot graphics [xmin=0, xmax=1,ymin=0.02,ymax=0.13] {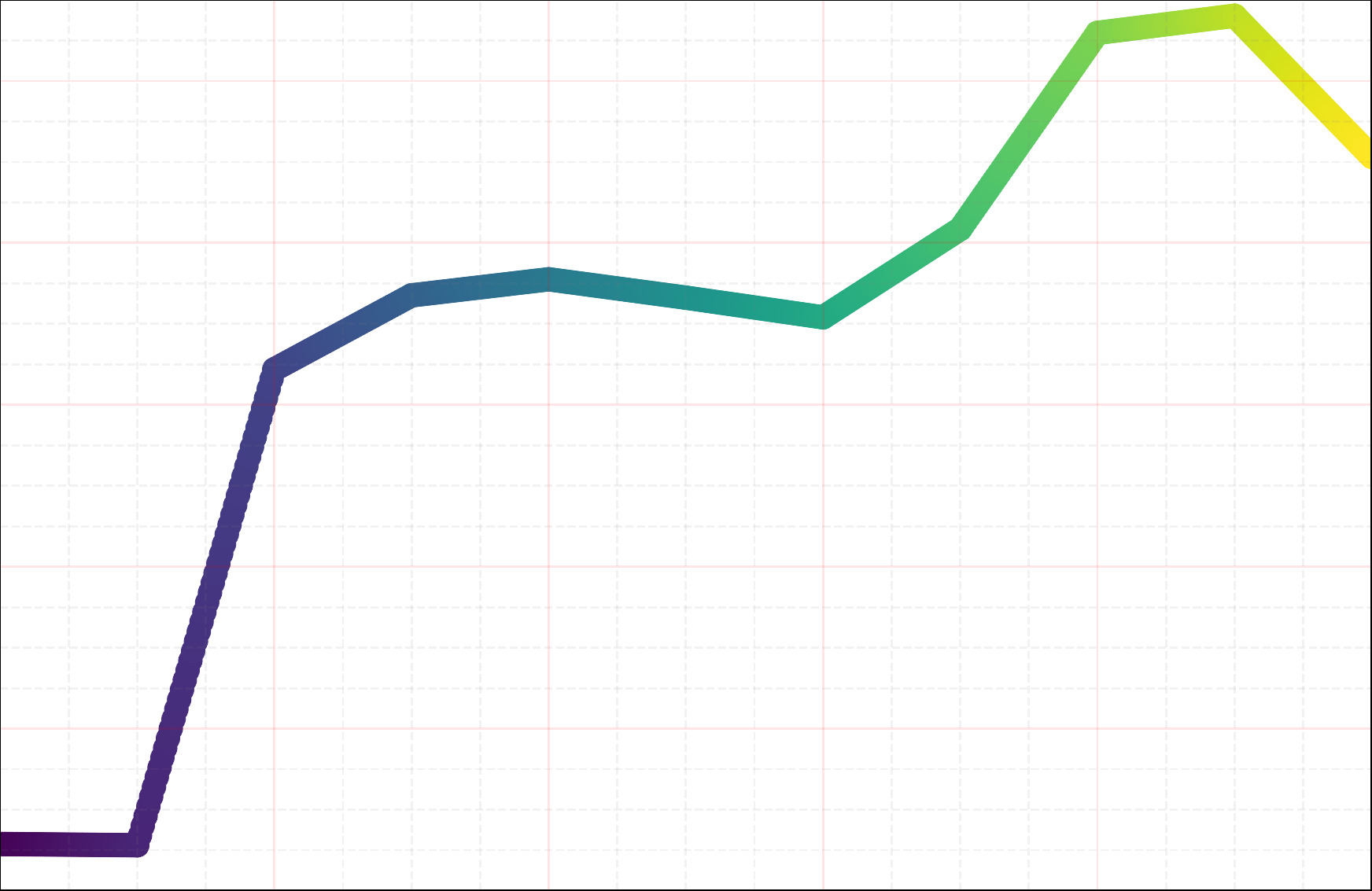};
		\end{axis}
	\end{tikzpicture}%
	\phantomcaption
	\end{subfigure}
	\begin{subfigure}[b]{0.235\textwidth}
	\begin{tikzpicture}
        \fill[shadecolor2, opacity=0] (0, 0) rectangle (\textwidth, 0.75\textwidth);
        \node [anchor=north west] at (0, 0.75\textwidth) {\emph{(f)}};
        \node [anchor=north west] at (1.6, 0.75\textwidth) {\scriptsize $50^\text{th}$ Percentile};
		\begin{axis}
			[
			at={(0.9cm, 0.75cm)},
			anchor=south west,
			xmin=0, xmax=1,
			ymin=0.03, ymax=0.18,
		]
		\addplot graphics [xmin=0, xmax=1,ymin=0.03,ymax=0.18] {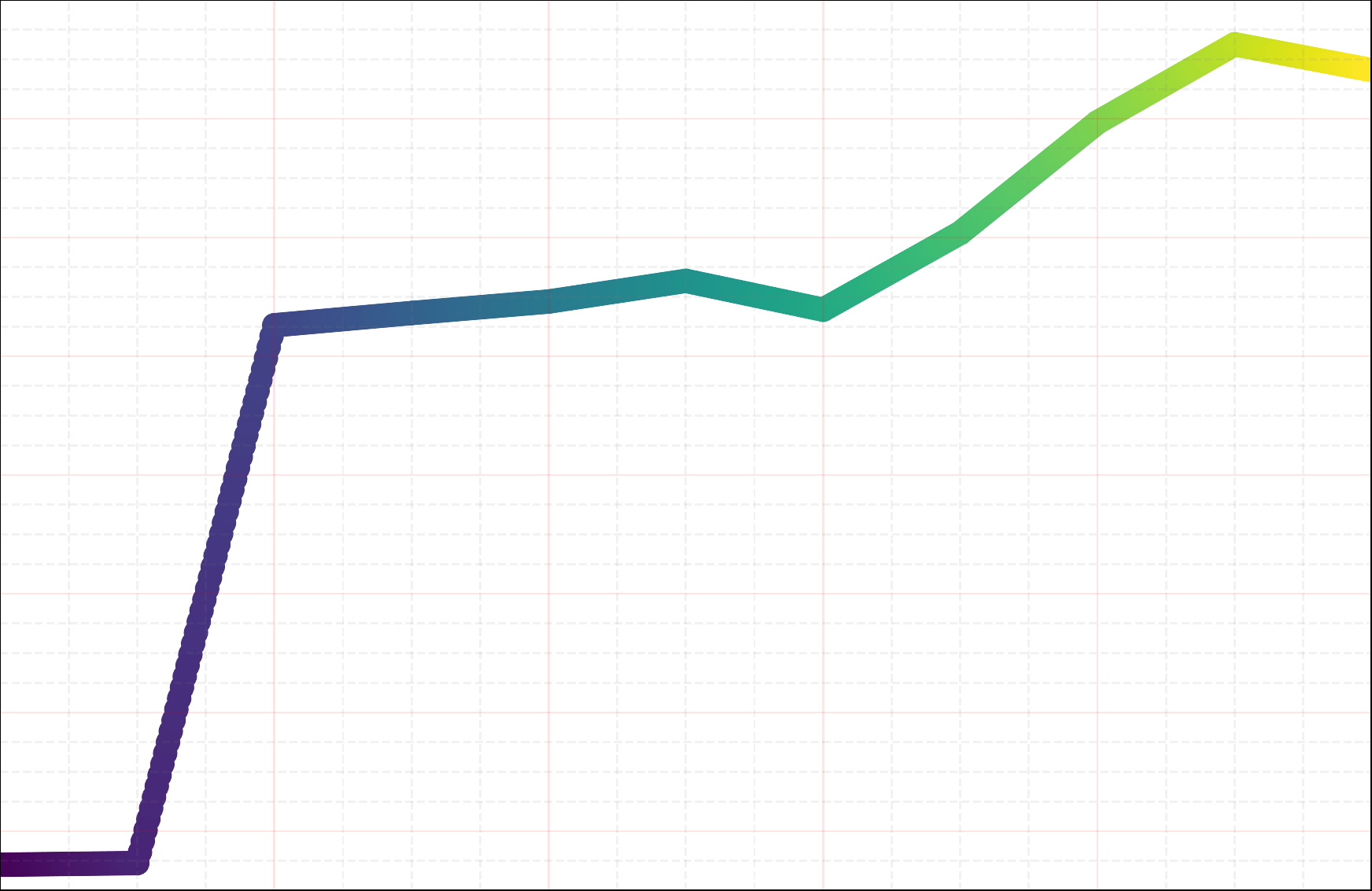};
		\end{axis}
	\end{tikzpicture}%
	\phantomcaption
	\end{subfigure}
	\hspace{0em}
	\begin{subfigure}[b]{0.235\textwidth}
	\begin{tikzpicture}
        \fill[shadecolor2, opacity=0] (0, 0) rectangle (\textwidth, 0.75\textwidth);
        \node [anchor=north west] at (0, 0.75\textwidth) {\emph{(g)}};
        \node [anchor=north west] at (1.6, 0.75\textwidth) {\scriptsize $60^\text{th}$ Percentile};
		\begin{axis}
			[
			at={(0.9cm, 0.75cm)},
			anchor=south west,
			xmin=0, xmax=1,
			ymin=0.045, ymax=0.225,
		]
		\addplot graphics [xmin=0, xmax=1,ymin=0.045,ymax=0.225] {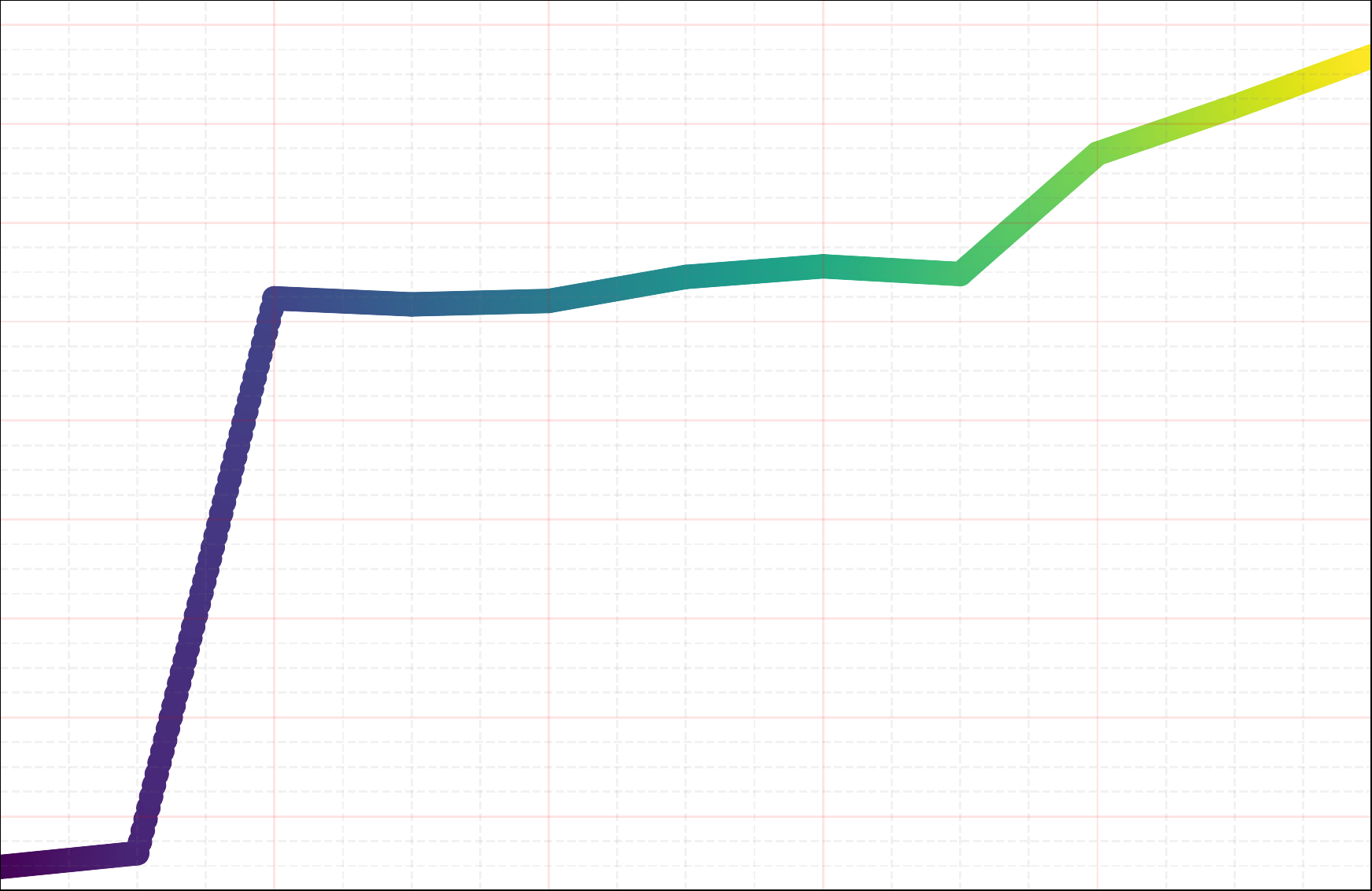};
		\end{axis}
	\end{tikzpicture}%
	\phantomcaption
	\end{subfigure}
	\begin{subfigure}[b]{0.235\textwidth}
	\begin{tikzpicture}
        \fill[shadecolor2, opacity=0] (0, 0) rectangle (\textwidth, 0.75\textwidth);
        \node [anchor=north west] at (0, 0.75\textwidth) {\emph{(h)}};
        \node [anchor=north west] at (1.6, 0.75\textwidth) {\scriptsize $70^\text{th}$ Percentile};
		\begin{axis}
			[
			at={(0.9cm, 0.75cm)},
			anchor=south west,
			xmin=0, xmax=1,
			ymin=0.07, ymax=0.3,
		]
		\addplot graphics [xmin=0, xmax=1,ymin=0.07,ymax=0.3] {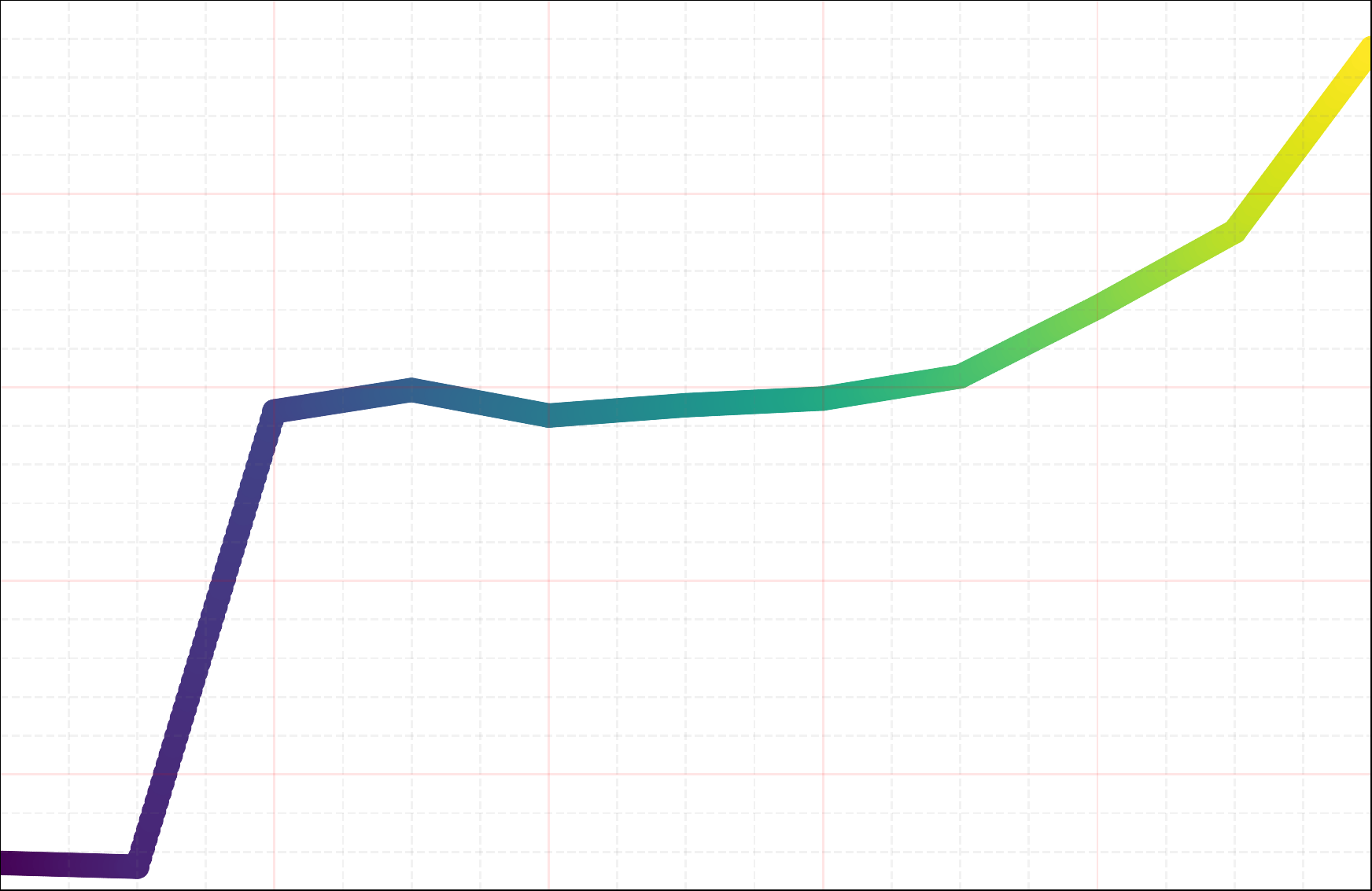};
		\end{axis}
	\end{tikzpicture}%
	\phantomcaption
	\end{subfigure}
	\begin{subfigure}[b]{0.235\textwidth}
	\begin{tikzpicture}
        \fill[shadecolor2, opacity=0] (0, 0) rectangle (\textwidth, 0.75\textwidth);
        \node [anchor=north west] at (0, 0.75\textwidth) {\emph{(i)}};
        \node [anchor=north west] at (1.6, 0.75\textwidth) {\scriptsize $80^\text{th}$ Percentile};
		\begin{axis}
			[
			at={(0.9cm, 0.75cm)},
			anchor=south west,
			xmin=0, xmax=1,
			ymin=0.14, ymax=0.45,
		]
		\addplot graphics [xmin=0, xmax=1,ymin=0.14,ymax=0.45] {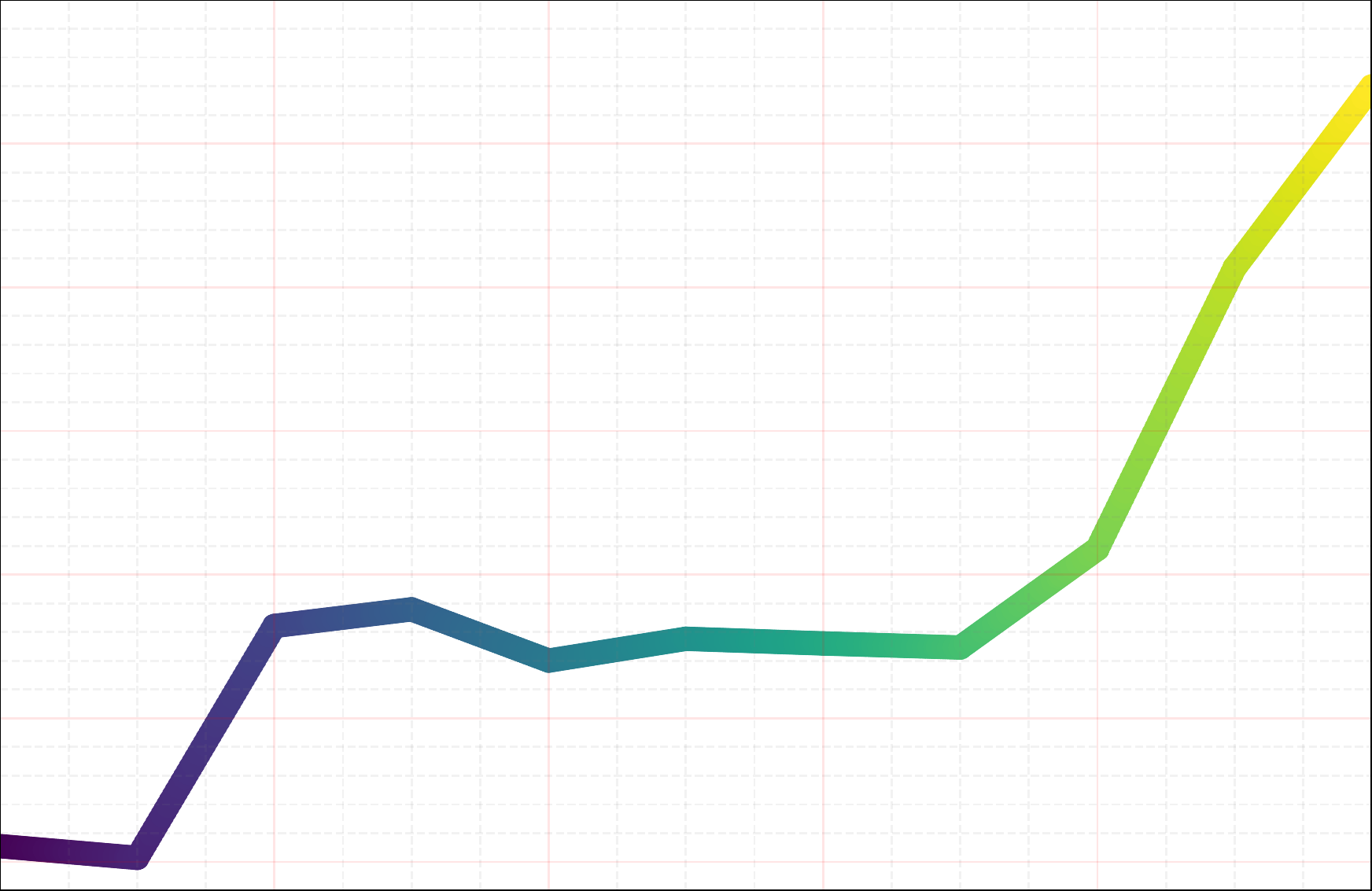};
		\end{axis}
	\end{tikzpicture}%
	\phantomcaption
	\end{subfigure}
	\begin{subfigure}[b]{0.235\textwidth}
	\begin{tikzpicture}
        \fill[shadecolor2, opacity=0] (0, 0) rectangle (\textwidth, 0.75\textwidth);
        \node [anchor=north west] at (0, 0.75\textwidth) {\emph{(j)}};
        \node [anchor=north west] at (1.6, 0.75\textwidth) {\scriptsize $90^\text{th}$ Percentile};
		\begin{axis}
			[
			at={(0.9cm, 0.75cm)},
			anchor=south west,
			xmin=0, xmax=1,
			ymin=0.25, ymax=0.6,
		]
		\addplot graphics [xmin=0, xmax=1,ymin=0.25,ymax=0.6] {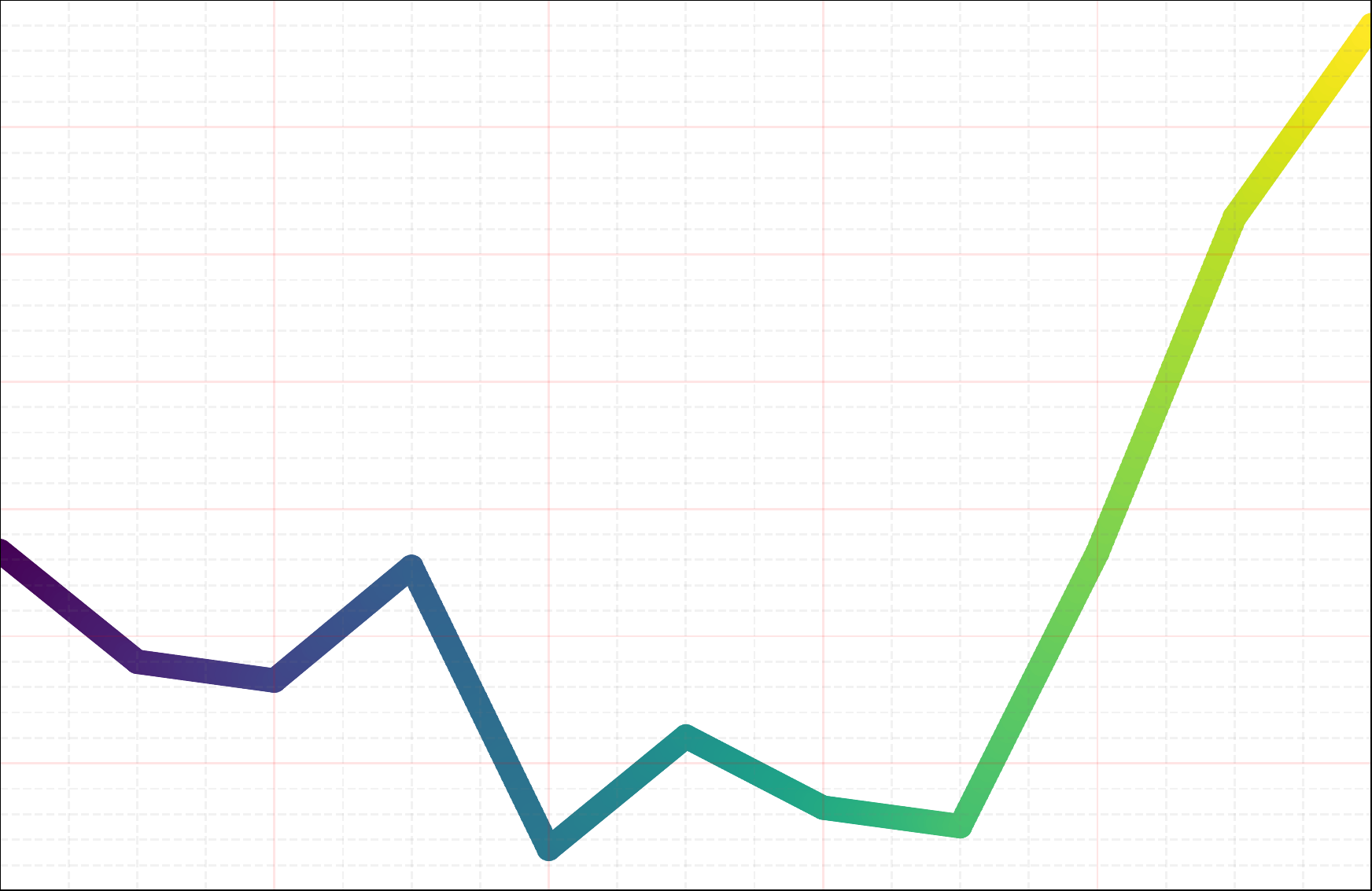};
		\end{axis}
	\end{tikzpicture}%
	\phantomcaption
	\end{subfigure}
	\hspace{0em}
	\begin{subfigure}[b]{0.235\textwidth}
	\begin{tikzpicture}
        \fill[shadecolor2, opacity=0] (0, 0) rectangle (\textwidth, 0.75\textwidth);
        \node [anchor=north west] at (0, 0.75\textwidth) {\emph{(k)}};
        \node [anchor=north west] at (1.6, 0.75\textwidth) {\scriptsize $100^\text{th}$ Percentile};
		\begin{axis}
			[
			at={(0.9cm, 0.75cm)},
			anchor=south west,
			xmin=0, xmax=1,
			ymin=0.5, ymax=1.4,
		]
		\addplot graphics [xmin=0, xmax=1,ymin=0.5,ymax=1.4] {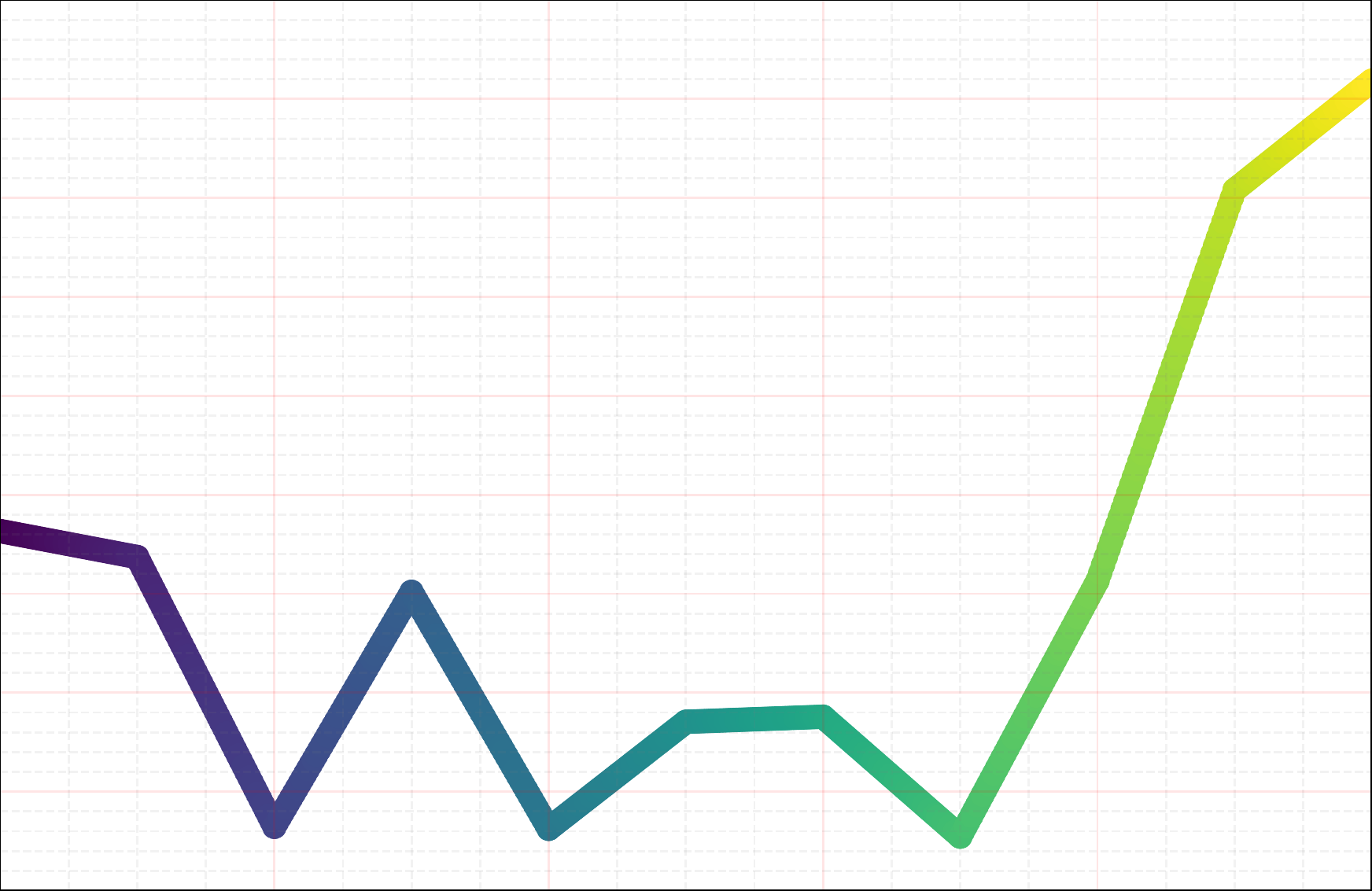};
		\end{axis}
	\end{tikzpicture}%
	\phantomcaption
	\end{subfigure}
	\hspace{0em}
	\caption[]{\textbf{Deciles of $I^1\cdot I^2$ distributions over 200 seeds vs. mixing parameter $\alpha$} In every decile there is a general monotonic relationship between $I^1\cdot I^2$ and the mixing parameter $\alpha$.\label{fig:importance_dots_deciles}}
\end{figure}
\section{Overlap Plots}\label{app: overlaps}

In this section we show plots of the overlap parameters corresponding to the various figures in the main text. These can help to illuminate some of the behaviours at the node level that give rise to the macroscopic phenomena we observe e.g. in the generalisation errors. 

\subsection{Effect of EWC on task similarity vs. forgetting}

These are the overlaps associated with~\autoref{fig:ewc}.

\begin{figure}[h]
	\pgfplotsset{
		width=1.1\textwidth,
		height=0.8\textwidth,
		scaled x ticks=false,
		every tick label/.append style={font=\tiny},
		ylabel={\scriptsize $\mathbf{Q}$},
		y label style={at={(axis description cs:-0.16, 0.5)}, rotate=0, anchor=south},
		x label style={at={(axis description cs:0.5, -0.42)}, rotate=0, anchor=south},
		ymin=0, ymax=1.4,
		xlabel={\scriptsize $s$},
		}
	\begin{subfigure}[b]{0.235\textwidth}
	\begin{tikzpicture}
        \fill[shadecolor2, opacity=0] (0, 0) rectangle (\textwidth, 0.75\textwidth);
        \node [anchor=north west] at (0, 0.75\textwidth) {\emph{(a)}};
        \node [anchor=north west] at (1.6, 0.75\textwidth) {\scriptsize $\lambda=0$};
		\begin{axis}
			[
			at={(0.9cm, 0.75cm)},
			anchor=south west,
			xmin=0, xmax=12000000,
		]
		\addplot graphics [xmin=0, xmax=12000000,ymin=0,ymax=1.4] {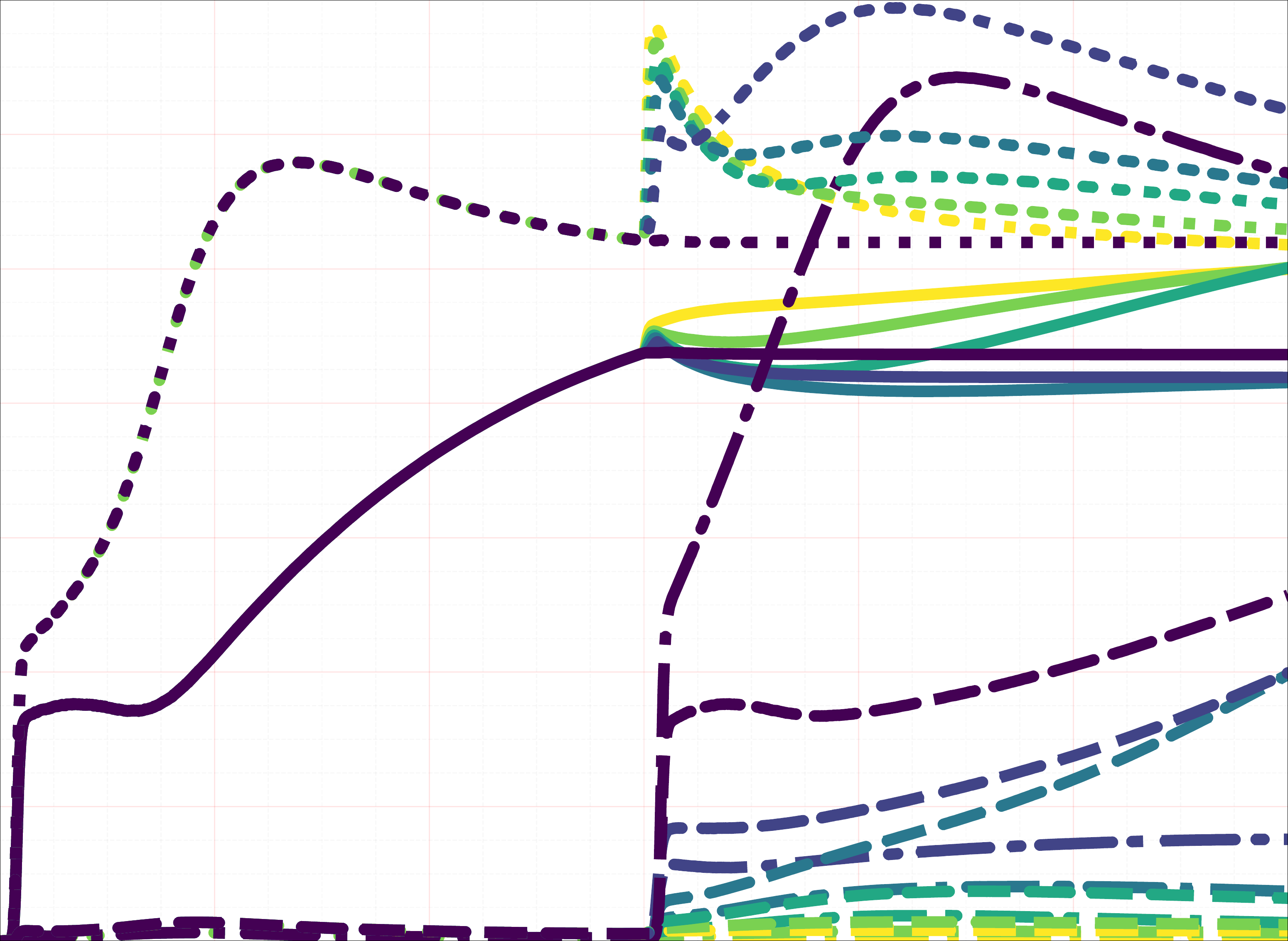};
		\end{axis}
	\end{tikzpicture}%
	\phantomcaption
	\end{subfigure}
	\begin{subfigure}[b]{0.235\textwidth}
	\begin{tikzpicture}
        \fill[shadecolor2, opacity=0] (0, 0) rectangle (\textwidth, 0.75\textwidth);
        \node [anchor=north west] at (0, 0.75\textwidth) {\emph{(b)}};
        \node [anchor=north west] at (1.6, 0.75\textwidth) {\scriptsize $\lambda=100$};
		\begin{axis}
			[
			at={(0.9cm, 0.75cm)},
			anchor=south west,
			xmin=0, xmax=12000000,
		]
		\addplot graphics [xmin=0, xmax=12000000,ymin=0,ymax=1.4] {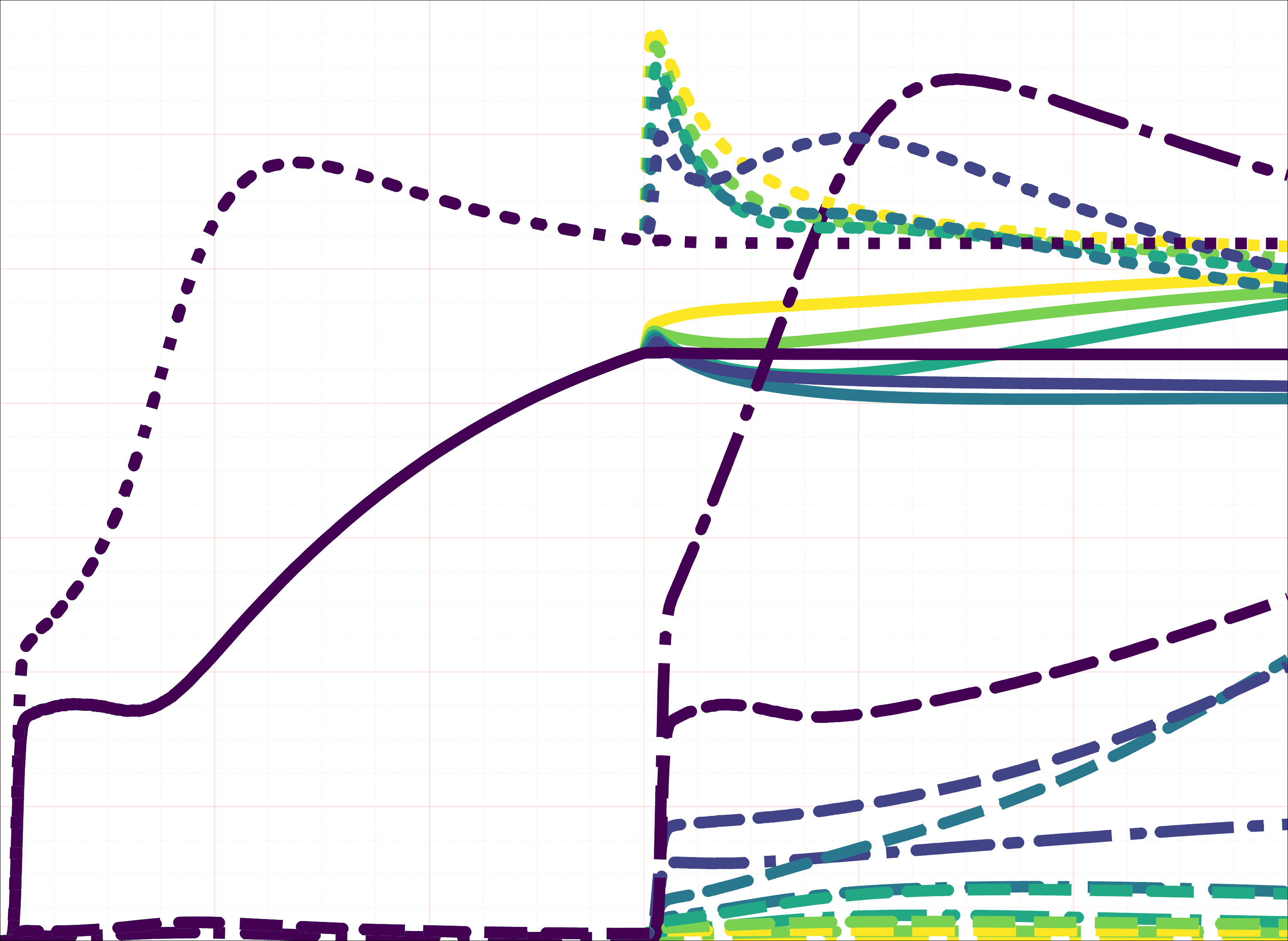};
		\end{axis}
	\end{tikzpicture}%
	\phantomcaption
	\end{subfigure}
	\hspace{0em}
	\begin{subfigure}[b]{0.235\textwidth}
	\begin{tikzpicture}
        \fill[shadecolor2, opacity=0] (0, 0) rectangle (\textwidth, 0.75\textwidth);
        \node [anchor=north west] at (0, 0.75\textwidth) {\emph{(c)}};
        \node [anchor=north west] at (1.6, 0.75\textwidth) {\scriptsize $\lambda=1000$};
		\begin{axis}
			[
			at={(0.9cm, 0.75cm)},
			anchor=south west,
			xmin=0, xmax=12000000,
		]
		\addplot graphics [xmin=0, xmax=12000000,ymin=0,ymax=1.4] {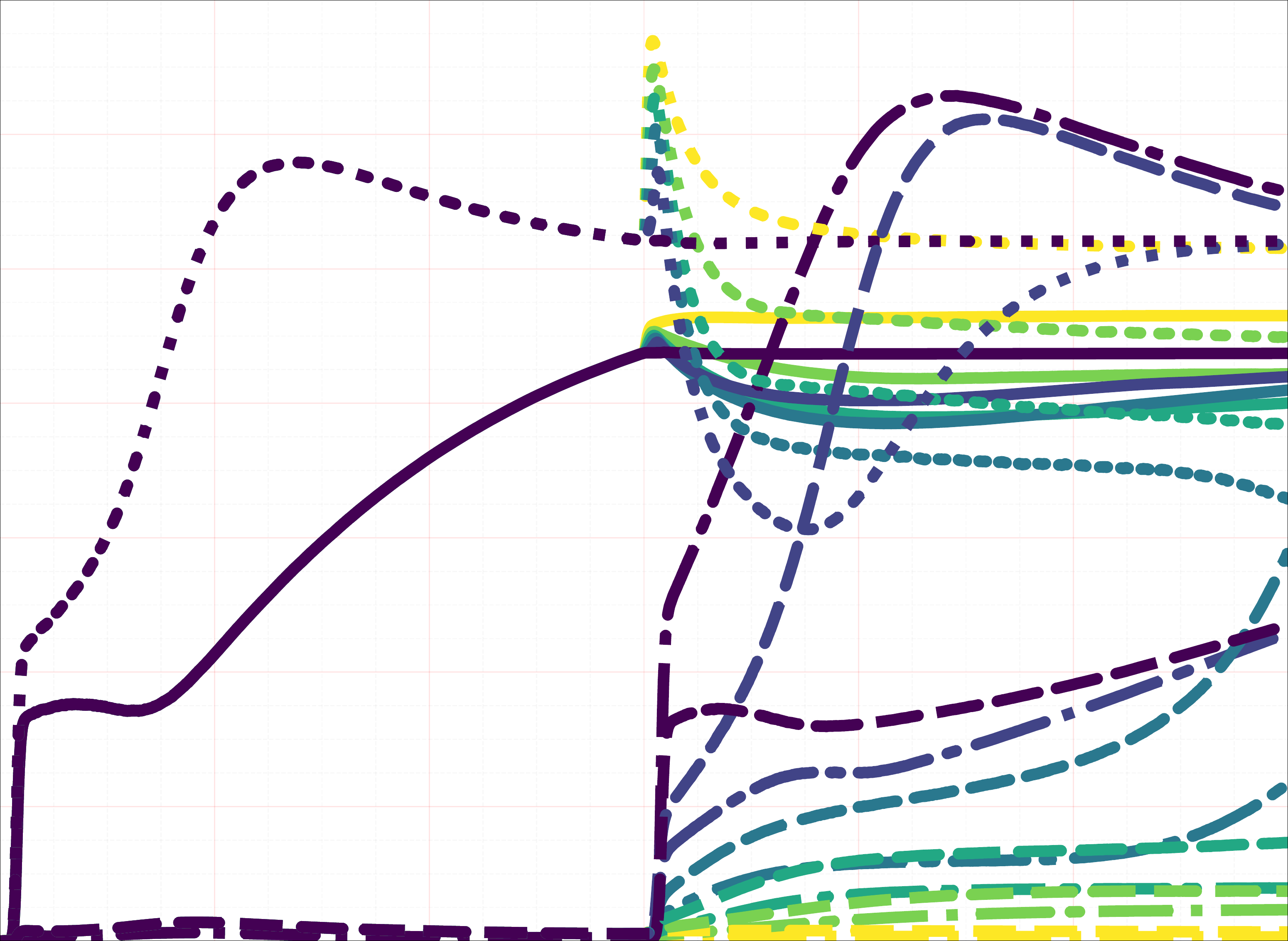};
		\end{axis}
	\end{tikzpicture}%
	\phantomcaption
	\end{subfigure}
	\begin{subfigure}[b]{0.235\textwidth}
	\begin{tikzpicture}
        \fill[shadecolor2, opacity=0] (0, 0) rectangle (\textwidth, 0.75\textwidth);
        \node [anchor=north west] at (0, 0.75\textwidth) {\emph{(d)}};
        \node [anchor=north west] at (1.6, 0.75\textwidth) {\scriptsize $\lambda=10000$};
		\begin{axis}
			[
			at={(0.9cm, 0.75cm)},
			anchor=south west,
			xmin=0, xmax=12000000,
		]
		\addplot graphics [xmin=0, xmax=12000000,ymin=0,ymax=1.4] {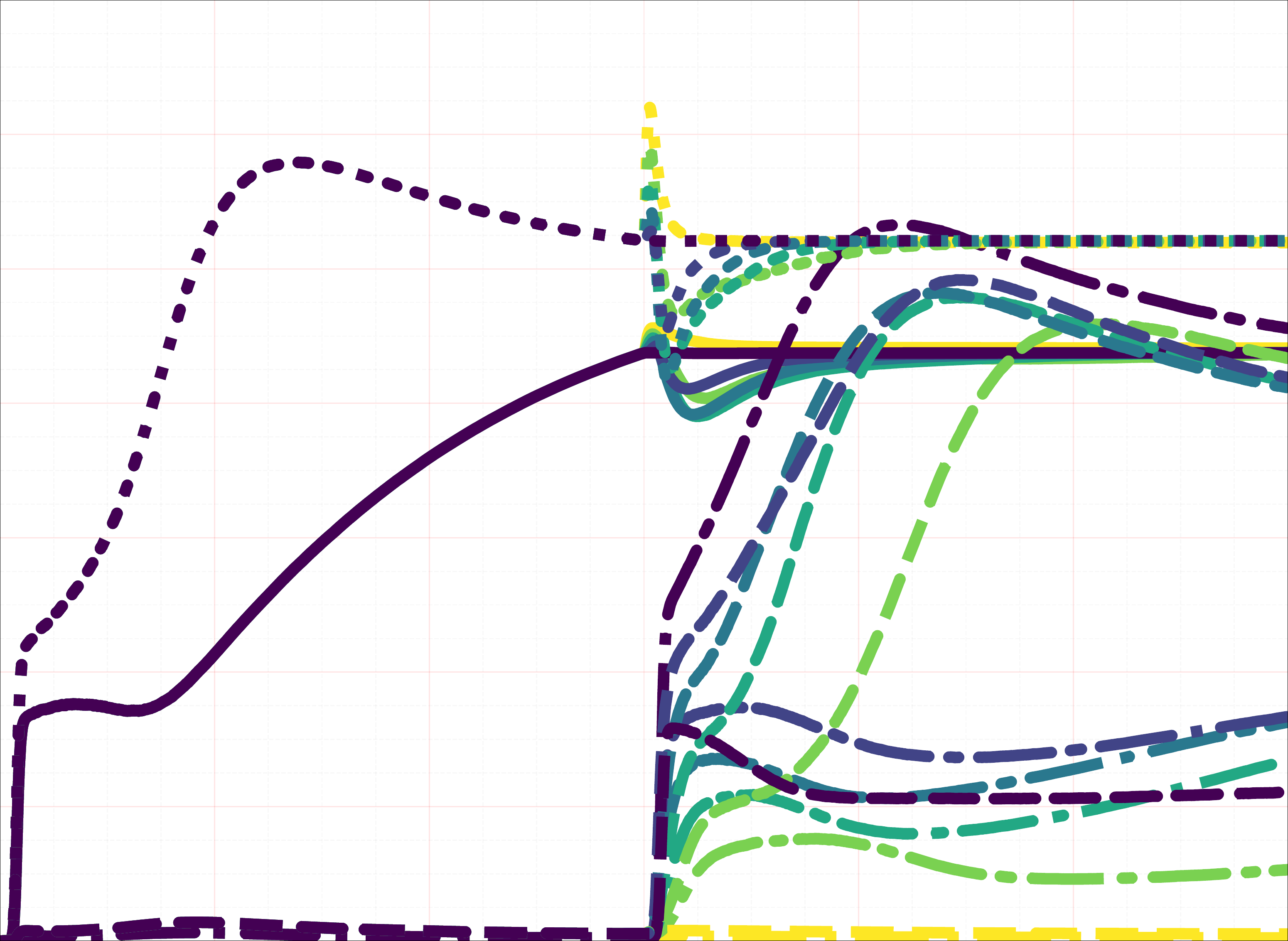};
		\end{axis}
	\end{tikzpicture}%
	\phantomcaption
	\end{subfigure}
	\hspace{0em}
	\caption[]{\textbf{Effect of EWC on task similarity vs. forgetting: student self-overlaps}\label{fig:self_overlaps_ewc}}
\end{figure}
\input{app_figs_code/ewc_student_teacher_overlaps}

\subsection{Effect of interleaving on task similarity vs. forgetting}

These are the overlaps associated with~\autoref{fig:interleave}.

\begin{figure}[h]
	\pgfplotsset{
		width=1.1\textwidth,
		height=0.8\textwidth,
		scaled x ticks=false,
		every tick label/.append style={font=\tiny},
		ylabel={\scriptsize $\mathbf{Q}$},
		y label style={at={(axis description cs:-0.16, 0.5)}, rotate=0, anchor=south},
		x label style={at={(axis description cs:0.5, -0.42)}, rotate=0, anchor=south},
		ymin=0, ymax=1.4,
		xlabel={\scriptsize $s$},
		}
	\begin{subfigure}[b]{0.235\textwidth}
	\begin{tikzpicture}
        \fill[shadecolor2, opacity=0] (0, 0) rectangle (\textwidth, 0.75\textwidth);
        \node [anchor=north west] at (0, 0.75\textwidth) {\emph{(a)}};
        \node [anchor=north west] at (1.6, 0.75\textwidth) {\scriptsize $T=100$};
		\begin{axis}
			[
			at={(0.9cm, 0.75cm)},
			anchor=south west,
			xmin=0, xmax=12000000,
		]
		\addplot graphics [xmin=0, xmax=12000000,ymin=0,ymax=1.4] {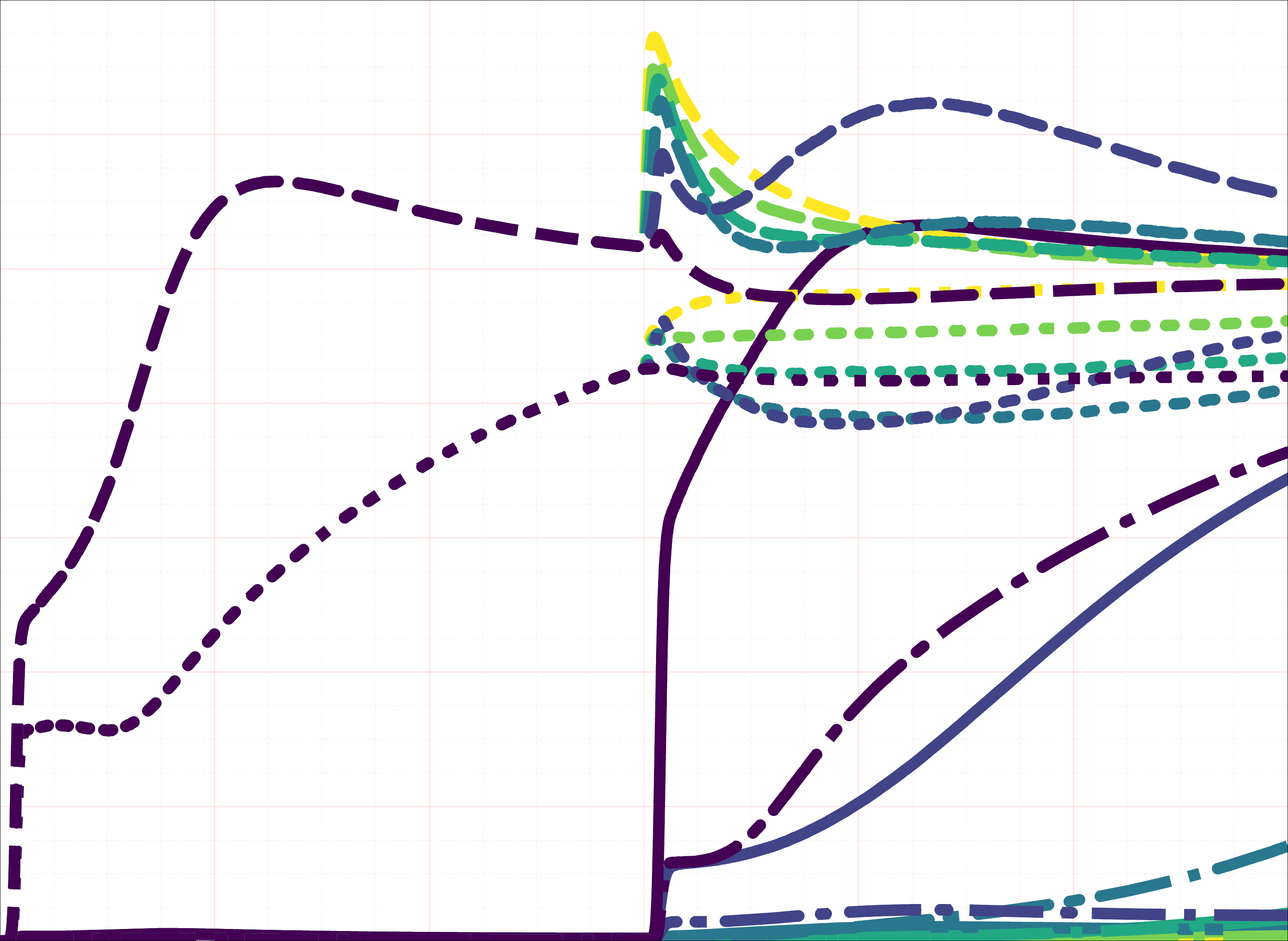};
		\end{axis}
	\end{tikzpicture}%
	\phantomcaption
	\end{subfigure}
	\begin{subfigure}[b]{0.235\textwidth}
	\begin{tikzpicture}
        \fill[shadecolor2, opacity=0] (0, 0) rectangle (\textwidth, 0.75\textwidth);
        \node [anchor=north west] at (0, 0.75\textwidth) {\emph{(b)}};
        \node [anchor=north west] at (1.6, 0.75\textwidth) {\scriptsize $T=10$};
		\begin{axis}
			[
			at={(0.9cm, 0.75cm)},
			anchor=south west,
			xmin=0, xmax=12000000,
		]
		\addplot graphics [xmin=0, xmax=12000000,ymin=0,ymax=1.4] {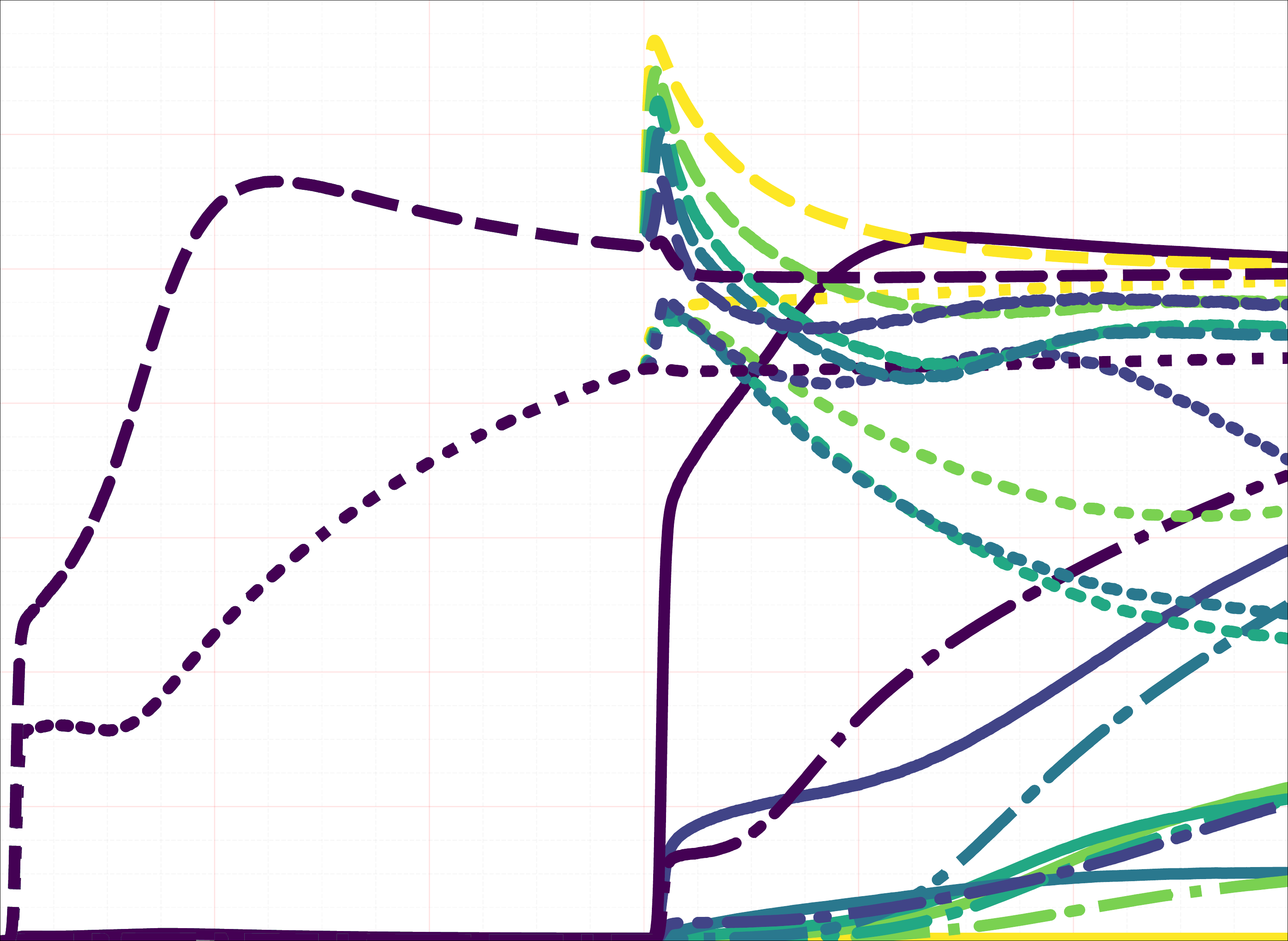};
		\end{axis}
	\end{tikzpicture}%
	\phantomcaption
	\end{subfigure}
	\hspace{0em}
	\begin{subfigure}[b]{0.235\textwidth}
	\begin{tikzpicture}
        \fill[shadecolor2, opacity=0] (0, 0) rectangle (\textwidth, 0.75\textwidth);
        \node [anchor=north west] at (0, 0.75\textwidth) {\emph{(c)}};
        \node [anchor=north west] at (1.6, 0.75\textwidth) {\scriptsize $T=2$};
		\begin{axis}
			[
			at={(0.9cm, 0.75cm)},
			anchor=south west,
			xmin=0, xmax=12000000,
		]
		\addplot graphics [xmin=0, xmax=12000000,ymin=0,ymax=1.4] {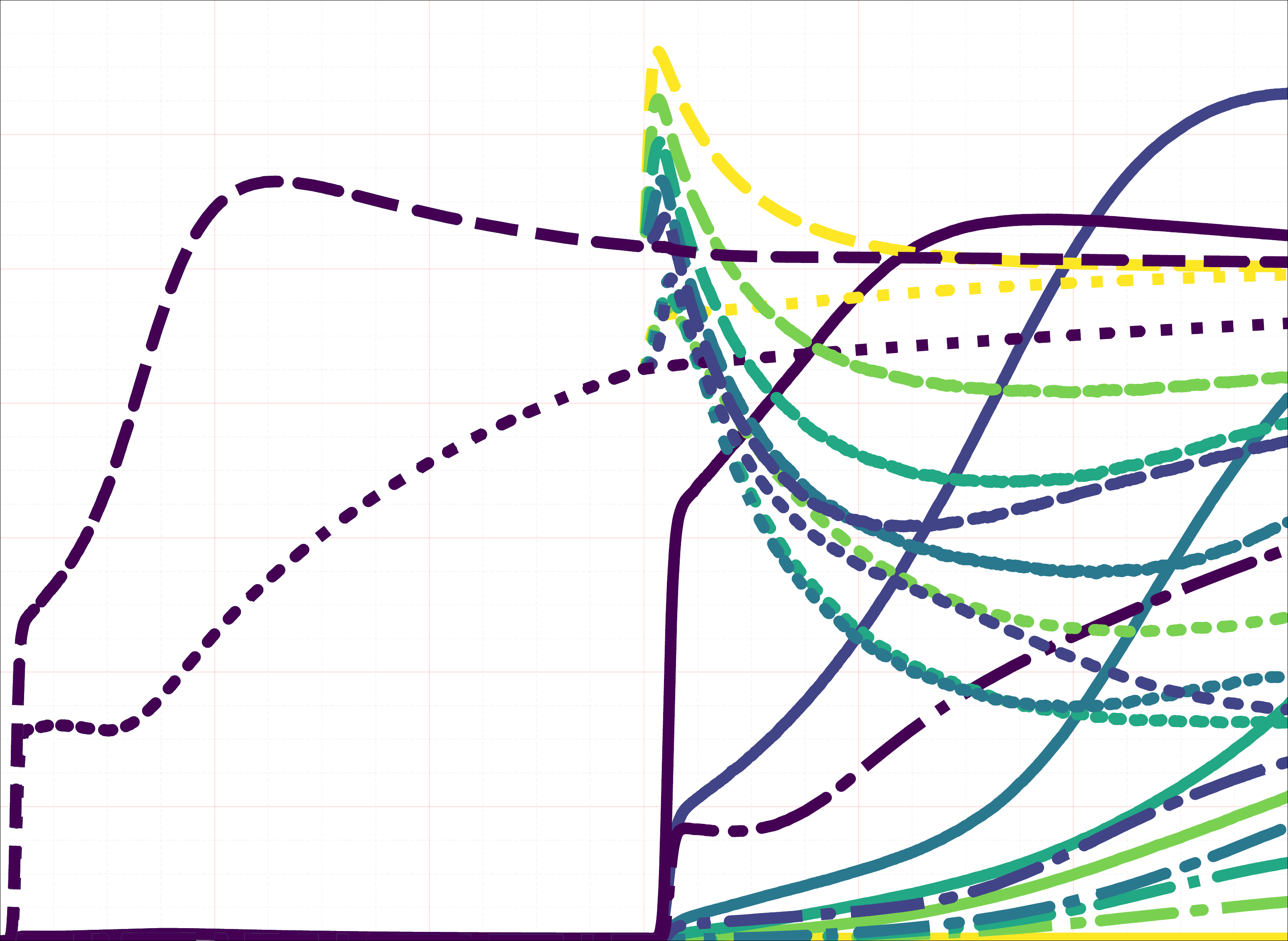};
		\end{axis}
	\end{tikzpicture}%
	\phantomcaption
	\end{subfigure}
	\begin{subfigure}[b]{0.235\textwidth}
	\begin{tikzpicture}
        \fill[shadecolor2, opacity=0] (0, 0) rectangle (\textwidth, 0.75\textwidth);
        \node [anchor=north west] at (0, 0.75\textwidth) {\emph{(d)}};
        \node [anchor=north west] at (1.6, 0.75\textwidth) {\scriptsize $T=1$};
		\begin{axis}
			[
			at={(0.9cm, 0.75cm)},
			anchor=south west,
			xmin=0, xmax=12000000,
		]
		\addplot graphics [xmin=0, xmax=12000000,ymin=0,ymax=1.4] {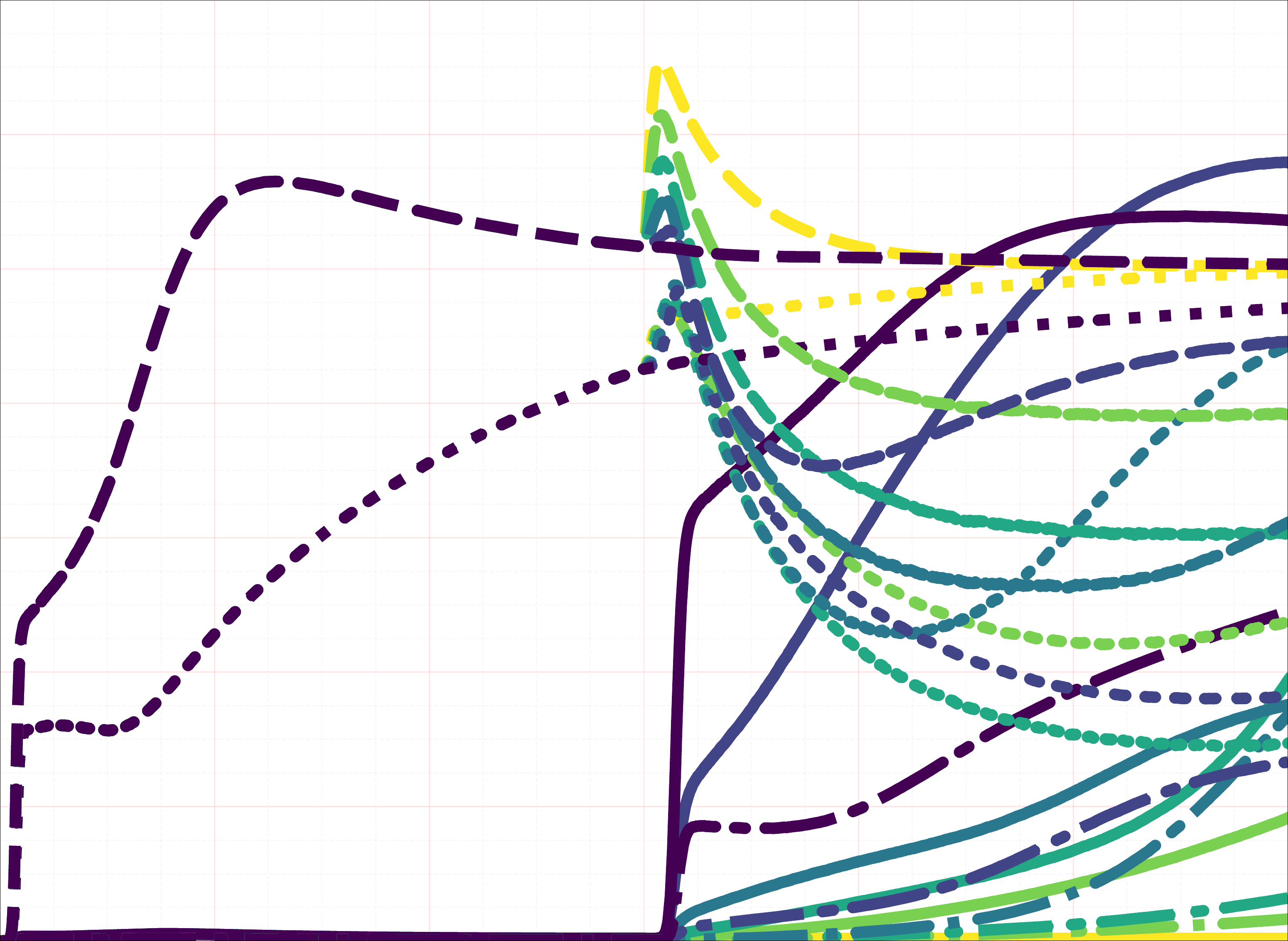};
		\end{axis}
	\end{tikzpicture}%
	\phantomcaption
	\end{subfigure}
	\hspace{0em}
	\caption[]{\textbf{Effect of interleaving on task similarity vs. forgetting: student self-overlaps}\label{fig:self_overlaps_interleave}}
\end{figure}
\input{app_figs_code/interleave_student_teacher_overlaps}

\newpage

\section{Transfer Plots}\label{app: transfer_supp}

This section contains plots on the `transfer' observed in the performance on task 2. We follow~\cite{lee2021continual} in defining transfer as the difference between the generalisation error on the second teacher at some step after switch and the generalisation error on the second teacher at the switch. Principally we could ask many of the questions related to forgetting we ask in this paper for transfer as well in the sense that the teacher-student framework permits a natural transfer analogue. However forgetting has been the focus of this work and thus we include transfer plots here for the interested reader. 

To help distinguish these plots from those showing quantities associated with forgetting, we use a different color scheme for transfer related plots:

\begin{figure*}[h]
    \centering
    \begin{tikzpicture}
        \node at (0.2, 2.25) {\small $V$};
        \node at (4.7, 1.8) {\small 1};
        \node at (0.5, 1.8) {\small 0};
	    \node [opacity=1, anchor=west] at (0.3, 2.25) {\includegraphics[width=0.25\textwidth]{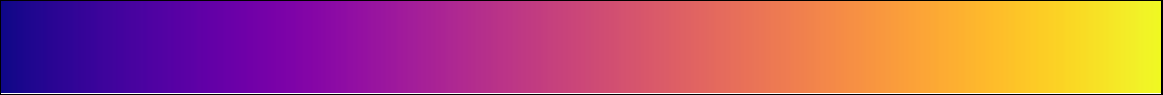}}; 
	\end{tikzpicture}
\end{figure*}

\begin{figure}[h]
	\pgfplotsset{
		width=1.1\textwidth,
		height=0.8\textwidth,
		scaled x ticks=false,
		every tick label/.append style={font=\tiny},
		y label style={at={(axis description cs:-0.16, 0.5)}, rotate=0, anchor=south},
		x label style={at={(axis description cs:0.5, -0.42)}, rotate=0, anchor=south},
		ymin=-6, ymax=-0.5,
		xlabel={\scriptsize $s$},
		}
	\begin{subfigure}[b]{0.235\textwidth}
	\begin{tikzpicture}
        \fill[shadecolor2, opacity=0] (0, 0) rectangle (\textwidth, 0.75\textwidth);
        \node [anchor=north west] at (0, 0.75\textwidth) {\emph{(a)}};
        \node [anchor=north west] at (1.6, 0.75\textwidth) {\scriptsize $\lambda=0$};
		\begin{axis}
			[
			at={(0.9cm, 0.75cm)},
			anchor=south west,
			xmin=0, xmax=12000000,
			ylabel={\scriptsize $\log\epsilon^\ddag$}
		]
		\addplot graphics [xmin=0, xmax=12000000,ymin=-6,ymax=-0.5] {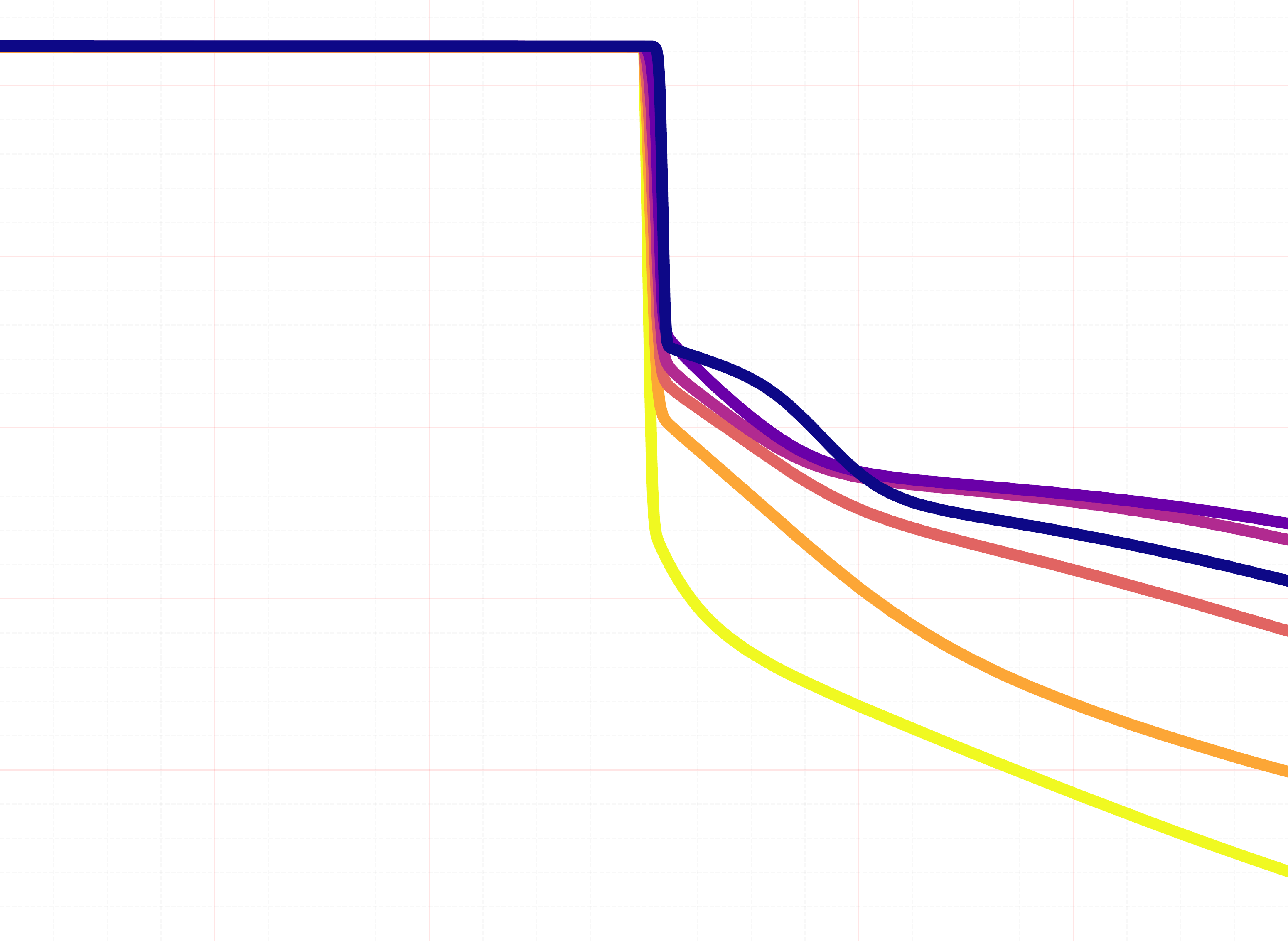};
		\end{axis}
	\end{tikzpicture}%
	\phantomcaption
	\label{fig:transfer_5a}
	\end{subfigure}
	\begin{subfigure}[b]{0.235\textwidth}
	\begin{tikzpicture}
        \fill[shadecolor2, opacity=0] (0, 0) rectangle (\textwidth, 0.75\textwidth);
        \node [anchor=north west] at (0, 0.75\textwidth) {\emph{(b)}};
        \node [anchor=north west] at (1.6, 0.75\textwidth) {\scriptsize $\lambda=100$};
		\begin{axis}
			[
			at={(0.9cm, 0.75cm)},
			anchor=south west,
			xmin=0, xmax=12000000,
			ylabel={\scriptsize $\log\epsilon^\ddag$}
		]
		\addplot graphics [xmin=0, xmax=12000000,ymin=-6,ymax=-0.5] {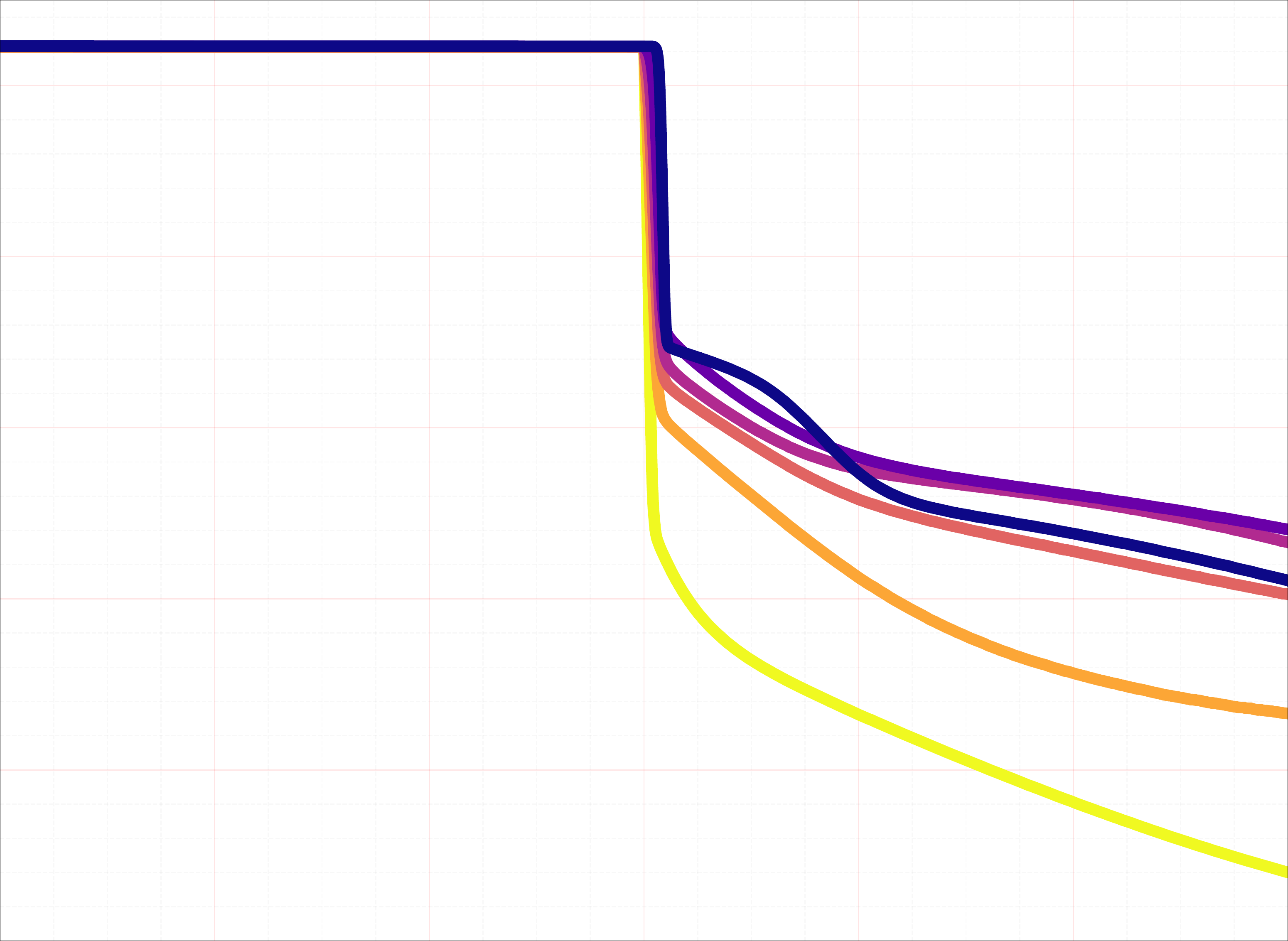};
		\end{axis}
	\end{tikzpicture}%
	\phantomcaption
	\label{fig:transfer_5b}
	\end{subfigure}
	\hspace{0em}
	\begin{subfigure}[b]{0.235\textwidth}
	\begin{tikzpicture}
        \fill[shadecolor2, opacity=0] (0, 0) rectangle (\textwidth, 0.75\textwidth);
        \node [anchor=north west] at (0, 0.75\textwidth) {\emph{(c)}};
        \node [anchor=north west] at (1.6, 0.75\textwidth) {\scriptsize $\lambda=1000$};
		\begin{axis}
			[
			at={(0.9cm, 0.75cm)},
			anchor=south west,
			xmin=0, xmax=12000000,
			ylabel={\scriptsize $\log\epsilon^\ddag$}
		]
		\addplot graphics [xmin=0, xmax=12000000,ymin=-6,ymax=-0.5] {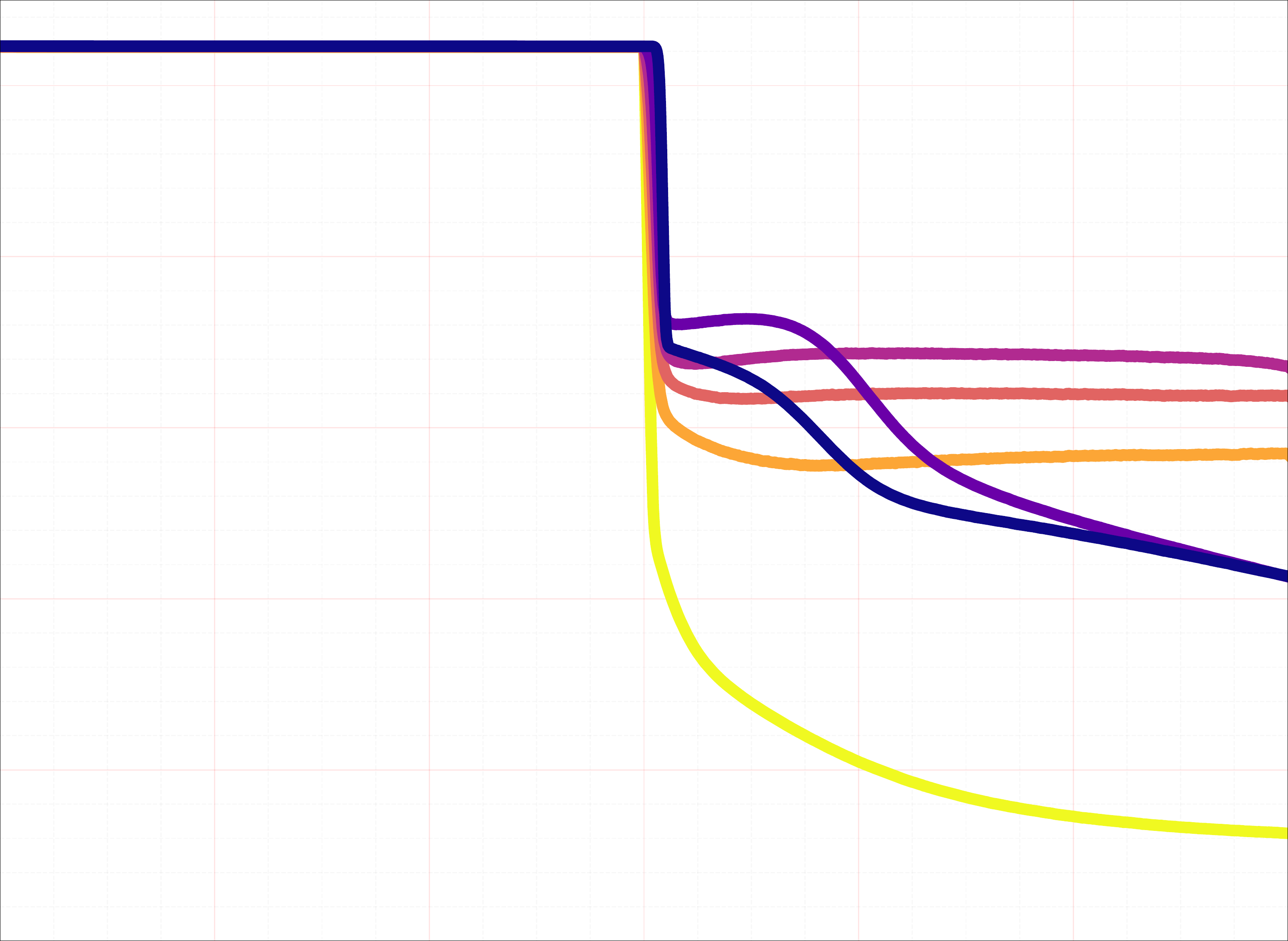};
		\end{axis}
	\end{tikzpicture}%
	\phantomcaption
	\label{fig:transfer_5c}
	\end{subfigure}
	\begin{subfigure}[b]{0.235\textwidth}
	\begin{tikzpicture}
        \fill[shadecolor2, opacity=0] (0, 0) rectangle (\textwidth, 0.75\textwidth);
        \node [anchor=north west] at (0, 0.75\textwidth) {\emph{(d)}};
        \node [anchor=north west] at (1.6, 0.75\textwidth) {\scriptsize $\lambda=10000$};
		\begin{axis}
			[
			at={(0.9cm, 0.75cm)},
			anchor=south west,
			xmin=0, xmax=12000000,
			ylabel={\scriptsize $\log\epsilon^\ddag$}
		]
		\addplot graphics [xmin=0, xmax=12000000,ymin=-6,ymax=-0.5] {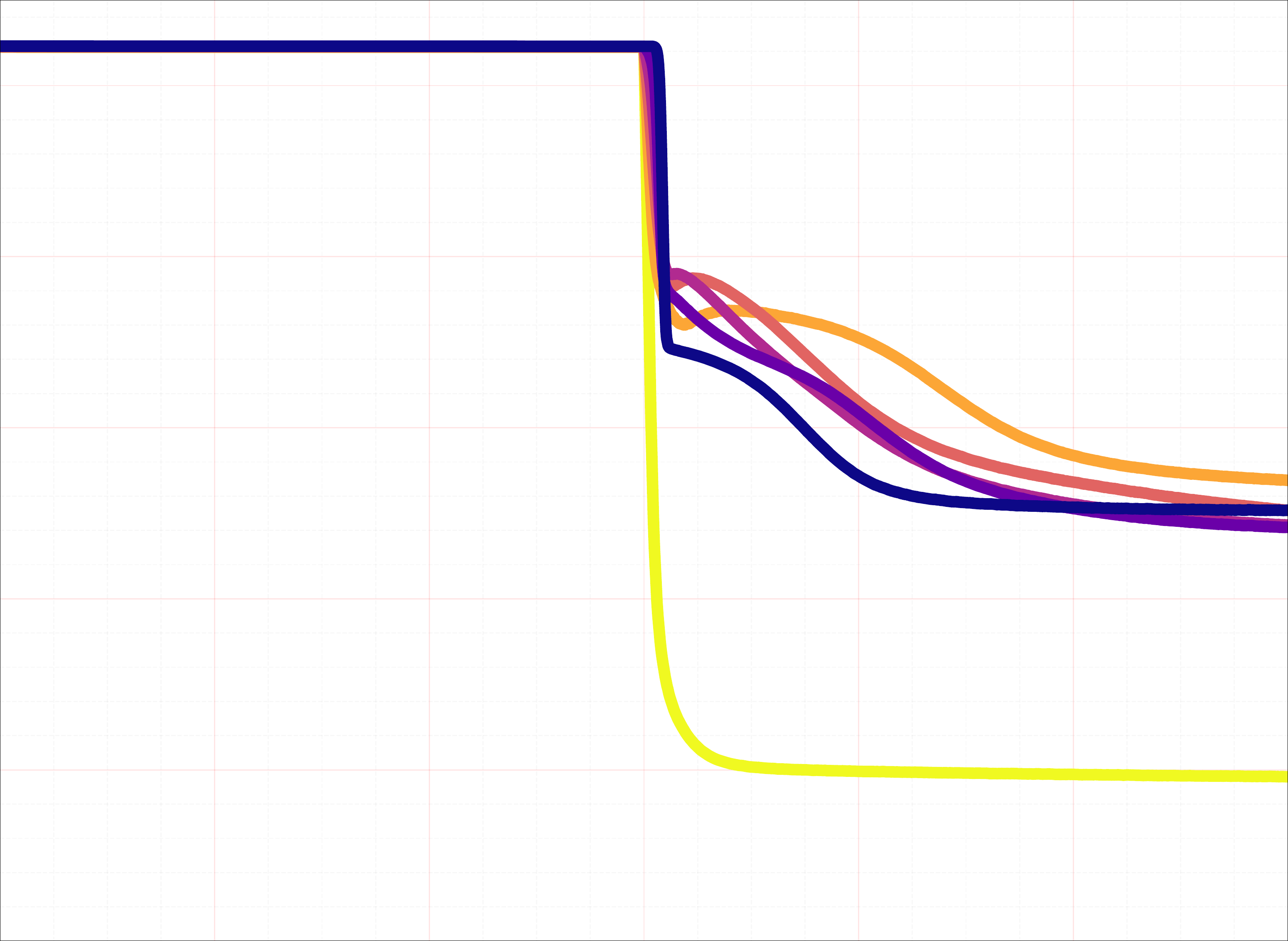};
		\end{axis}
	\end{tikzpicture}%
	\phantomcaption
	\label{fig:transfer_5d}
	\end{subfigure}
	\hspace{0em}
	\caption[]{\textbf{Effect of EWC on task similarity vs. forgetting: transfer}\label{fig:transfer_ewc}}
\end{figure}
\begin{figure}[h]
	\pgfplotsset{
		width=1.1\textwidth,
		height=0.8\textwidth,
		scaled x ticks=false,
		every tick label/.append style={font=\tiny},
		y label style={at={(axis description cs:-0.16, 0.5)}, rotate=0, anchor=south},
		x label style={at={(axis description cs:0.5, -0.42)}, rotate=0, anchor=south},
		xlabel={\scriptsize $s$},
		}
	\begin{subfigure}[b]{0.235\textwidth}
	\begin{tikzpicture}
        \fill[shadecolor2, opacity=0] (0, 0) rectangle (\textwidth, 0.75\textwidth);
        \node [anchor=north west] at (0, 0.75\textwidth) {\emph{(a)}};
        \node [anchor=north west] at (1.6, 0.75\textwidth) {\scriptsize $T=100$};
		\begin{axis}
			[
			at={(0.9cm, 0.75cm)},
			anchor=south west,
			xmin=0, xmax=12000000,
			ymin=-7.5, ymax=-0.5,
			ylabel={\scriptsize $\log\epsilon^\ddag$}
		]
		\addplot graphics [xmin=0, xmax=12000000,ymin=-7.5,ymax=-0.5] {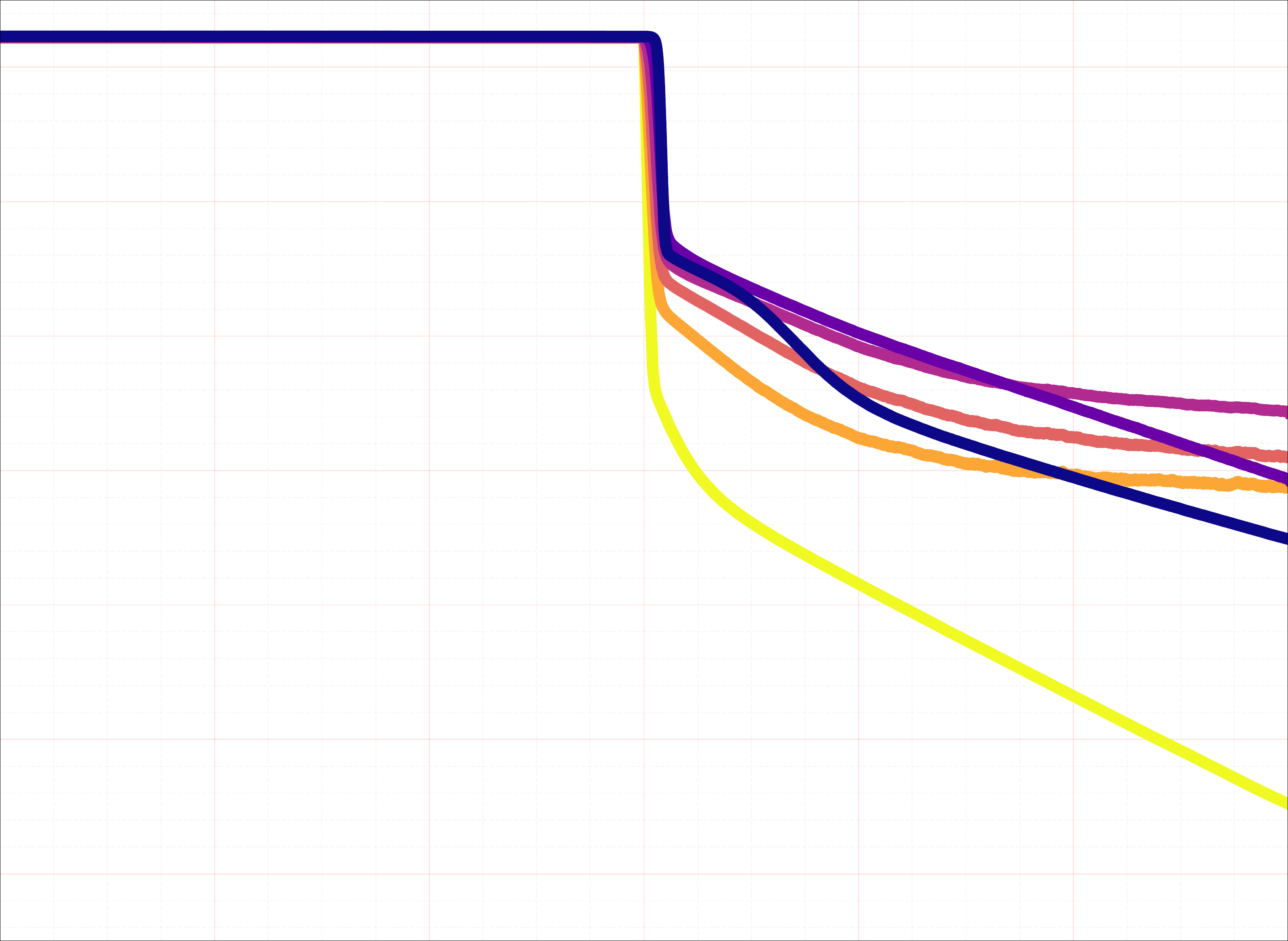};
		\end{axis}
	\end{tikzpicture}%
	\phantomcaption
	\label{fig:transfer_6a}
	\end{subfigure}
	\begin{subfigure}[b]{0.235\textwidth}
	\begin{tikzpicture}
        \fill[shadecolor2, opacity=0] (0, 0) rectangle (\textwidth, 0.75\textwidth);
        \node [anchor=north west] at (0, 0.75\textwidth) {\emph{(b)}};
        \node [anchor=north west] at (1.6, 0.75\textwidth) {\scriptsize $T=10$};
		\begin{axis}
			[
			at={(0.9cm, 0.75cm)},
			anchor=south west,
			xmin=0, xmax=12000000,
			ymin=-7.5, ymax=-0.5,
			ylabel={\scriptsize $\log\epsilon^\ddag$}
		]
		\addplot graphics [xmin=0, xmax=12000000,ymin=-7.5,ymax=-0.5] {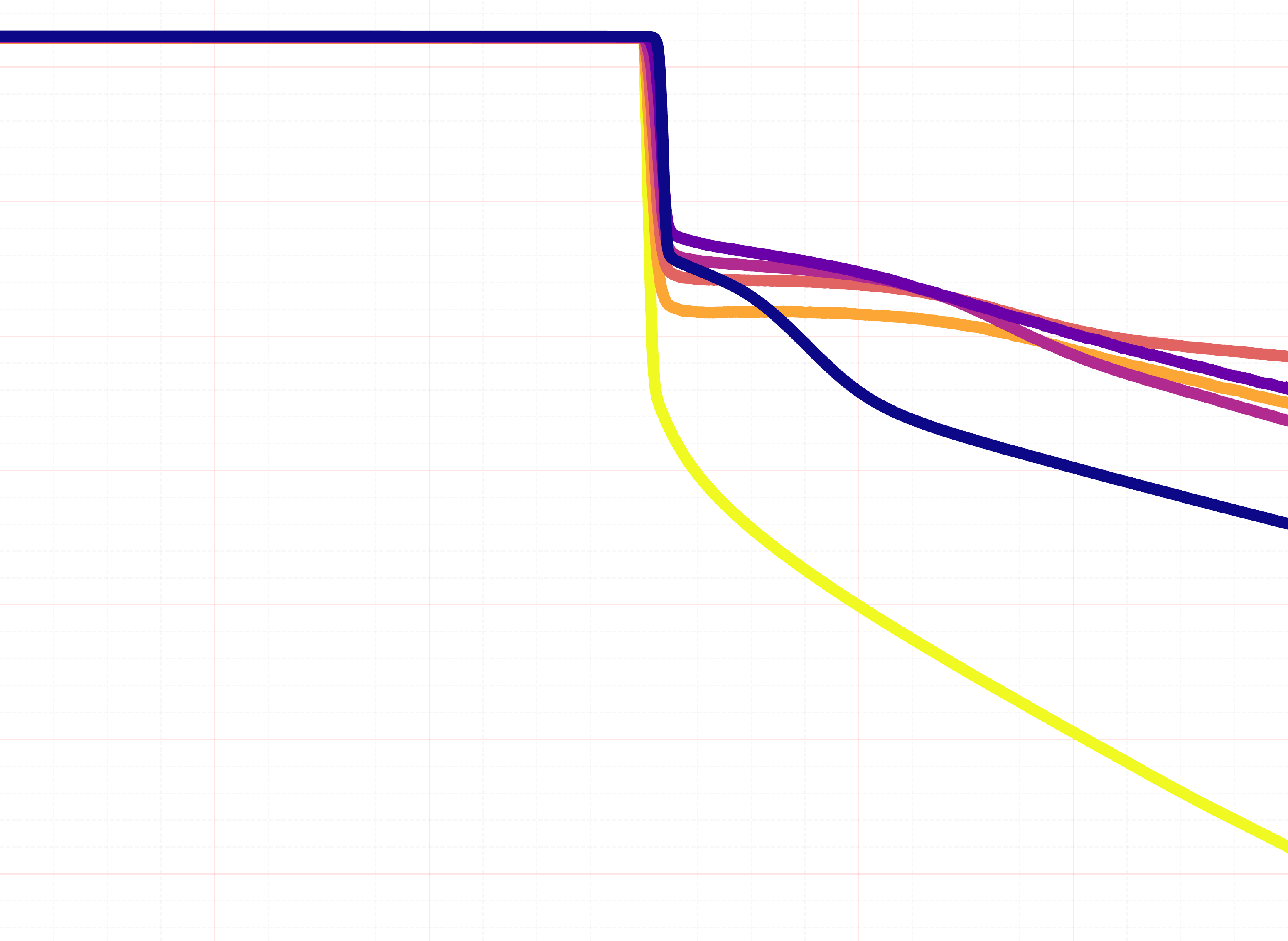};
		\end{axis}
	\end{tikzpicture}%
	\phantomcaption
	\label{fig:transfer_6b}
	\end{subfigure}
	\hspace{0em}
	\begin{subfigure}[b]{0.235\textwidth}
	\begin{tikzpicture}
        \fill[shadecolor2, opacity=0] (0, 0) rectangle (\textwidth, 0.75\textwidth);
        \node [anchor=north west] at (0, 0.75\textwidth) {\emph{(c)}};
        \node [anchor=north west] at (1.6, 0.75\textwidth) {\scriptsize $T=2$};
		\begin{axis}
			[
			at={(0.9cm, 0.75cm)},
			anchor=south west,
			xmin=0, xmax=12000000,
			ymin=-7.5, ymax=-0.5,
			ylabel={\scriptsize $\log\epsilon^\ddag$}
		]
		\addplot graphics [xmin=0, xmax=12000000,ymin=-7.5,ymax=-0.5] {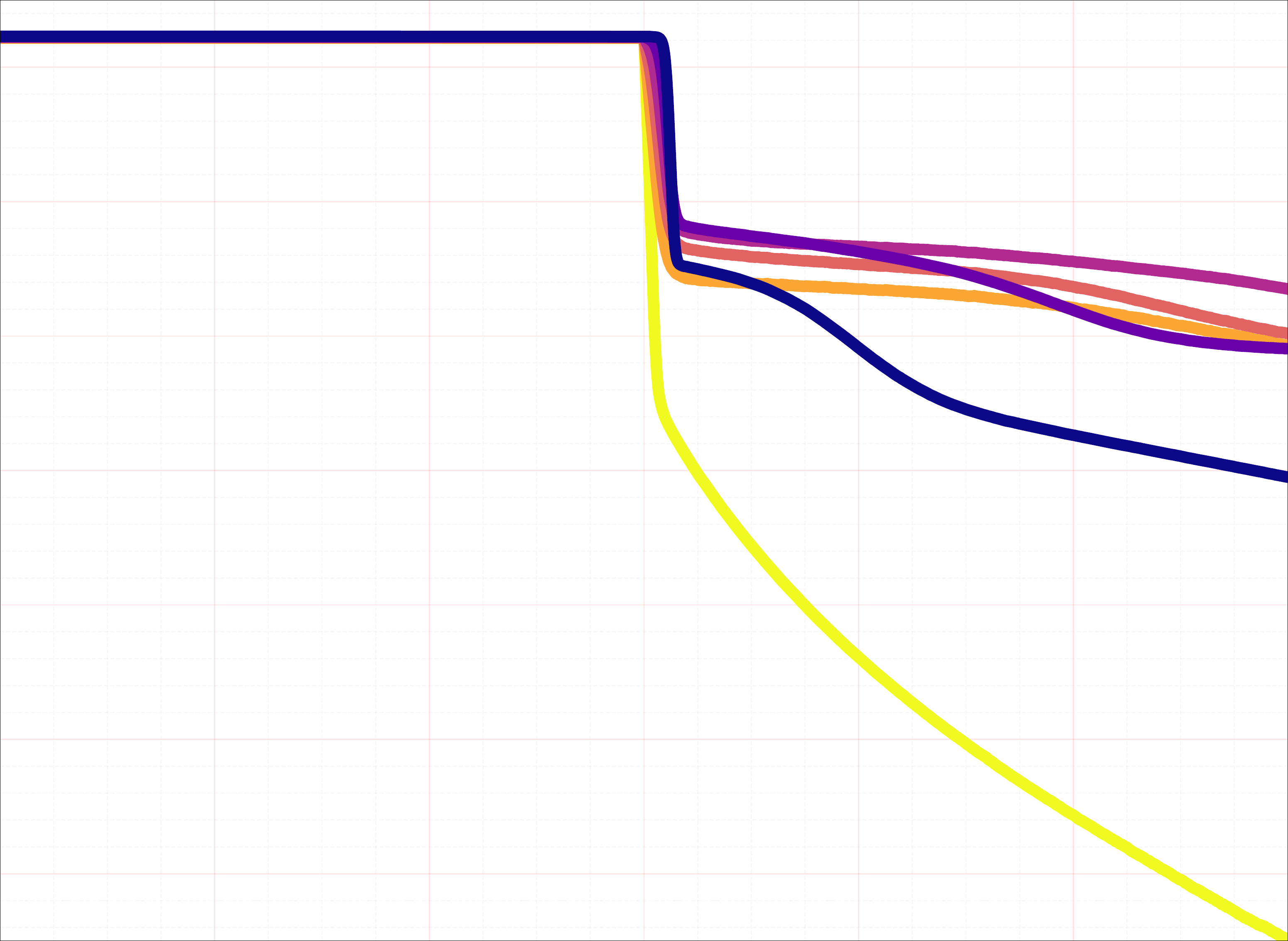};
		\end{axis}
	\end{tikzpicture}%
	\phantomcaption
	\label{fig:transfer_6c}
	\end{subfigure}
	\begin{subfigure}[b]{0.235\textwidth}
	\begin{tikzpicture}
        \fill[shadecolor2, opacity=0] (0, 0) rectangle (\textwidth, 0.75\textwidth);
        \node [anchor=north west] at (0, 0.75\textwidth) {\emph{(d)}};
        \node [anchor=north west] at (1.6, 0.75\textwidth) {\scriptsize $T=1$};
		\begin{axis}
			[
			at={(0.9cm, 0.75cm)},
			anchor=south west,
			xmin=0, xmax=12000000,
			ymin=-7.5, ymax=-0.5,
			ylabel={\scriptsize $\log\epsilon^\ddag$}
		]
		\addplot graphics [xmin=0, xmax=12000000,ymin=-7.5,ymax=-0.5] {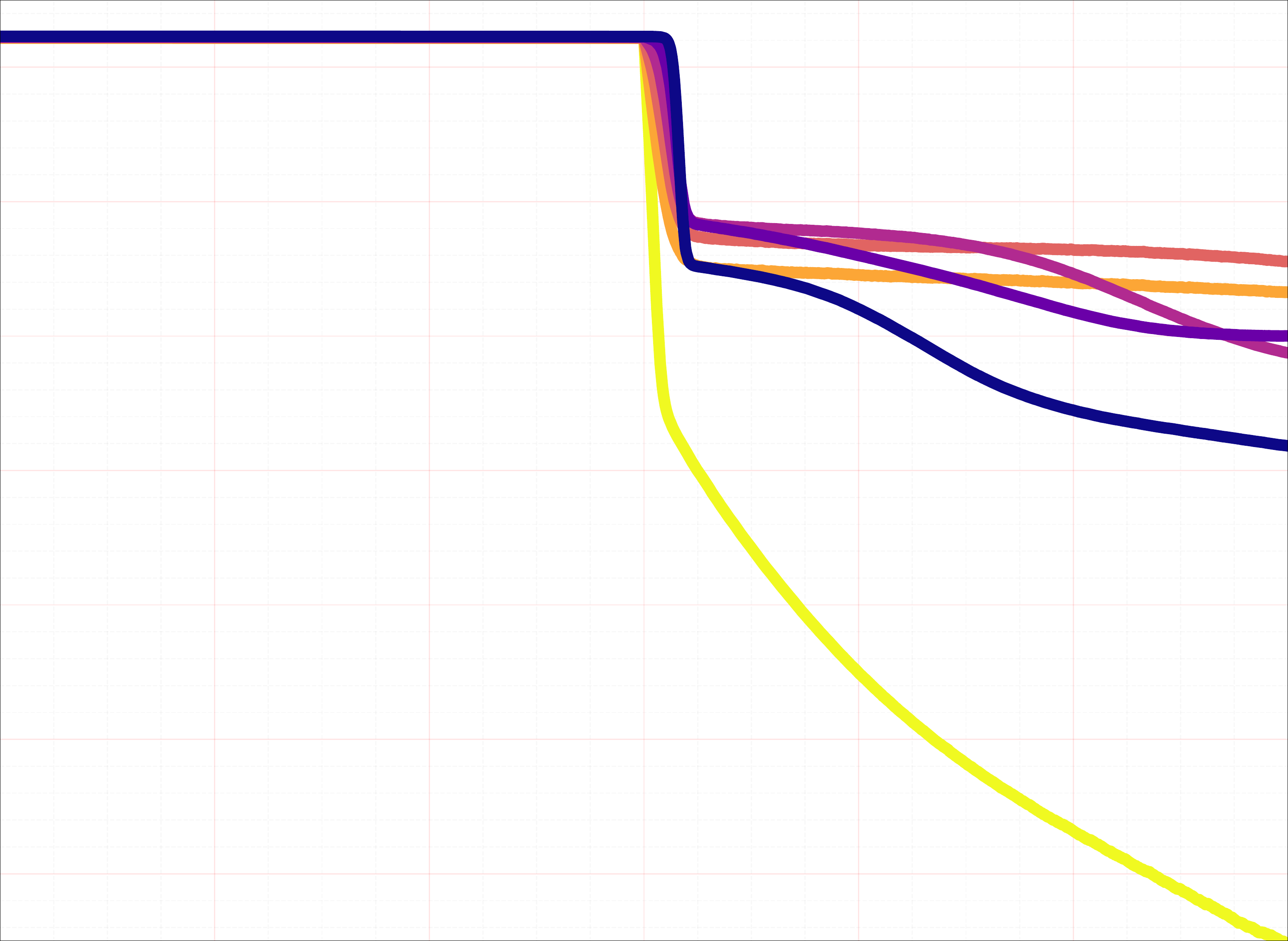};
		\end{axis}
	\end{tikzpicture}%
	\phantomcaption
	\label{fig:transfer_6d}
	\end{subfigure}
	\hspace{0em}
	\caption[]{\textbf{Effect of replay on task similarity vs. forgetting: transfer} \label{fig:transfer_interleave}}
\end{figure}
\begin{figure}[h]
\centering
    \pgfplotsset{
		width=0.4\textwidth,
		height=0.26\textwidth,
		scaled x ticks=false,
		y label style={at={(axis description cs:-0.12, 0.5)}, rotate=0, anchor=south},
		x label style={at={(axis description cs:0.5, -0.3)}, rotate=0, anchor=south},
		}
	\begin{tikzpicture}
        \fill[shadecolor2, opacity=0] (0, 0) rectangle (0.5\textwidth, 0.23\textwidth);
		\begin{axis}
			[
			at={(1.7cm, 0.9cm)},
			anchor=south west,
			xmin=0, xmax=20000000,
			ymin=-6.8, ymax=-0.5,
            xlabel={$s$},
			ylabel={$\log\epsilon^\ddag$},
		]
		\addplot graphics [xmin=0, xmax=20000000,ymin=-6.8,ymax=-0.5] {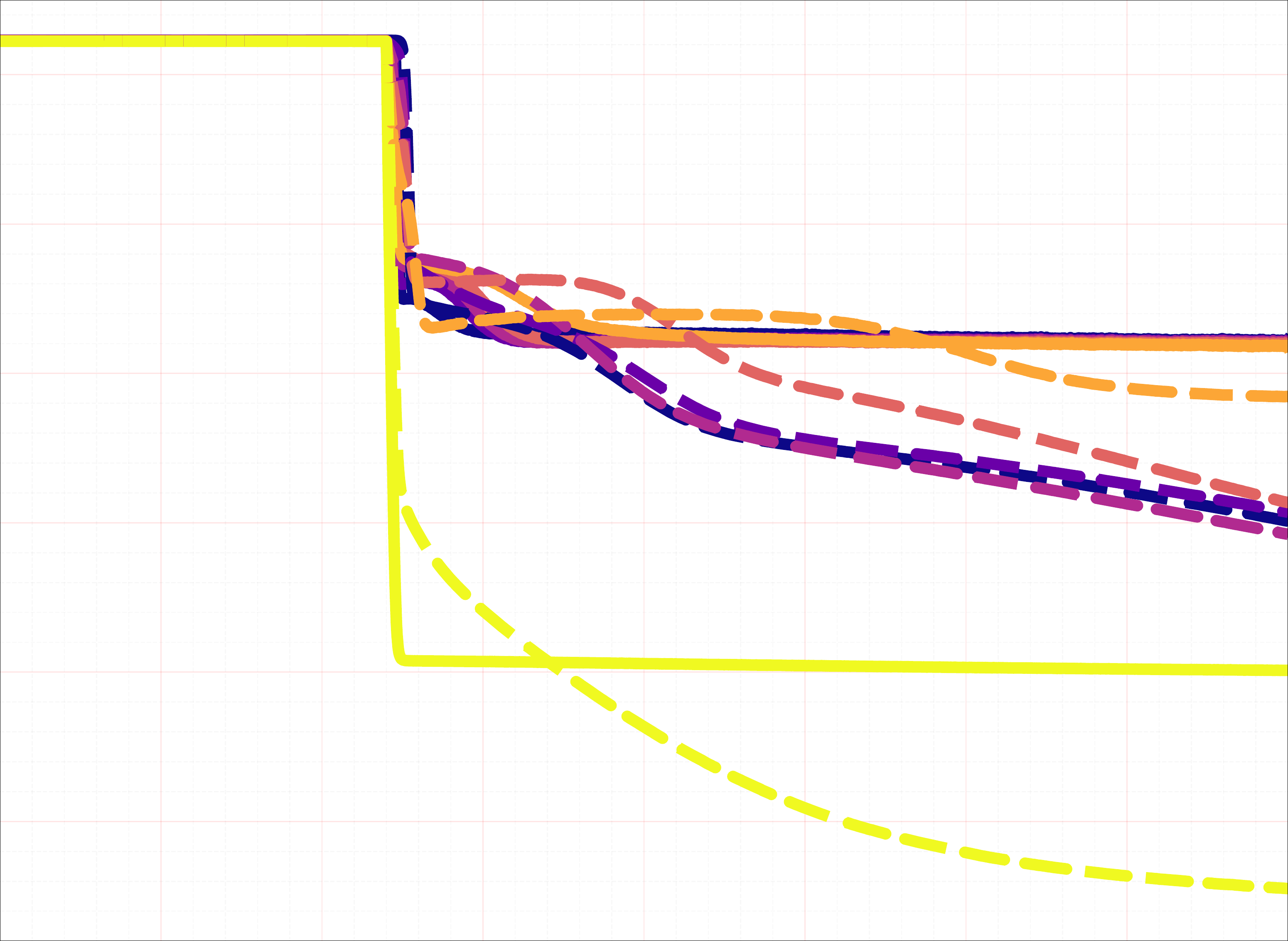};
		\end{axis}
	\end{tikzpicture}%
	\caption[]{\textbf{EWC vs. interleaved replay transfer} In~\autoref{fig:7a} we compared strong interleaving with strong elastic weight consolidation. Here we show the transfer performance. The solid lines show EWC and the dashed lines show interleaved training. You can see clearly that the performance of EWC plateaus; this is because movement in the node that specialised on the first task is so highly penalised and the network is less flexible as a result. On the other hand transfer performance (as well as backward transfer to the first task as seen by the plot in the main text) continues to improve under interleaved training for highly aligned and orthogonal tasks.\label{fig: transfer_combat_method_comparison}}
\end{figure}
\begin{figure}[h]
\centering
    \pgfplotsset{
		width=0.4\textwidth,
		height=0.26\textwidth,
		scaled x ticks=false,
		y label style={at={(axis description cs:-0.12, 0.5)}, rotate=0, anchor=south},
		x label style={at={(axis description cs:0.5, -0.3)}, rotate=0, anchor=south},
		}
	\begin{tikzpicture}
        \fill[shadecolor2, opacity=0] (0, 0) rectangle (0.5\textwidth, 0.23\textwidth);
		\begin{axis}
			[
			at={(1.7cm, 0.9cm)},
			anchor=south west,
			xmin=0, xmax=20000000,
			xtick={6000000, 20000000},
			ymin=-7.6, ymax=-0.5,
            xlabel={$s$},
			ylabel={$\log\epsilon^\ddag$},
		]
		\addplot graphics [xmin=0, xmax=20000000,ymin=-7.6,ymax=-0.5] {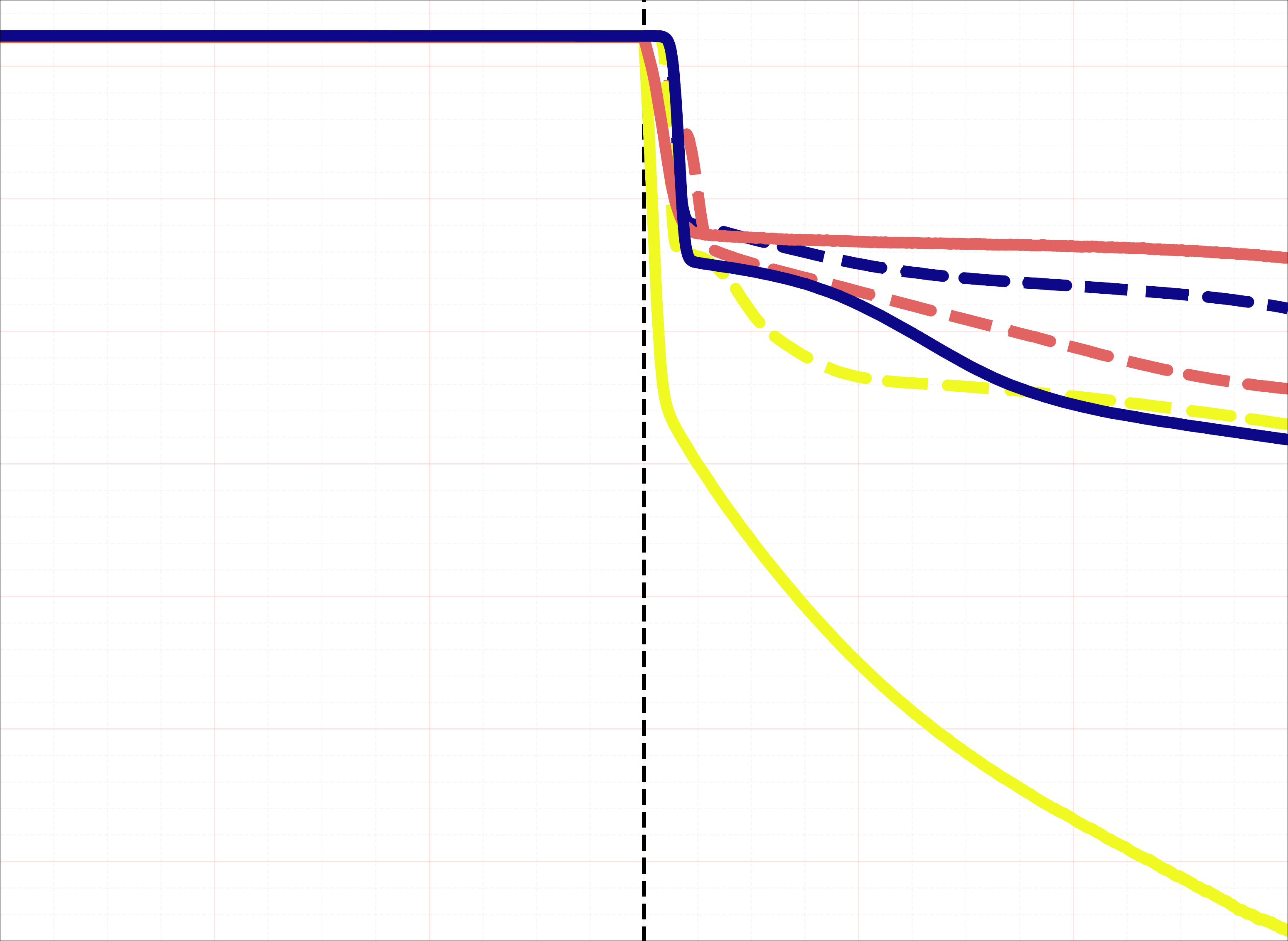};
		\end{axis}
	\end{tikzpicture}%
	\caption[]{\textbf{Catastrophic slowing transfer} In~\autoref{fig:7b} we identify an effect we call \emph{catastrophic slowing} where interleaved replay can not aid forgetting in the intermediate task regime. Here we show that transfer is also poor in this regime. The dashed lines show trajectories when we re-initialise at the task boundary. While transfer (and forgetting) is better for the aligned and orthogonal cases under interleaving than under re-initialising, it is worse for intermediately related tasks.\label{fig: transfer_catastrophic_slowing}}
\end{figure}
\begin{figure}[t]
\centering
    \pgfplotsset{
		width=0.4\textwidth,
		height=0.32\textwidth,
		scaled x ticks=false,
		y label style={at={(axis description cs:-0.12, 0.5)}, rotate=0, anchor=south},
		x label style={at={(axis description cs:0.5, -0.3)}, rotate=0, anchor=south},
		}
	\begin{tikzpicture}
        \fill[shadecolor2, opacity=0] (0, 0) rectangle (0.5\textwidth, 0.35\textwidth);
		\begin{axis}
			[
			at={(1.7cm, 1.4cm)},
			anchor=south west,
			xmin=0, xmax=1,
			ymin=2.5, ymax=6,
            xlabel={$V$},
			ylabel={$\mathcal{T}_{10,000}$},
		]
		\addplot graphics [xmin=0, xmax=1,ymin=2.5,ymax=6] {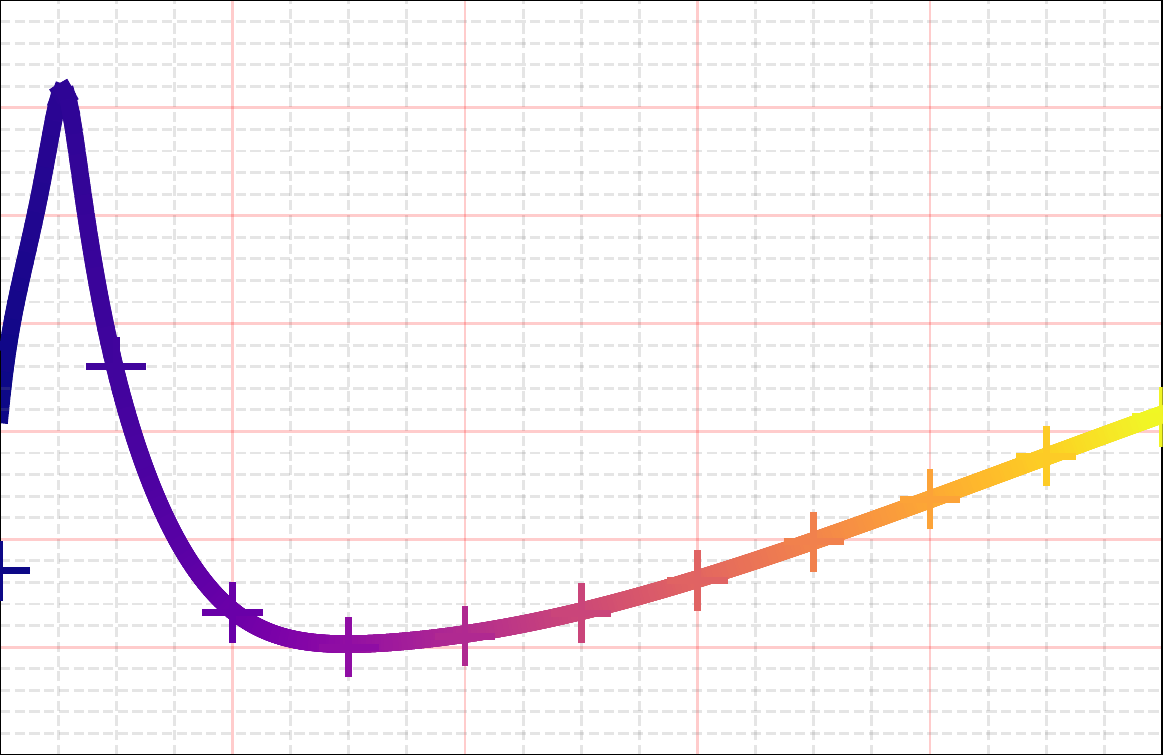};
		\end{axis}
	\end{tikzpicture}
	\caption[]{\textbf{Transfer vs. task similarity}: This is the transfer equivalent of~\autoref{fig:1b}. For a full discussion of the implications of this plot, see~\cite{lee2021continual} from which this is reproduced.}
\end{figure}

\end{document}